\pdfoutput=1

\documentclass[11pt]{article}

\usepackage[preprint]{acl}

\usepackage{times}
\usepackage{latexsym}

\usepackage[T1]{fontenc}

\usepackage[utf8]{inputenc}

\usepackage{microtype}

\usepackage{inconsolata}

\usepackage{bm,bbm}
\usepackage{subcaption}
\usepackage{graphicx}
\usepackage{amsmath}
\usepackage{amssymb}
\usepackage{booktabs}
\usepackage{adjustbox}
\usepackage{makecell}
\usepackage{multirow}
\usepackage{bbding}
\usepackage{xcolor}
\usepackage{color, colortbl}
\definecolor{citecolor}{HTML}{2980b9}
\definecolor{linkcolor}{HTML}{c0392b}

\usepackage{tabularx}
\usepackage{enumerate}
\usepackage{pifont}
\usepackage{wasysym}
\usepackage{float}

\usepackage{soul}
\usepackage{algorithm}
\usepackage{algorithmicx}
\usepackage{algpseudocode}

\usepackage{todonotes}

\usepackage{amsthm}
\newtheorem{theorem}{Theorem}

\newcommand{\name}{\textcolor{black}{\textsc{IntactKV}}}
\newcommand{\namebos}{\name$_\textrm{[B]}$}
\newcommand{\nameprompt}{\name$_\textrm{[P]}$}
\newcommand{\bos}{\texttt{[BOS]}~}
\newcommand{\m}[1]{\mathbf{#1}}

\newcommand\bK{\ensuremath{{\bm K}}}

\newcommand\bW{\ensuremath{{\bm W}}}

\newcommand\bq{\ensuremath{{\bm q}}}
\newcommand\bx{\ensuremath{{\bm x}}}
\newcommand\by{\ensuremath{{\bm y}}}

\newcommand\bh{\ensuremath{{\bm h}}}

\newcommand\bV{\ensuremath{{\bm V}}}

\newcommand\bs{\ensuremath{{\bm s}}}

\newcommand{\Rbb}{\mathbb{R}}

\newcommand\sm{{\rm softmax}}

\newcommand{\norm}[1]{{\Vert #1 \Vert}}
\newcommand{\norminf}[1]{{\Vert #1 \Vert}_\infty}
\newcommand{\normtwoinf}[1]{{\Vert #1 \Vert}_{2,\infty}}

\title{IntactKV: Improving Large Language Model Quantization by\\ Keeping Pivot Tokens Intact}

\author{Ruikang Liu$^1$ ~~~~~ Haoli Bai$^2$ ~~~~~ Haokun Lin$^3$ ~~~~~ Yuening Li$^4$ ~~~~~ Han Gao$^2$ \\ \textbf{Zhengzhuo Xu}$^1$ ~~~~~ \textbf{Lu Hou}$^2$ ~~~~~ \textbf{Jun Yao}$^2$ ~~~~~ \textbf{Chun Yuan}$^{1\dagger}$\thanks{\dag\ Corresponding author.}\\
$^{1}$Shenzhen International Graduate School, Tsinghua University ~~ $^{2}$Huawei Noah’s Ark Lab \\ $^{3}$Institute of Automation, Chinese Academy of Sciences ~~ $^{4}$The Chinese University of Hong Kong\\
\texttt{\normalsize \{liuruikang.cs, xzzthu\}@gmail.com ~~ \{baihaoli, han.g, houlu3, yaojun97\}@huawei.com}\\\texttt{\normalsize haokun.lin@cripac.ia.ac.cn ~~ yuening@link.cuhk.edu.hk ~~ yuanc@sz.tsinghua.edu.cn}
}

\begin{document}
\maketitle
\begin{abstract}
Large language models (LLMs) excel in natural language processing but demand intensive computation. To mitigate this, various quantization methods have been explored, yet they compromise LLM performance. 
This paper unveils a previously overlooked type of outliers in LLMs. Such outliers are found to allocate most of the attention scores on initial tokens of input, termed as \textit{pivot tokens}, which are crucial to the performance of quantized LLMs.
Given that, we propose \textit{IntactKV} to generate the KV cache of pivot tokens losslessly from the full-precision model. 
The approach is simple and easy to combine with existing quantization solutions with no extra inference overhead. Besides, \name\ can be calibrated as additional LLM parameters to boost the quantized LLMs further with minimal training costs. Mathematical analysis also proves that \name\ effectively reduces the upper bound of quantization error. 
Empirical results show that \name\ brings consistent improvement over various quantization methods across different LLMs and downstream tasks, leading to the new state-of-the-art for LLM quantization. The codes are available at \href{https://github.com/ruikangliu/IntactKV}{https://github.com/ruikangliu/IntactKV}.
\end{abstract}

\section{Introduction}

Large language models~(LLMs) have achieved remarkable progress in various tasks and benchmarks in natural language processing~\cite{brown2020language,bubeck2023sparks,touvron2023llama,team2023gemini}.   
Nonetheless, the rise of LLMs also increases computational intensity and memory requirements.
This motivates various research to decrease the inference cost of LLMs, e.g., quantizaiton~\cite{frantar2022gptq,shao2023omniquant,lin2023awq}, pruning~\cite{frantar2023sparsegpt,liu2023deja,sun2023simple,zhangplug}, and speculative decoding~\cite{chen2023accelerating,leviathan2023fast,cai2024medusa}, e.t.c.


Among these methods, network quantization converts the network parameters or activations from floating-point to fixed-point formats, which is a popular technique to reduce the model size and computational resources. Nevertheless, quantization inevitably affects the performance of LLMs. The leading cause comes from the outliers in LLM activations, which are sensitive to network quantization~\cite{dettmers2022llm, xiao2023smoothquant, lin2023awq}. As workarounds, there are efforts to either use mixed-precision formats~\cite{dettmers2022llm} or re-scale network weights of the outlier channels~\cite{lin2023awq}. 
These methods are all built based on the premise that outliers persist in fixed channels across all tokens. However, we find this is not the case for all outliers in LLMs.

In this paper, we discover a new type of outlier that is overlooked by previous quantization methods. These outliers exhibit extremely high values at only the \bos and some other common tokens (e.g., ``,'' and ``.'') at the beginning of the input, which is referred to as \textbf{\textit{pivot tokens}}.
We find the extreme values of these outliers make the self-attention concentrate on the pivot tokens, leaving the rest of the tokens untouched. This is also known as attention sinks~\cite{xiao2023efficient}, which is critical to the model performance~\cite{xiao2023efficient,bondarenko2023quantizable}. The effect of quantization on these pivot tokens should be carefully studied to improve the quantized LLMs.

\captionsetup[subfloat]{labelsep=none,format=plain,labelformat=empty,justification=centering}
\begin{figure*}[t]
\centering
\subfloat[(a) Output activations of LLaMA-30B Layer 24]{
        \includegraphics[width=0.24\linewidth]{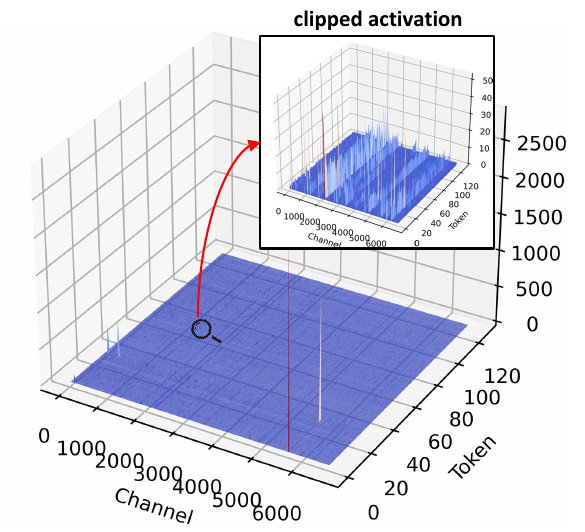}
        \label{fig:act+attn-a}
    }
\subfloat[(b) Output activations of LLaMA-2-7B Layer 24]{
        \includegraphics[width=0.24\linewidth]{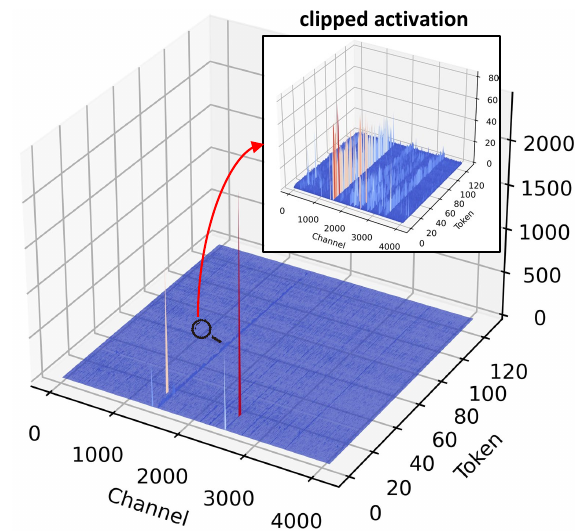}
        \label{fig:act+attn-b}
    }
\subfloat[(c) Attention map of \\LLaMA-30B Layer 24]{
        \includegraphics[width=0.24\linewidth]{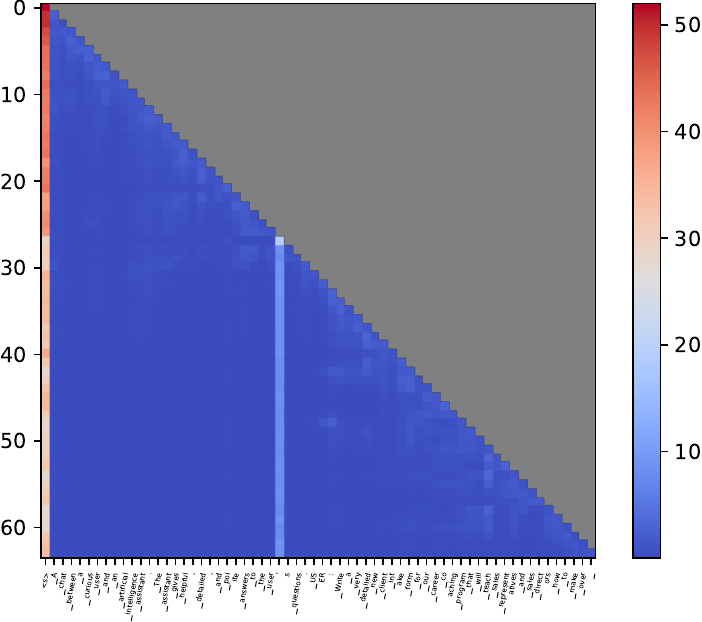}
        \label{fig:act+attn-c}
    }    
\subfloat[(d) Attention map of \\LLaMA-2-7B Layer 24]{
        \includegraphics[width=0.24\linewidth]{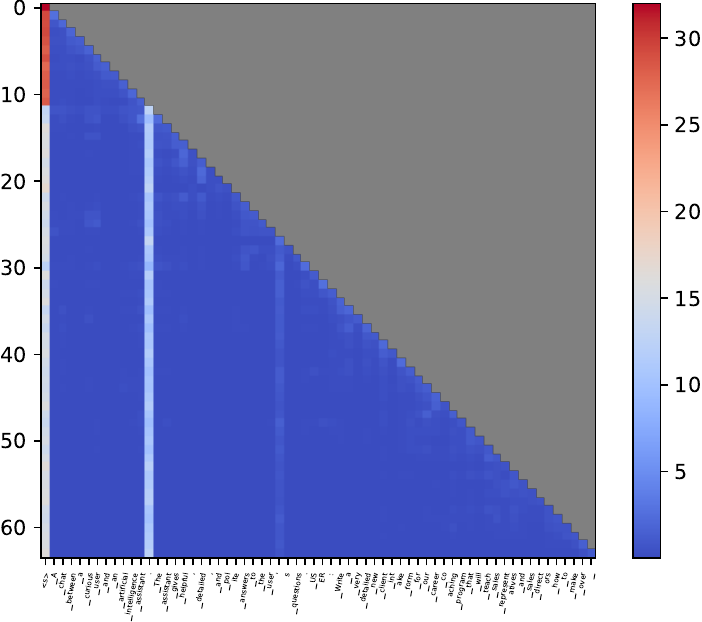}
        \label{fig:act+attn-d}
    }
	\caption{Visualizations of  Transformer output and attention scores of LLaMA-30B and LLaMA-2-7B. Observations: (1) There are token-specific outliers that can be orders of magnitudes larger than the rest of the tokens (enlarged in the box). Such tokens occur at the \bos token, the 28th token "\texttt{'}" in LLaMA-30B and 13th token "\texttt{.}" in LLaMA-2-7B, which are referred to as \textbf{\textit{pivot tokens}}; (2) These outliers over pivot tokens make the attention scores concentrated on themselves, which are likely to be affected by quantization. 
 More details can be found in Appendix~\ref{apdx-subsec:vis_details}.
 }
\label{fig:act+attn}
\end{figure*}

\begin{figure*}[t]
\centering
\subfloat[(a) LLaMA-13B]{
        \includegraphics[width=0.24\linewidth]{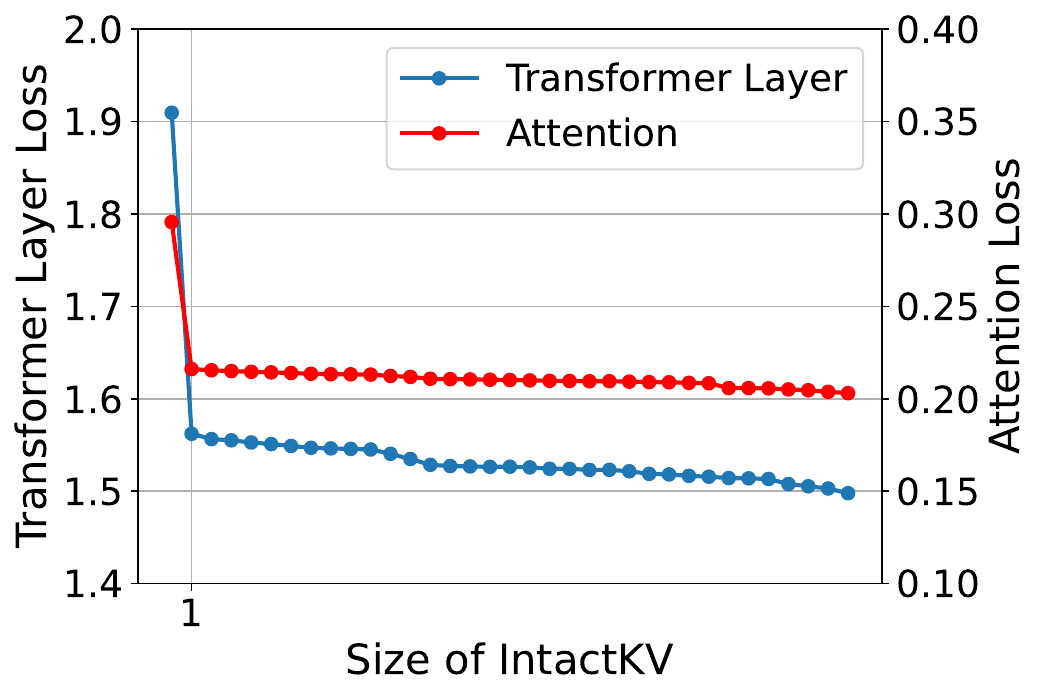}
        \label{fig:kv-size-a}
    }
\subfloat[(b) LLaMA-30B]{
        \includegraphics[width=0.24\linewidth]{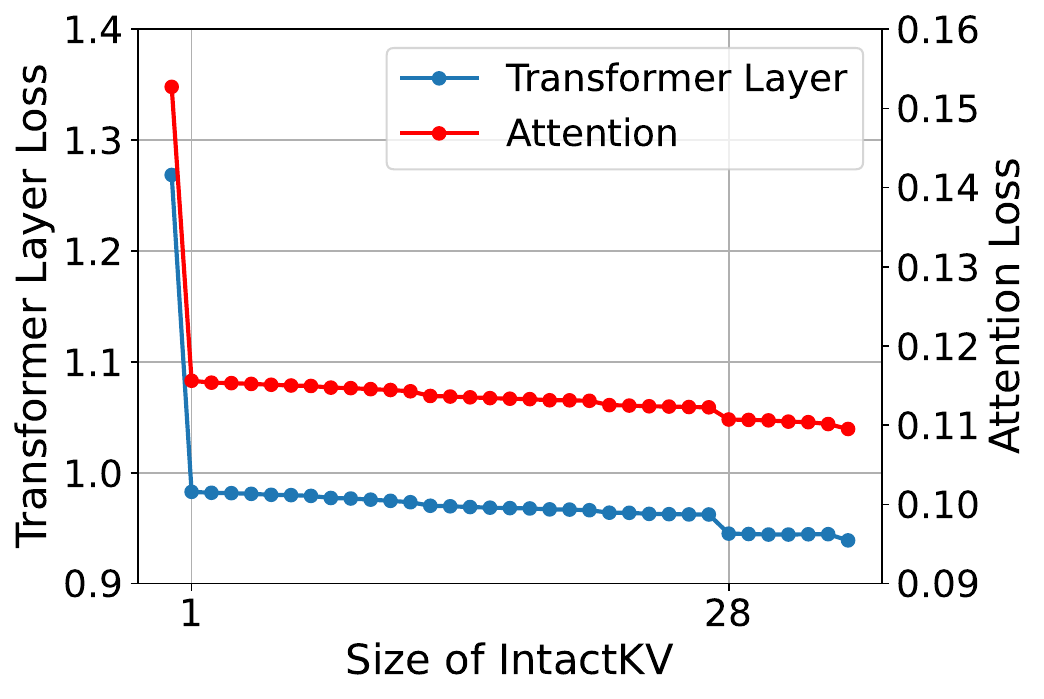}
        \label{fig:kv-size-b}
    }
\subfloat[(c) LLaMA-2-7B]{
        \includegraphics[width=0.24\linewidth]{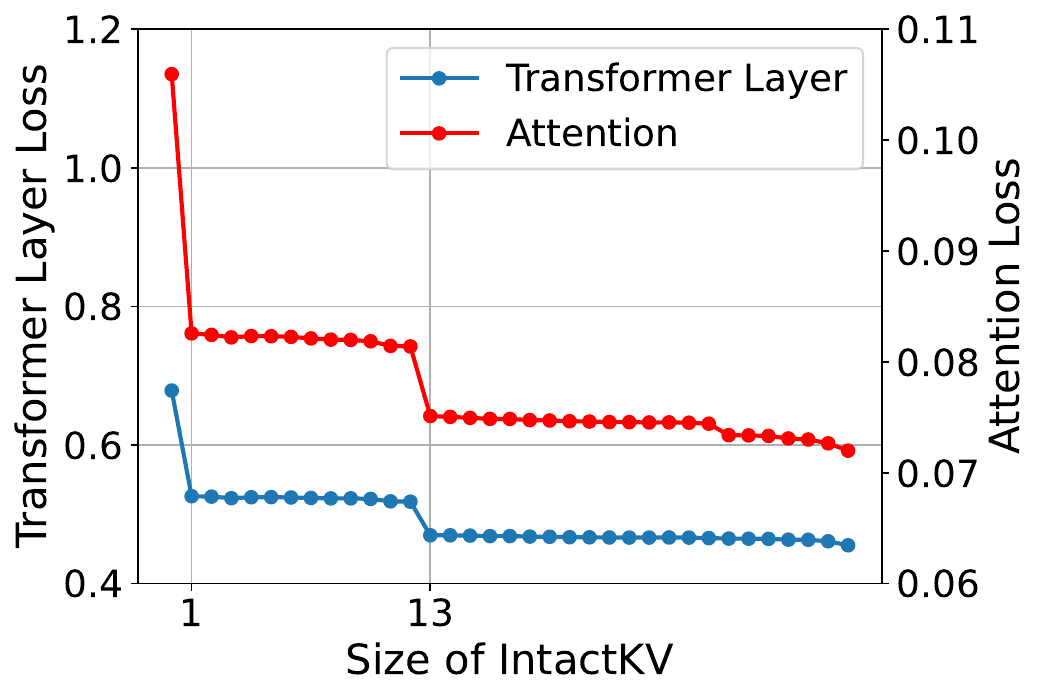}
        \label{fig:kv-size-c}
    }
\subfloat[(d) LLaMA-2-70B]{
        \includegraphics[width=0.24\linewidth]{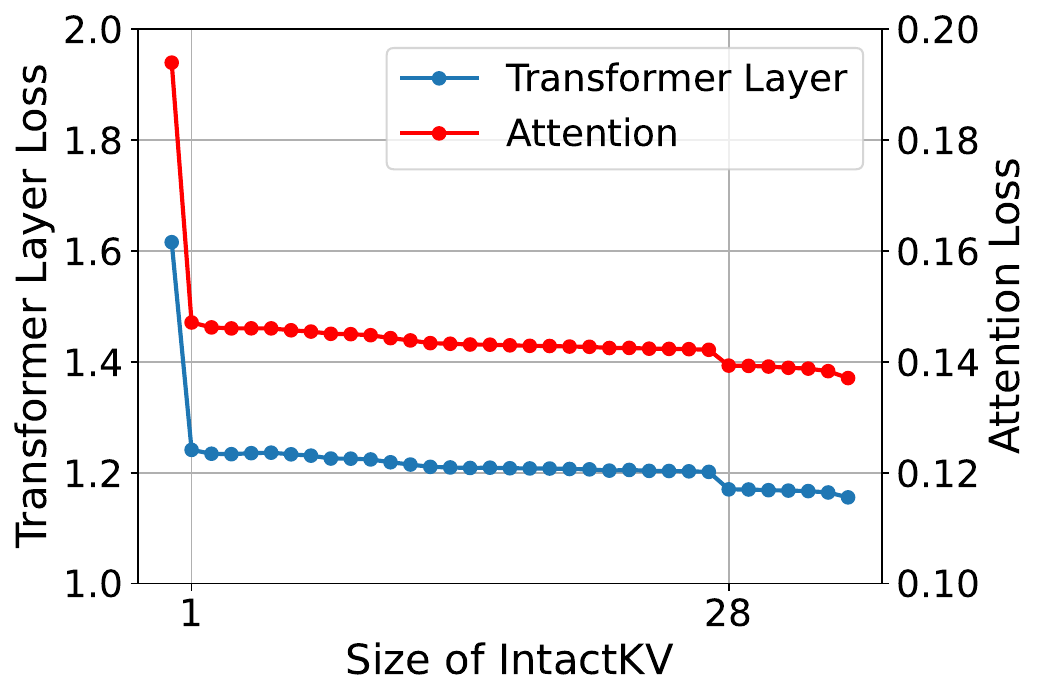}
        \label{fig:kv-size-d}
    }
\caption{The mean squared error (MSE) of the last Transformer layer and attention layers w.r.t. the varying sizes of \name.
Observations: (1) The MSE continues to drop as the size of \name\ increases. (2) Including the pivot tokens' KV cache in \name\ leads to the most significant decrease in the quantization loss, demonstrating the importance of the pivot tokens' KV cache. More experiment details can be found in Appendix~\ref{apdx-sec:kv-size-details}.}
\label{fig:kv-size}
\end{figure*}

Towards that end, we are motivated to propose \name, a simple strategy that is orthogonal to most existing quantization solutions.
The key idea behind \name\ is to \textit{generate the lossless KV cache of pivot tokens from the full-precision model}. By keeping the KV cache of pivot tokens intact, quantization error accumulated on the output of self-attention will be effectively alleviated in the rest of the decoding steps.  The integration of \name\ comes with no additional inference overhead.
Moreover, \name\ can also serve as \textit{extra trainable parameters} in addition to the LLM backbone. The calibration process of \name\ follows the convention of PTQ~\cite{bai2022towards,frantar2022gptq}, which further decreases the quantization error. 
To get more insights from \name, we also provide mathematical analysis and the results show that \name\ can effectively lower the upper bound of quantization error.

Empirical results show that \name\ consistently improves the capability of different quantization methods (e.g. AWQ~\cite{lin2023awq}, GPTQ~\cite{frantar2022gptq}, OmniQuant~\cite{shao2023omniquant} and QuaRot~\cite{ashkboos2024quarot}) on various open-sourced LLMs (e.g., LLaMA and Vicuna) across different tasks and benchmarks such as PPL, MMLU, commonsense QA, and MT-bench, achieving new state-of-the-art results for weight-only quantization as well as weight and activation quantization, e.g., lossless INT4 weight-only quantization for Vicuna-v1.3-13B on commonsense QA tasks.
Moreover, calibrating \name\ with INT4  quantization even matches the full-precision model on aligning with human preferences, as evaluated by GPT-4~\cite{bubeck2023sparks} on MT-bench.

\section{Motivation}
\label{sec:motivation}

\subsection{Preliminaries on LLM Quantization}
\label{subsec:preliminaries}
Network quantization is popularly studied in the literature of efficient LLMs~\cite{frantar2022gptq, lin2023awq, shao2023omniquant}. It allows larger throughput by reducing the model size and leads to practical inference speedup. 
Given the full-precision weight $\m w$, quantization aims to convert it to the low-bit representation $\hat{\m w}$. The general $b$-bit uniform quantization $\mathcal{Q}_b(\cdot)$ can be represented as
\begin{equation}
\label{eq:quant}
    \hat{\m w} = \mathcal{Q}_b(\m w) = s \cdot \Pi_{\Omega(b)}(\m w/s),
\end{equation}
where $s$ is the quantization step size, and $\Pi_{\Omega(b)}$ is the projection function onto the set of $b$-bit integers $\Omega(b) = \{0, 1, ..., 2^{b}-1\}$. 
While we mainly focus on weight-only quantization, Equation~\ref{eq:quant} can be similarly used to quantize activations and KV cache of LLMs to increase the inference throughput~\cite{xiao2023smoothquant,shao2023omniquant,hooper2024kvquant}.


Following most existing works in LLM quantization, we focus on post-training quantization~(PTQ)~\cite{frantar2022gptq,lin2023awq}, since it does not introduce extra training overhead as those in quantization-aware training~(QAT)~\cite{liu2023llm,li2023loftq}. 
Quantization inevitably downgrades LLMs in low-bit settings, where the outliers in quantized LLMs are found to be the cause of the deterioration~\cite{dettmers2022llm}.
In the next, we study the details of how these outliers affect the LLM quantization. 




\captionsetup[subfloat]{labelsep=none,format=plain,labelformat=empty,justification=centering}
\begin{figure*}[t]
\centering
\subfloat[(a) The overview of \name.]{
        \includegraphics[width=0.625\linewidth]{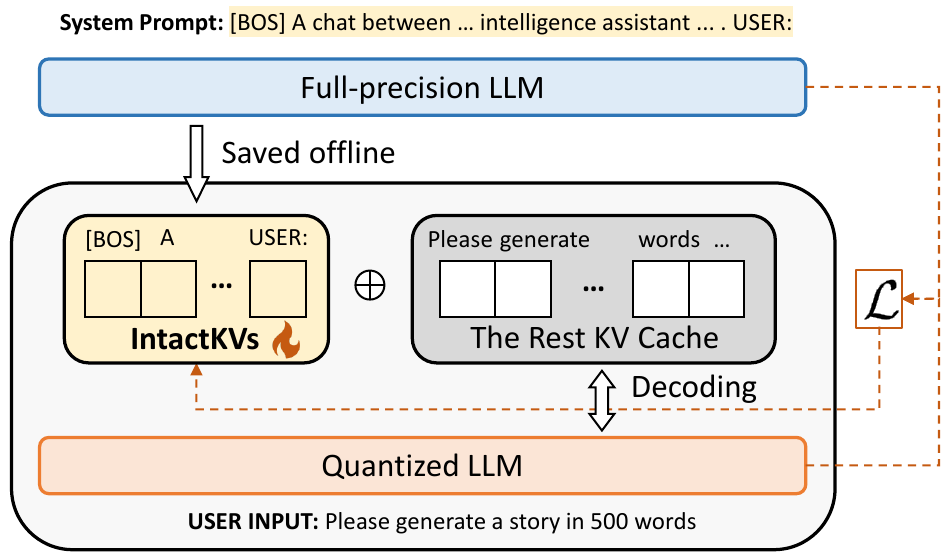}
        \label{fig:method-a}
    }
\subfloat[(b) Pseudo Code of Inference.]{
        \includegraphics[width=0.36\linewidth]{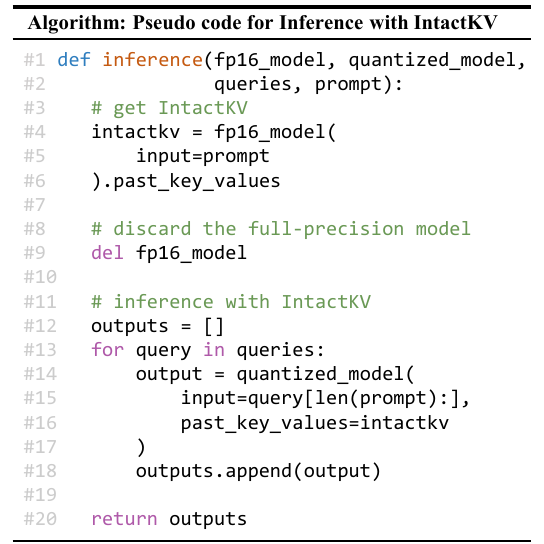}
        \label{fig:method-b}
    }
    \caption{The overview of the proposed \name\ applied for the supervised fine-tuned LLM. The full-precision model takes the system prompt as input and generates the \name\ losslessly as the prefix concatenated with the rest of the KV cache of quantized LLMs. \name\ can be further calibrated by minimizing the mean squared error $\mathcal{L}$ between the full-precision and quantized LLMs.}
\label{fig:method}
\end{figure*}

\subsection{Revisiting Outliers in LLMs}
\label{subsec:downgrade}

We discover a new type of outlier that is specific to particular tokens, which leads the attention sink~\cite{xiao2023efficient} that is critical to the performance of LLMs.




\paragraph{A New Variant of Outlier.}
Different from the outliers that persist in several fixed channels across different tokens~\citep{dettmers2022llm, xiao2023smoothquant, lin2023awq}, we find a new variant of outlier that is specific to some initial tokens of the input sequence.
By visualizing the activation of Transformer layer output in  Figure~\ref{fig:act+attn-a} and Figure~\ref{fig:act+attn-b}, there exist peaks with magnitudes over 1e3. 
These outliers can be hundreds of times larger than the previous outliers that persist in fixed channels across all tokens, as enlarged in Figure~\ref{fig:act+attn-a} and Figure~\ref{fig:act+attn-b}. More visualizations can be found in Appendix~\ref{apdx-sec:viss}.
It is found that such huge outliers usually occur at the \bos token and some other uninformative initial tokens (e.g., "\texttt{.}" or "\texttt{,}") at particular channels, regardless of the rest of the input sequence.
We thus name these tokens \textit{\textbf{pivot tokens}} given their dominating values in the activation.
Recently, a concurrent work~\citep{sun2024massive} also discovers such outliers with more detailed studies.




\paragraph{Pivot Tokens Exhibit Attention Sinks.}

We hypothesize that the outliers over these pivot tokens may propagate to queries and keys in the self-attention. Consequently, the attention scores will be concentrated on these pivot tokens than the rest ones, a.k.a \textit{attention sinks}~\cite{xiao2023efficient}.
To verify the hypothesis, we plot the attention scores in Figure~\ref{fig:act+attn-c} and Figure~\ref{fig:act+attn-d}. It can be found that the pivot tokens indeed dominate the attention scores, especially for the first token (i.e., \bos). This corresponds to the observations in attention sinks~\citep{xiao2023efficient}, which are empirically verified to be critical to the model performance. 
The recent study by~\citep{bondarenko2023quantizable} also shows that concentrating on these tokens naturally helps the attention head do nothing but simply a partial update of the residual.
In the decoding stage of LLMs, all generated tokens need to interact with pivot tokens through self-attention. However, as mentioned in Section~\ref{subsec:preliminaries}, network quantization would inevitably distort the output from the full-precision model. The concentrated scores of pivot tokens thus can be further deviated by quantization, which downgrades the model performance.

\section{Method}
\label{sec:method}

In this section, we introduce \textsc{IntactKV}, a simple and easy-to-implement method to improve the quantized LLMs. The key idea behind this is to keep the KV cache of the pivot tokens intact, i.e., without any distortion raised by quantization.
An overview of our method can be found in Figure~\ref{fig:method}. 



 \subsection{Preserving the KV Cache of Pivot Tokens}
\label{sec:pivotkv}



According to Section~\ref{subsec:downgrade}, the attention sinks of pivot tokens are likely to deteriorate by quantization. 
To alleviate this issue, we propose \name, a simple yet effective strategy to keep these pivot tokens intact. Specifically, as illustrated in Figure~\ref{fig:method-a}, we leverage the full-precision LLM to generate the lossless KV cache of pivot tokens, which is saved offline. The quantized LLM then loads \name\ as the prefix to concatenate with the rest of the KV cache and continues with the regular auto-regressive decoding process.
The pseudo code of the inference scheme with \name\ is presented in Figure~\ref{fig:method-b}.

In order to study the benefits of \name, we conduct a preliminary test on the mean squared error~(MSE) of the attention and transformer layer output. 
From Figure~\ref{fig:kv-size}, it is natural that the increasing size of \name\ gives the monotonically decreasing MSE on both the attention and transformer layers.
More importantly, it is found the pivot tokens observed in Section~\ref{subsec:downgrade} (e.g., \bos and other delimiter tokens) give the most significant decrease on the MSE, which demonstrates the importance of their KV cache. This aligns with the observations in Figure~\ref{fig:act+attn} that pivot tokens exhibit outliers with extreme values and attention sinks. 




\paragraph{The Choice of Pivot Tokens and \name.}
It is the key design to choose the pivot tokens and the associated \name. Given the observations in Figure~\ref{fig:kv-size}, one can naively pick pivot tokens with the most MSE reduction for \name. However, this is in fact not the case. Since \name\ acts as the prefix to the KV cache of quantized LLMs, it must start from the very first token, and be consecutive in length. This ensures it to be input agnostic, and the full-precision LLMs can be safely discarded once \name\ is generated.
Next, we provide practical solutions to this problem for different LLMs.

\begin{itemize}
\item For pre-trained LLMs, we propose the \name\ of size one that only contains \bos KV cache. 
It is a convention to prepend \bos to the input of pre-trained LLMs.
Moreover, as illustrated in Section~\ref{sec:motivation}, \bos is the pivot token with extreme outlier and attention scores. Besides, the KV cache of \bos has a great impact on the MSE of the quantized model. Employing a lossless \bos KV cache is thus believed to decrease the quantization loss. 
\item 
For supervised fine-tuned (SFT) models, when the input follows the system prompt, we argue that extending \name\ to the same length of the system prompt can further improve quantized LLMs. 
In addition to \texttt{[BOS]}, other tokens appearing at the beginning of the input sequence also have the potential to serve as pivot tokens (see Figure~\ref{fig:act+attn}). 
The system prompt is usually prepended to the input, which allows it to cover 
more pivot tokens.
As shown in Figure~\ref{fig:kv-size}, remedying the quantization error of these pivot tokens' KV cache can be helpful to compensate for the quantization error. 
We find that for Vicuna models, system prompt is enough to cover all the pivot tokens, more details can be found in Appendix~\ref{apdx-subsec:vicuna_vis}. 
\end{itemize}



\paragraph{Overhead of \name.}
Finally, we highlight that \name\ does not introduce extra latency overhead during inference. 
Besides, as \name\ is pre-computed, the pre-filling stage of the quantized LLMs can be accelerated as well. The memory overhead to save \name\ is also negligible compared with the LLM backbone. For instance, there are only 34 tokens of the system prompt for Vicuna-v1.5-7B, and thus \name\ takes only 0.13\% of the LLM model parameters.



\subsection{\name\ as Trainable Parameters}
\label{sec:trainable}

Since \name\ is pre-computed and saved offline, it can be treated as extra trainable parameters aside from the LLM backbone to further boost the quantized LLMs.
Despite there being no information loss at the pivot tokens, the quantization may still introduce errors to the KV cache during the decoding stage. 
As shown in Figure~\ref{fig:method-a}, we calibrate \name\ to compensate for the quantization error accumulated in the following tokens.
While there are various metrics to characterize the quantization discrepancy~\cite{frantar2022gptq,shao2023omniquant,liu2023llm}, we adopt the mean squared error of the transformer layer output between the full-precision LLM and quantized LLM, a simple yet most widely used metric, i.e., 
\begin{equation}
\label{eq:mse-loss}
\begin{aligned}
    \mathcal L(\Theta)=\frac{1}{2}\sum_{l=1}^L\|f_l(\m w, \m x)-f_l(\hat{\m w}, \m x; \Theta)\|_2^2,
\end{aligned}
\end{equation}
where $\Theta$ denotes the set of \name, $f_l$ is the mapping function for the $l$-th Transformer layer, and $L$ is the number of Transformer layers in LLM. $\m x$ is the input sequence, while  $\m w,\hat{\m w}$ are full-precision and quantized weights respectively. 
Note that the full-precision model is only required during the calibration process, and it can be safely discarded afterward. 
It is empirically found that calibration of system prompt  \name\ in SFT models generally gives more improvement than the calibration of \bos \name\ in pre-trained LLMs. This matches the intuition that a larger size of \name\ increases the potential to compensate for quantization errors. 


As we focus on the post-training quantization, the training of \name\ is highly lightweight since the only learnable parameters introduced are \name, i.e., the KV cache of pivot tokens. It takes only as few as 20 epochs on a calibration set with 128 samples. Besides, training with a quantized model further lowers the memory cost. The calibration process takes about only 10 minutes for a 7B model and less than 20 minutes for a 13B model on one computing device.

\subsection{Theoretical Analysis}
In this section, we provide a theoretical view of how the proposed \name\ benefits the quantized LLM. 
For the clarity of presentation, our analysis is built on the self-attention module of a Transformer layer, while it can be readily extended to the FFN module and multiple layers. 

Specifically, we denote $\bK, \bV\in \Rbb^{n\times d}$ as the KV cache during the decoding stage, and $\bq \in \Rbb^{d}$ is the query vector, where $n$ and $d$ are the sequence length and head dimension. Recall that the output of each attention head $\bh\in\Rbb^d$ is computed as
\begin{equation}
\bh = \sm(\bq\bK^\top/\sqrt{d})\bV\bW^O,
\end{equation}
where $\bW^O\in\Rbb^{d\times d}$ is the weight matrix of the projection layer. By quantizing the LLMs, there will be errors accumulated on the KV cache, denoted as $\Delta \bK , \Delta\bV \in \Rbb^{n\times d}$. 
Therefore, we are interested in showing how $\Delta \bK$ and $\Delta \bV$ are propagated to the change of attention head $\Delta \bh$, and to what extent \name\ alleviates the distortion.

\begin{theorem}\label{theo:error analysis}
    Given the query vector $\bq\in\Rbb^d$ and the change of KV caches $\Delta \bK , \Delta\bV \in \Rbb^{n\times d}$, the change of the attention head $\Delta\bh$ is bounded by 
    \begin{align*}
        \norm {\Delta\bh}_2 \le & \, C_1 \normtwoinf {\Delta \bK} \norm{\Delta\bV}_F \,+ \\
        &  + C_2 \normtwoinf {\Delta \bK}
        + C_3 \norm{\Delta\bV}_F,
    \end{align*}
    where $C_1 = \frac{n^{3/2} }{ \sqrt{d} } C_3 \norm\bq_2, C_2 = C_1 \norm{\bV}_2$ and $C_3=\norm{\bW^O}_2$.
\end{theorem}

The proof to Theorem~\ref{theo:error analysis} can be found in Appendix~\ref{appendix:proof}.
We preserve the terms w.r.t. $\Delta \bK$ and $\Delta \bV$ of interests, and leave the rest as constants. 
Note that $\Delta \bK$ can be further separated by the pivot tokens $\Delta \bK_{p}$ and rest tokens $\Delta \bK_{\textbackslash p}$, and similar notations hold for $\Delta \bV$.
Therefore, we have $\normtwoinf{\Delta \bK} = \max\big( \normtwoinf{\Delta \bK_p}, \normtwoinf{\Delta \bK_{\textbackslash p}}\big),$ and $\norm{\Delta\bV}_F= \sqrt{\norm{\Delta \bV_p}_F^2+ \norm{\Delta \bV_{\textbackslash p}}_F^2 }$.
With \name\, we have $\normtwoinf{\Delta \bK_p} = \|\Delta \bV_p\|_F = 0$ since they are generated losslessly, which decreases the upper bound of $\|\Delta \bh\|_2$. Moreover, it can further reduce the bound by incorporating more pivot tokens. This also aligns with the observation in Figure~\ref{fig:kv-size} that a larger size of \name\ gives a lower MSE of the attention module.







\section{Experiments}
\begin{table*}[t]
\begin{center}
\resizebox{\linewidth}{!}{
\begin{tabular}{l|ccccccc}
\toprule
\textbf{Method} & \textbf{LLaMA-7B} & \textbf{LLaMA-13B} & \textbf{LLaMA-30B} & \textbf{LLaMA-65B} & \textbf{LLaMA-2-7B} & \textbf{LLaMA-2-13B} & \textbf{LLaMA-2-70B} \\ 
\midrule
FP16            & 7.36               & 6.82                & 6.15                & 5.83                & 7.28               & 6.75                & 5.73                \\
\midrule
RTN             & 9.15               & 7.89                & 6.85                & 6.33                & 8.97               & 7.60                & 6.27                \\
\rowcolor{orange!10} 
\ \ +\namebos     & 8.52      & 7.66       & 6.69       & 6.20       & 8.61      & 7.48       & 6.13       \\ 
GPTQ            & 8.59               & 7.49                & 6.73                & 6.29                & 9.58               & 7.43                & 6.33                \\
\rowcolor{orange!10} 
\ \ +\namebos     & 8.30      & 7.42       & 6.62       & 6.23       & 9.27      & 7.36       & 6.28       \\ 
OmniQuant       & 8.26               & 7.39                & 6.65                & 6.18                & 8.35               & 7.43                & 6.12                \\
\rowcolor{orange!10} 
\ \ +\namebos     & 8.25      & 7.39       & 6.64       & 6.18       & 8.33      & 7.40       & 6.11       \\ 
AWQ             & 8.26               & 7.38                & 6.59                & 6.16                & 8.31               & 7.32                & 6.05                \\
\rowcolor{orange!10} 
\ \ +\namebos     & \textbf{8.12}      & \textbf{7.36}       & \textbf{6.54}       & \textbf{6.12}       & \textbf{8.18}      & \textbf{7.29}       & \textbf{6.04}       \\ 
\bottomrule
\end{tabular}
}
\end{center}
\vspace{-10pt}
\caption{INT3-group128 weight-only quantization results of LLaMA and LLaMA-2 Models on C4 dataset.
}
\label{tab:ppl}
\end{table*}

\begin{table*}[!t]
\begin{center}
\resizebox{\linewidth}{!}{
\begin{tabular}{l|ccccc|lclll}
\toprule
\textbf{Task Acc}            & \multicolumn{5}{c|}{\textbf{MMLU (5 shot)  average}}                                            & \multicolumn{5}{c}{\textbf{Common Sense QA (0 shot)  average}}       \\ \cmidrule{2-11} 
\textbf{Vicuna Family}  & \textbf{v1.5-7B} & \textbf{v1.5-13B} & \textbf{v1.3-7B} & \textbf{v1.3-13B} & \textbf{v1.3-33B} & \multicolumn{1}{c}{\textbf{v1.5-7B}} & \textbf{v1.5-13B} & \multicolumn{1}{c}{\textbf{v1.3-7B}} & \multicolumn{1}{c}{\textbf{v1.3-13B}} & \multicolumn{1}{c}{\textbf{v1.3-33B}} \\ \midrule
FP16                         & 49.84\%          & 55.78\%           & 47.12\%          & 52.10\%           & 59.30\%           & 65.33\%                              & 68.38\%           & 64.52\%                              & 67.22\%                               & 69.53\%                               \\ \midrule
RTN                          & 44.62\%          & 51.44\%           & 39.33\%          & 44.56\%           & 53.18\%           & 61.36\%                              & 66.12\%           & 59.05\%                              & 63.43\%                               & 67.33\%                               \\
\rowcolor{orange!10}                  
\ \ +\namebos     & 45.93\% & 51.89\%  & 41.74\% & 46.73\%  & 55.20\%  & 61.94\%                     & 65.91\%  & 61.26\%                     & 63.94\%                      & \textbf{67.95\%}                      \\ 
GPTQ                         & 43.99\%          & 52.95\%           & 40.12\%          & 47.83\%           & 55.84\%           & 58.61\%                              & 66.34\%           & 59.56\%                              & 65.11\%                               & 66.66\%                               \\
\rowcolor{orange!10}                  
\ \ +\namebos     & 44.86\% & 52.49\%  & 41.55\% & 48.53\%  & 56.32\%  & 59.12\%                     & 66.53\%  & 60.46\%                     & \textbf{65.13\%}                      & 67.93\%                      \\ 
OmniQuant                    & 46.62\%          & 52.82\%           & 42.95\%          & 48.23\%           & 55.21\%           & 62.30\%                              & 65.58\%           & 60.89\%                              & 64.62\%                               & 67.61\%                               \\
\rowcolor{orange!10}                  
\ \ +\namebos     & 46.27\% & 52.67\%  & 43.85\% & 48.31\%  & 55.51\%  & 62.01\%                     & 65.67\%  & 60.66\%                     & 64.89\%                      & 67.61\%                      \\ 
AWQ                          & 46.45\%          & 52.92\%           & 43.08\%          & 48.56\%           & 56.09\%           & 62.18\%                              & 66.51\%           & 60.75\%                              & 64.56\%                               & 67.67\%                               \\
\rowcolor{orange!10}                  
\ \ +\namebos     & \textbf{46.87\%} & \textbf{53.58\%}  & \textbf{44.67\%} & \textbf{49.05\%}  & \textbf{56.91\%}  & \textbf{62.49\%}                     & \textbf{66.93\%}  & \textbf{61.93\%}                     & 65.02\%                      & 67.90\%                      \\ 
\bottomrule
\end{tabular}

}
\end{center}
\vspace{-10pt}
\caption{INT3-group128 weight-only quantization results of Vicuna models
on 5-shot MMLU and 0-shot QA tasks.}
\label{tab:all-models}
\end{table*}

\subsection{Settings}

\paragraph{Models.}
We evaluate the proposed \name\ on various sizes of open-sourced LLMs, including LLaMA~\citep{touvron2023llama} (7B-65B), LLaMA-2~\citep{touvron2023llama2} (7B-70B), Vicuna-v1.3~\citep{chiang2023vicuna} (7B-33B) and Vicuna-v1.5 (7B-13B). We denote models that keep intact \bos KV as \namebos, and models that keep intact system prompt KV as \nameprompt. 

\paragraph{Quantization Methods.}
We mainly consider weight-only quantization methods, including round-to-nearest quantization (RTN), GPTQ \citep{frantar2022gptq}, the state-of-the-art OmniQuant \citep{shao2023omniquant} and AWQ \citep{lin2023awq}.
For GPTQ, we use AutoGPTQ with C4 calibration set following \citep{frantar2022gptq} to reproduce all results. For AWQ and OmniQuant, we use the official code or checkpoint with Pile \citep{gao2020pile} and WikiText2 \citep{merity2016wiki} calibration set respectively, following~\citep{lin2023awq, shao2023omniquant}. More implementation details can be found in Appendix~\ref{apdx-sec:quant-method-details}. We adopt asymmetric group-wise quantization with a group size of 128 and mainly focus on INT3 and INT4 quantization since INT8 is empirically lossless on various task metrics \citep{dettmers2022llm}.

Our \name\ can be readily combined with these existing weight-only quantization methods, and the experiment results are shown in Section~\ref{subsec:main-results}.
Moreover, aside from weight-only quantization, the proposed \name\ can be similarly applied for KV cache quantization and extended to activation quantization, as detailed in Section~\ref{sec:kvquant-exp} and Section~\ref{sec:actquant-exp}. It is worth noting that the integration of \name\ with weight-only/KV cache/activation quantization comes with no extra inference cost and works as an effective plugin to effectively boost the accuracy of quantized models.

\paragraph{Evaluation.}
For pre-trained LLMs (i.e., LLaMA and LLaMA-2), we report the perplexity (PPL) of language generation on C4 \citep{raffel2020c4} and WikiText2 \citep{merity2016wiki} dataset. 
For SFT models (i.e., Vicuna-v1.3 and v1.5), we conduct evaluation over a wide range of downstream tasks. We test the zero and five-shot performance on the Massively Multitask Language Understanding (MMLU)~\citep{hendrycks2020mmlu} benchmark. Meanwhile, we also evaluate seven zero-shot commonsense QA tasks: OBQA \citep{mihaylov2018openbookqa}, WinoGrande \citep{sakaguchi2021winogrande}, ARC-Challenge, ARC-Easy \citep{clark2018arc}, BoolQ \citep{clark2019boolq}, HellaSwag \citep{zellers2019hellaswag}, and LAMBADA \citep{paperno2016lambada}. Additionally, we evaluate quantized Vicuna on MT-bench \citep{zheng2023mt-bench}, a high-quality dataset consisting of 80 open-ended multi-turn questions, to gauge their alignment with human preferences. The responses generated by quantized models are judged by GPT-4 with a total score of 10. More evaluation details can be found in Appendix~\ref{apdx-sec:eval}.


\paragraph{Implementation Details}
\begin{table*}[!t]
\begin{center}
\resizebox{\linewidth}{!}{
\begin{tabular}{l|ccccc|ccccc}
\toprule
\multicolumn{1}{c|}{\multirow{2}{*}{\textbf{Method}}} & \multicolumn{5}{c|}{\textbf{MMLU (0 shot)}}    & \multicolumn{5}{c}{\textbf{MMLU (5 shot)}} \\ \cmidrule{2-11} 
\multicolumn{1}{c|}{}       & \textbf{Hums}    &  \textbf{STEM}   & \textbf{Social}  & \textbf{Others}  & \textbf{Avg}    & \textbf{Hums}    &  \textbf{STEM}    & \textbf{Social}  & \textbf{Others}  & \textbf{Avg}    \\ 
\midrule
FP16          & 47.89\%          & 39.96\%          & 58.86\%          & 57.34\%          & 50.77\%          & 49.78\%          & 40.46\%          & 60.61\%          & 58.24\%          & 52.10\%          \\ \cmidrule{1-11} RTN           & 42.06\%          & 32.87\%          & 47.61\%          & 49.51\%          & 43.02\%          & 42.42\%          & 34.46\%          & 50.34\%          & 51.57\%          & 44.56\%          \\
\rowcolor{orange!10}\ \ +\namebos & 42.49\%          & 35.35\% & 50.37\% & 52.44\% & 44.98\% & 44.65\% & 36.98\%          & 53.04\% & 52.84\% & 46.73\% \\ 
GPTQ          & 45.06\%          & 35.88\%          & 52.23\%          & 51.26\%          & 46.09\%          & 45.82\%          & 37.57\%          & 54.83\%          & 53.64\%          & 47.83\%          \\
\cellcolor{orange!10}\ \ +\namebos & \cellcolor{orange!10} 44.72\%          &\cellcolor{orange!10} 35.42\% &\cellcolor{orange!10} 52.94\% &\cellcolor{orange!10} 52.07\% &\cellcolor{orange!10} 46.22\% &\cellcolor{orange!10} 45.61\% &\cellcolor{orange!10} 38.34\%          &\cellcolor{orange!10} 55.83\% &\cellcolor{orange!10} 55.31\% &\cellcolor{orange!10} 48.53\% \\ 
OmniQuant     & 43.51\% & {36.85\%}          & 52.16\%          & 53.05\%          & 46.18\%          & 45.91\%          & 37.44\% & 55.31\%          & 54.94\%          & 48.23\%          \\
\rowcolor{orange!10}
\ \ +\namebos & 44.19\%          & 36.61\% & 53.33\% & 53.52\% & 46.72\% & 46.27\% & 37.54\%          & 54.99\% & 54.94\% & 48.31\% \\ 
AWQ           & 45.14\%          &   36.18\%        & 52.55\%          & 53.79\%          & 46.84\%          & {46.65\%}          &  37.64\%         & 55.54\%          & 54.87\%          & 48.56\%          \\
\rowcolor{orange!10}
\ \ +\namebos & {45.91\%}          & 36.65\% & {53.75\%} & {54.60\%} & \textbf{47.64\%} & 46.57\%  &         {38.40\%} & {56.03\%} & {55.95\%} & \textbf{49.05\%} \\ 
\bottomrule
\end{tabular}
}
\end{center}
\vspace{-10pt}
\caption{INT3-group128 weight-only quantization results of Vicuna-v1.3-13B on MMLU benchmarks.
}
\label{tab:mmlu}
\end{table*}

\begin{table*}[!t]
\begin{center}
\resizebox{\linewidth}{!}{
\begin{tabular}{l|l|cccccccc}
\toprule
\multicolumn{1}{c|}{\textbf{\#bits}} & \multicolumn{1}{c|}{\textbf{Method}} & \textbf{OBQA}    & \textbf{WinoGrande} & \textbf{ARC-C}   & \textbf{ARC-E}   & \textbf{BoolQ}   & \textbf{HellaSwag} & \textbf{LAMBADA} & \textbf{Avg}     \\ 
\midrule
FP16                        & -                           & 45.40\%          & 71.03\%          & 47.70\%          & 73.70\%          & 82.81\%          & 77.00\%          & 72.91\%          & 67.22\%          \\ \cmidrule{1-10} 
\multirow{8}{*}{w3g128}  
& RTN                         & 44.00\%          & 70.96\%          & 44.03\%          & 67.30\%          & 80.40\%          & 73.33\%          & 64.00\%          & 63.43\%         \\
&  
\cellcolor{orange!10}\ \ +\namebos   & \cellcolor{orange!10}44.80\% & \cellcolor{orange!10}69.93\% & \cellcolor{orange!10}45.05\% & \cellcolor{orange!10}68.35\% & \cellcolor{orange!10}79.42\% & \cellcolor{orange!10}74.81\% & \cellcolor{orange!10}65.22\% & \cellcolor{orange!10}63.94\%  \\                           
& GPTQ                        & 45.20\%          & 69.77\%          & 46.08\%          & 70.33\%          & 81.90\%          & 74.89\%          & 67.59\%          & 65.11\%          \\
&
\cellcolor{orange!10}\ \ +\namebos   & \cellcolor{orange!10}44.00\% & \cellcolor{orange!10}70.80\% & \cellcolor{orange!10}44.97\% & \cellcolor{orange!10}70.75\% & \cellcolor{orange!10}81.35\% & \cellcolor{orange!10}75.03\% & \cellcolor{orange!10}69.03\% & \cellcolor{orange!10}\textbf{65.13\%} \\                          
& OmniQuant                   & 45.20\%          & 69.22\%          & 45.22\%          & 68.90\%          & 80.95\%          & 74.72\%          & 68.15\%          & 64.62\%          \\
&
\cellcolor{orange!10}\ \ +\namebos   & \cellcolor{orange!10}45.40\% & \cellcolor{orange!10}70.32\% & \cellcolor{orange!10}45.31\% & \cellcolor{orange!10}68.86\% & \cellcolor{orange!10}81.28\% & \cellcolor{orange!10}74.52\% & \cellcolor{orange!10}68.52\% & \cellcolor{orange!10}64.89\% 
\\                       
& AWQ                         & 42.80\%          & 68.98\%          & 46.08\%          & 68.98\%          & 81.31\%          & 74.97\%          & 68.78\%          & 64.56\%          \\
&
\cellcolor{orange!10}\ \ +\namebos   & \cellcolor{orange!10}43.20\% & \cellcolor{orange!10}69.46\% & \cellcolor{orange!10}46.16\% & \cellcolor{orange!10}69.74\% & \cellcolor{orange!10}81.80\% & \cellcolor{orange!10}75.11\% & \cellcolor{orange!10}69.67\% & \cellcolor{orange!10}65.02\% 
\\ \cmidrule{1-10} 
\multirow{8}{*}{w4g128}  
& RTN                         & 45.20\%          & 71.43\%          & 48.04\%          & 73.15\%          & 82.87\%          & 76.56\%          & 70.62\%          & 66.84\%          \\
&
\cellcolor{orange!10}\ \ +\namebos   & \cellcolor{orange!10}44.80\% & \cellcolor{orange!10}71.51\% & \cellcolor{orange!10}47.44\% & \cellcolor{orange!10}73.36\%& \cellcolor{orange!10}82.75\% & \cellcolor{orange!10}77.01\% & \cellcolor{orange!10}70.99\% & \cellcolor{orange!10}66.84\% \\                           
& GPTQ                        & 44.60\% & 70.01\%          & 47.87\% & 73.32\%          & 82.23\%          & 76.55\%          & 71.78\% & 66.62\%          \\
&
\cellcolor{orange!10}\ \ +\namebos   & \cellcolor{orange!10}45.00\% & \cellcolor{orange!10}71.35\% & \cellcolor{orange!10}46.76\% & \cellcolor{orange!10}73.02\% & \cellcolor{orange!10}83.33\% & \cellcolor{orange!10}77.00\% & \cellcolor{orange!10}71.55\% & \cellcolor{orange!10}66.86\% \\                         
& OmniQuant                   & 45.60\%          & 70.56\%          & 46.76\%          & 73.02\%          & 82.81\%          & 76.74\%          & 70.41\%          & 66.56\%          \\
&
\cellcolor{orange!10}\ \ +\namebos   & \cellcolor{orange!10}45.20\% & \cellcolor{orange!10}71.43\% & \cellcolor{orange!10}46.25\% & \cellcolor{orange!10}72.52\% & \cellcolor{orange!10}82.63\% & \cellcolor{orange!10}76.90\% & \cellcolor{orange!10}70.31\% & \cellcolor{orange!10}66.46\% \\                      
& AWQ                         & 45.20\%          & 70.32\%          & 47.27\%          & 73.91\%          & 82.81\%          & 76.79\%          & 71.32\%          & 66.80\%          \\
\rowcolor{orange!10} \cellcolor{white!0}
&
\ \ +\namebos  & 45.60\%          & 71.19\% & 47.10\%          & 73.32\% & 82.72\% & 76.95\% & 71.38\%          & \textbf{66.89\%} \\ 
\bottomrule
\end{tabular}
}
\end{center}
\vspace{-10pt}
\caption{Weight-only quantization results of Vicuna-v1.3-13B on seven 0-shot commonsense QA tasks.
}
\label{tab:qa}
\end{table*}

For evaluation on PPL, MMLU, and commonsense QA tasks, we adopt \namebos\ that only includes \bos KV since the input sequence of these tasks does not use any system prompt. 
For evaluation of SFT models on MT-bench, we adopt \nameprompt\ to keep an intact system prompt KV cache. The system prompt of Vicuna can be found in Appendix~\ref{apdx-sec:prompt}.
For training the cached \name, we randomly sample 128 samples from ShareGPT\footnote{\url{https://huggingface.co/datasets/Aeala/ShareGPT_Vicuna_unfiltered}} dataset as our calibration dataset, consisting of multi-turn ChatGPT \citep{chatgpt} conversations. 
The layer-wise MSE loss defined in Equation~\ref{eq:mse-loss} is calculated on the response of ChatGPT.
We use AdamW optimizer with learning rate $2\times10^{-4}$, training for 160 optimizer update steps with a gradient accumulation step of 16, i.e., 20 epochs. 
As mentioned in Section~\ref{sec:trainable}, training \namebos\ leads to comparable performance compared with vanilla \name. Instead, the calibration of \nameprompt\ has more potential to improve quantized LLMs with longer system prompt. 
Thus, we primarily evaluate the \nameprompt\ with KV cache of system prompt as trainable parameters in the following experiments.
For weight and activation quantization, we further quantize \name\ to lower bits to avoid extra inference overhead, which only incurs negligible accuracy loss. More details of activation quantization can be found in Section~\ref{sec:actquant-exp}.

\subsection{Main Results}
\label{subsec:main-results}

\paragraph{Results on Language Generation Tasks.}
We first integrate our proposed \name\ with RTN, GPTQ, OmniQuant, and AWQ on LLaMA and LLaMA-2 models.
The effect of this integration on model accuracy is measured by the perplexity (PPL) metric, with results on the C4 dataset detailed in Table \ref{tab:ppl}, and results on the WikiText2 dataset in Table \ref{apdx-tab:ppl}.
As indicated in these tables, \name\ notably enhances the generative capabilities of quantized models across various LLMs and quantization methods, with AWQ+\name\  consistently achieving new state-of-the-art (SOTA) results.
These findings demonstrate the efficacy of \name\ in improving quantized LLMs and particularly highlight the effectiveness of utilizing the lossless KV cache from full-precision models. We provide more experiment results on LLaMA-3 and other heterogeneous LLMs (e.g. OPT) in Appendix~\ref{apdx-subsec:ppl}. \name\ significantly improves different quantized LLMs, especially for LLaMA-3 models with larger quantization error. These results further prove the compatibility of our \name\ with various LLM backbones.

\paragraph{Results on MMLU Tasks.}
\begin{table}[t]
\centering
\resizebox{1\linewidth}{!}{
\begin{tabular}{l|cc}
\toprule
\textbf{Method}    & \textbf{Vicuna-v1.5-7B}   & \textbf{Vicuna-v1.5-13B} \\ \midrule
FP16                  & 5.31                      & 5.52            \\ \midrule
RTN                & 4.34                      & 5.13            \\
                           \rowcolor{orange!10}  
                           \ \ +\nameprompt      &  4.72         &  5.27       \\
                           \rowcolor{orange!10}     
                           \ \ +\nameprompt+Cal   &  \textbf{4.73}        &  \textbf{5.30}  \\
                           OmniQuant          & 4.78                      & 5.05            \\
                           \cellcolor{orange!10}\ \ +\nameprompt      &  \cellcolor{orange!10}\textbf{4.94}                 &   \cellcolor{orange!10}5.10          \\
                           \cellcolor{orange!10}\ \ +\nameprompt+Cal   &  \cellcolor{orange!10}4.85        &   \cellcolor{orange!10}\textbf{5.24} \\
                           AWQ                & 4.74                      & 5.17            \\
                           \rowcolor{orange!10} 
                           \ \ +\nameprompt      & 4.68                      & 5.34            \\
                           \rowcolor{orange!10}     
                           \ \ +\nameprompt+Cal   & \textbf{4.84}             & \textbf{5.44}   \\
\bottomrule
\end{tabular}
}
\caption{GPT-4 evaluation of INT3-group128 weight-only quantized Vicuna-v1.5 models on MT-Bench. The scores are on a scale of 10.}
\label{tab:mt_bench_3bit} 
\vspace{-0.5em}
\end{table}
\begin{figure*}[t]
\centering
\subfloat[(a) Vicuna-v1.3-7B]{
        \includegraphics[width=0.23\linewidth]{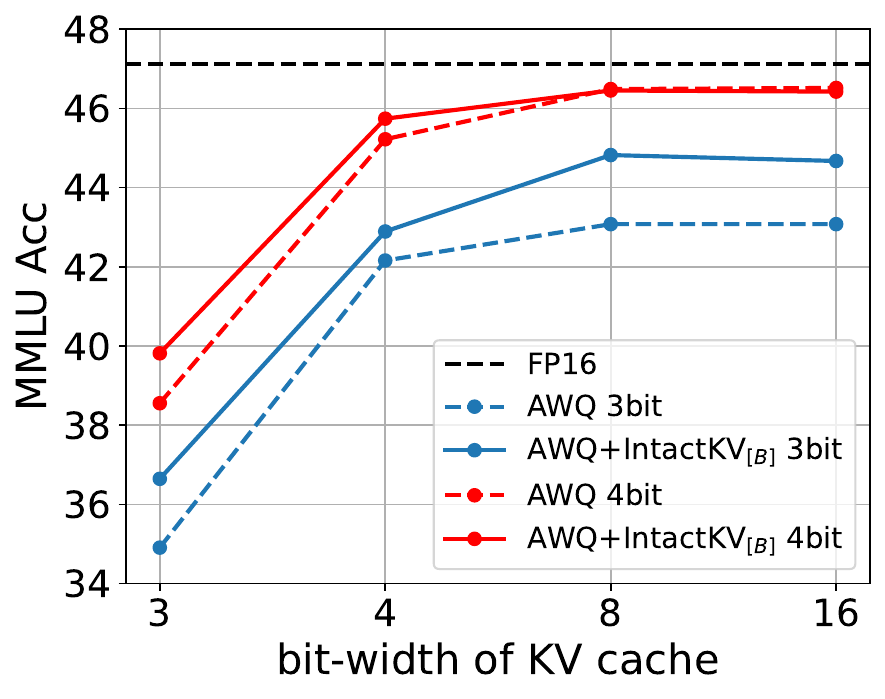}
    }
\subfloat[(b) Vicuna-v1.3-13B]{
        \includegraphics[width=0.23\linewidth]{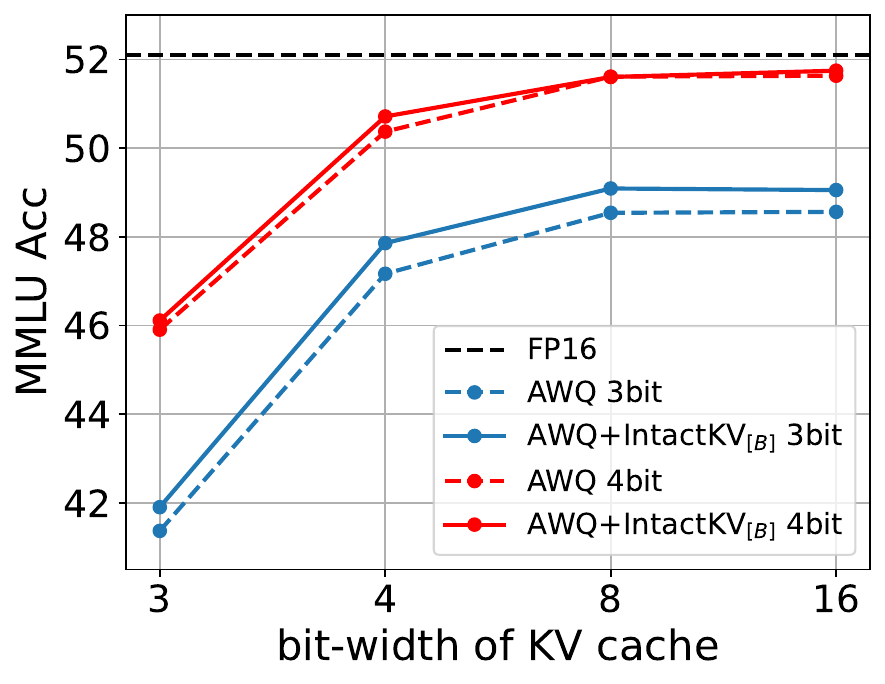}
    }
\subfloat[(c) Vicuna-v1.5-7B]{
        \includegraphics[width=0.24\linewidth]{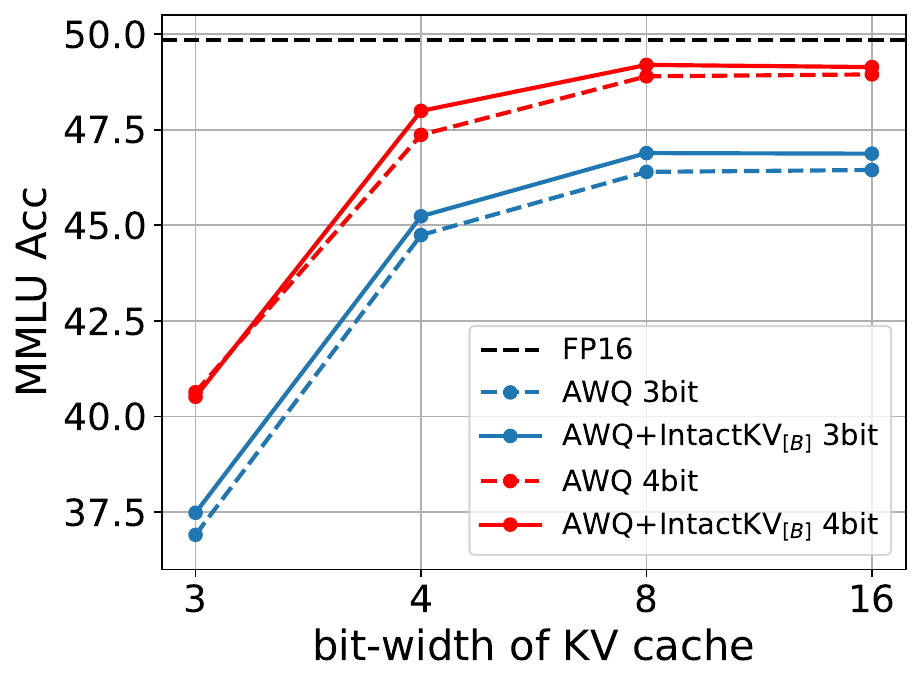}
    }
\subfloat[(d) Vicuna-v1.5-13B]{
        \includegraphics[width=0.23\linewidth]{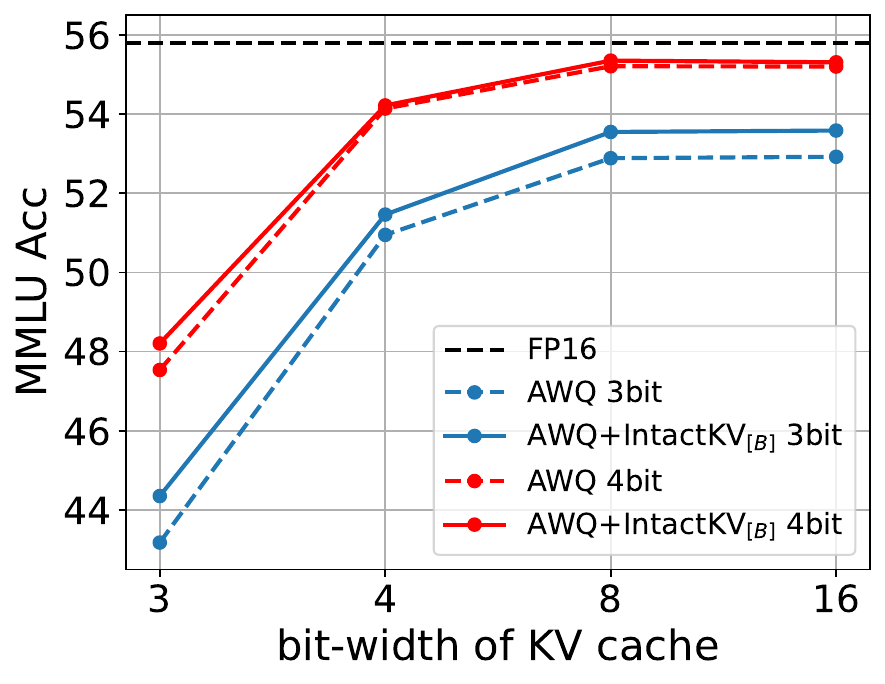}
    }
\vspace{-1ex}
\caption{Results of weight and KV cache quantization with different bit-widths on 5-shot MMLU benchmark. Note that this is additional to INT3/4 weight-only quantization. 
Blue and red lines indicate quantizing model weights to INT3 and INT4, respectively. We apply asymmetric per-head dynamic quantization to the KV cache.}
\label{fig:kv-quant}
\end{figure*}

\begin{table*}[!h]
\begin{center}
\resizebox{\linewidth}{!}{
\begin{tabular}{l|cc|cc|cc|cc|cc}
\toprule
\multicolumn{1}{c|}{\multirow{2}{*}{\textbf{Method}}} & \multicolumn{2}{c|}{\textbf{LLaMA-7B}} & \multicolumn{2}{c|}{\textbf{LLaMA-13B}} & \multicolumn{2}{c|}{\textbf{LLaMA-2-7B}} & \multicolumn{2}{c|}{\textbf{LLaMA-2-13B}}                       & \multicolumn{2}{c}{\textbf{LLaMA-3-8B}}                        \\  
\cmidrule{2-11} \multicolumn{1}{c|}{}                       & \textbf{C4}    & \textbf{WikiText2}           & \textbf{C4}    & \textbf{WikiText2} & \textbf{C4}    & \textbf{WikiText2} & \textbf{C4}    & \textbf{WikiText2} & \textbf{C4}    & \textbf{WikiText2}\\ \midrule
FP16 & 7.36 & 5.69 & 6.82 & 5.08 & 7.28 & 5.48 & 6.75 & 4.89 & 9.48 & 6.15           \\\midrule
OmniQuant & 17.03 & 12.17 & 15.65 & 11.16 & 21.40 & 14.74 & 16.24 & 12.28 & - & -           \\
\rowcolor{orange!10} \ \ +\namebos & 16.24 & 11.32 & 13.87 & 10.04 & 20.01 & 13.70 & 15.91 & 11.00 & - & - \\
QuaRot & 8.23 & 6.29 & 7.40 & 5.55 & 8.30 & 6.11 & 7.51 & 5.39 & 13.42 & 8.21           \\
\rowcolor{orange!10} \ \ +\namebos &\textbf{8.05}& \textbf{6.15} & \textbf{7.32} & \textbf{5.45} &\textbf{8.12}& \textbf{5.97} & \textbf{7.25} & \textbf{5.21} & \textbf{12.23} & \textbf{7.54}\\
\bottomrule
\end{tabular}
}
\end{center}
\vspace{-10pt}
\caption{INT4 weight and activation quantization results of LLaMA models on C4 and WikiText2 datasets.
}
\label{tab:ppl_wa}
\end{table*}
For SFT models, we implement \name\ on the quantized Vicuna models and evaluate the multi-task problem-solving ability on the MMLU benchmark. Table \ref{tab:mmlu} presents the detailed zero-shot and five-shot results for Vicuna-v1.3-13B.
The results demonstrate that \name\ significantly enhances the performance of quantized models across all categories of tasks and various quantization methods for Vicuna-v1.3-13B. 
Moreover, performance of Vicuna family under the five-shot setting is outlined in Table \ref{tab:all-models}.
Remarkably, \name\ achieves an average improvement of 1.05\% over OmniQuant and 0.8\% over AWQ across five model sizes, with AWQ+\name\ exhibiting superior performance over all the other quantized models. More results on MMLU are provided in Appendix~\ref{apdx-subsec:mmlu}.

\paragraph{Results on Commonsense QA Tasks.}
We further evaluate the quantized Vicuna models on zero-shot commonsense QA tasks.
The results of Vicuna-v1.3-13B, as detailed in Table \ref{tab:qa}, indicate that \name\ enables significant improvements over various quantization methods. For example, AWQ+\name\ surpasses the average accuracy of AWQ by 0.46\% under INT3-g128 quantization.
Additionally, Table \ref{tab:all-models} presents the average accuracy for various sizes of Vicuna models. In these evaluations, our \name\ leads to an average accuracy improvement of 0.45\% across different LLMs and quantization methods, which strongly demonstrates the efficacy of our proposed \name. More results on commonsense QA tasks can be found in Appendix~\ref{apdx-subsec:qa}.


\paragraph{Results on MT-Bench.}
To evaluate the quantized models' generation capabilities in multi-turn conversations and their alignment with human preferences, we use GPT-4 to score the responses of quantized models on MT-Bench.
We also calibrate \name, denoted as \name+Cal.
From Table \ref{tab:mt_bench_3bit}, \name\ significantly boosts the quantized model and \name+Cal further enhances generation quality by compensating for the quantization error.
For example, the 3-bit Vicuna-v1.5-13B quantized by AWQ has been improved from 5.17 to 5.34 by using the \name, which can be further boosted to 5.44 with trainable \name.
We provide INT4 quantization results in Table~\ref{apdx-tab:mt_bench_4bit}. Remarkably, with trainable \name, AWQ+\name\ even matches the full-precision model under INT4 quantization, while all other methods clearly lag behind the full-precision model.
These results demonstrate the effectiveness of \name\ as well as treating \name\ as trainable parameters.
Notably, the training process for the 7B model takes only 10 minutes on a single computing device, which is quite lightweight. In Appendix~\ref{apdx-sec:gbias}, we further demonstrate the effectiveness of calibrating \name\ by comparing it with group bias tuning, a commonly used fine-tuning strategy for quantized models. \name\ calibration can achieve better or comparable results with group bias tuning while using significantly fewer trainable parameters. Besides, \name\ calibration serves as a more versatile calibration strategy for quantized models, which is suitable for various quantization settings.

\subsection{Extension to KV Cache Quantization}
\label{sec:kvquant-exp}
\name\ can be readily applied to KV cache quantization to further decrease memory requirements.
We employ a mixed-precision strategy that keeps \name\ in FP16 while the rest of the KV cache is quantized to lower bits. This only induces negligible memory overhead since \name\ only contains the KV cache of the first few tokens. Note that this does not bring any additional inference costs since in the workflow of KV cache quantization, all quantized KV cache needs to be de-quantized back to FP16 before the matrix multiplication. Keeping \name\ in FP16 reduces the overhead of de-quantization, i,e., we only need to cheaply concatenate the FP16 \name\ with the rest de-quantized KV cache together.
From Figure~\ref{fig:kv-quant}, \name\ notably improves AWQ across different models and KV cache bit widths under the INT3 weight quantization. For INT4 weight quantization, AWQ+\name\ still gains an average accuracy increase of 0.27\% over the original quantized model. We also notice that quantizing the KV cache to INT8 leads to almost no performance drop on the MMLU benchmark. When equipped with \name, INT8 KV cache can even surpass vanilla AWQ-quantized models with FP16 KV cache, especially under INT3 weight quantization.

\subsection{Extension to Activation Quantization}
\label{sec:actquant-exp}
In Table~\ref{tab:ppl_wa}, we provide experiment results of combining \name\ with OmniQuant~\cite{shao2023omniquant} and QuaRot~\cite{ashkboos2024quarot} for weight and activation quantization. The implementation details can be found in Appendix~\ref{apdx-sec:quant-method-details}. To avoid extra inference costs, we need to quantize the whole KV cache to lower bits and can not keep the KV cache of pivot tokens intact. However, as detailed in Appendix~\ref{apdx-sec:wa}, we find that \name\ has a significantly smoother distribution compared with the rest of the KV cache. Therefore, the full-precision \name\ can be readily quantized to lower bits with negligible accuracy loss, thus rendering \name\ amenable to weight and activation quantization with no extra inference costs. As shown in Table~\ref{tab:ppl_wa}, our \name\ significantly surpasses the performance of original quantized models for two different quantization methods, improving the PPL by 1.07 for OmniQuant and 0.31 for QuaRot on average. When combined with QuaRot, our \name\ archives new state-of-the-art (SOTA) results on INT4 weight and activation quantization.

\section{Conclusions}
In this paper, we propose \name, a simple and easy-to-combine method to improve large language model quantization.
The research is motivated by the previously overlooked outliers over pivot tokens, which lead to attention sinks that are critical to the performance of quantized LLMs.
By generating \name\ with the full-precision model, the quantization error accumulated over the attention scores can be effectively alleviated. \name\ can also be calibrated as additional parameters to the LLM backbone, further improving the quantized LLMs.
Experiments show that combining the proposed \name\ gives consistent improvement on various sizes of LLMs and across multiple downstream tasks, leading to new state-of-the-art results for large language model quantization. 

\section{Limitations}
\label{sec:lim}
More experiments may be needed for LLM evaluation. LLMs are being applied to a wide range of tasks, posing high demands on various model abilities.
When quantizing LLMs to low bits, these abilities may be affected to varying degrees. Therefore, a comprehensive evaluation is required to gauge the capabilities of quantized LLMs. Although we experiment on several downstream tasks, such as PPL, MMLU, commonsense QA, and MT-bench, we note that this may not be enough to assess all abilities of LLMs. For example, how long context affects quantized models still remains unknown.

\section{Ethics Statement}
The development of LLM quantization techniques can further democratize LLMs, lowering the costs of LLM serving and enabling more people to get access to advanced AI assistants. Nonetheless, LLM itself may inherit certain social biases from training data concerning gender, race, etc. Quantization can not mitigate such biases. Therefore, caution must be taken when using quantized LLMs.



\clearpage

\appendix
\label{appendix:proof}
\begin{table*}[h!]
\begin{center}
\resizebox{\linewidth}{!}{
\begin{tabular}{l|ccccccc}
\toprule
\textbf{Method} & \textbf{LLaMA-7B} & \textbf{LLaMA-13B} & \textbf{LLaMA-30B} & \textbf{LLaMA-65B} & \textbf{LLaMA-2-7B} & \textbf{LLaMA-2-13B} & \textbf{LLaMA-2-70B} \\ \midrule
FP16            & 5.69               & 5.08                & 4.09                & 3.52                & 5.48               & 4.89                & 3.33                \\ \midrule
RTN             & 6.98               & 5.88                & 4.84                & 4.22                & 6.65               & 5.52                & 3.99                \\
\rowcolor{orange!10} 
\ \ +\namebos     & 6.52               & 5.70                & 4.69       & 4.05       & 6.40      & 5.44       & 3.84       \\ 
GPTQ            & 6.62               & 5.68                & 4.75                & 4.20                & 7.29               & 5.52                & 4.02                \\
\rowcolor{orange!10} 
\ \ +\namebos     & 6.51               & 5.62                & 4.63       & 4.12       & 7.00      & 5.46       & 3.97       \\ 
OmniQuant       & 6.20               & \textbf{5.46}                & 4.59                & 3.95                & \textbf{6.10}               & 5.32                & 3.81                \\
\rowcolor{orange!10} 
\ \ +\namebos     & \textbf{6.18 }              & \textbf{5.46}                & 4.58       & 3.95       & \textbf{6.10}      & 5.31       & 3.80       \\ 
AWQ             & 6.34               & 5.53                & 4.60                & 3.95                & 6.25               & 5.32                & 3.75                \\
\rowcolor{orange!10} 
\ \ +\namebos     & 6.23               & 5.49                & \textbf{4.54}       & \textbf{3.89}       & 6.14      & \textbf{5.29}       & \textbf{3.72}       \\ 
\bottomrule
\end{tabular}
}
\end{center}
\vspace{-10pt}
\caption{INT3-group128 weight-only quantization results of LLaMA and LLaMA-2 models on WikiText2 dataset.}
\label{apdx-tab:ppl}
\end{table*}
\begin{table}[!h]
\begin{center}
\resizebox{\linewidth}{!}{
\begin{tabular}{l|cc|cc}
\toprule
\multicolumn{1}{c|}{\multirow{2}{*}{\textbf{Method}}} & \multicolumn{2}{c|}{\textbf{LLaMA-3-8B}}                       & \multicolumn{2}{c}{\textbf{LLaMA-3-70B}}                        \\  
\cmidrule{2-5} \multicolumn{1}{c|}{}                       & \textbf{C4}    & \textbf{WikiText2}           & \textbf{C4}    & \textbf{WikiText2}\\ \midrule
FP16                                         & 9.48 & 6.15           & 7.20 & 2.87           \\\midrule
RTN                                          & 18.96 & 12.05          & 18.65 & 8.01           \\
\rowcolor{orange!10} \ \ +\namebos           &16.89 & 10.77         & 14.11 & 5.43 \\
GPTQ                                         & 51.69 & 26.14           & 5.1E4 & 5.1E4           \\
\rowcolor{orange!10} \ \ +\namebos           &13.08 & 8.32         & 3.5E4 & 4.5E4 \\
OmniQuant                                    & 14.46 & 9.09           & 9.04 & 5.29           \\
\rowcolor{orange!10} \ \ +\namebos           &13.99 & 8.88         & 8.83 & 5.02 \\
AWQ                                          & 12.69 & 8.15          & 8.55 & 4.66           \\
\rowcolor{orange!10} \ \ +\namebos           &\textbf{12.42}& \textbf{7.97}         & \textbf{8.35} & \textbf{4.41} \\
\bottomrule
\end{tabular}
}
\end{center}
\vspace{-10pt}
\caption{INT3-group128 weight-only quantization results of LLaMA-3 on C4 and WikiText2 datasets.
}
\label{apdx-tab:ppl_llama3}
\end{table}

\section{Proof of Theorem 1}



\begin{proof}
Denote the output of the softmax function as the score $\bs$, i.e., $\bs = \sm(\frac{\bq\bK^\top}{\sqrt{d}})$, and also define the error output from the softmax function as $\Delta\bs$.
To show the error of the attention head, we first justify how the error propagates from the score to the attention head. 
\begin{align*}
    &\norm {\Delta\bh}_2  = \left\|[(\bs + \Delta\bs)(\bV+\Delta\bV)-\bs\bV]\bW^O \right\| _2 \\
    &\le \left(  \norm {\Delta\bs}_2\norm{\bV+\Delta\bV}_2 + \norm {\bs}_2 \norm{\Delta\bV}_2  \right) \norm{\bW^O}_2 \\
    &\le \big( \norm{\Delta\bs}_2 (\norm{\bV}_2 \!+ \!\norm{\Delta\bV}_F ) + \norm{\Delta\bV}_F \big) \norm{\bW^O}_2 ,
    \end{align*}
where the inequalities are because
\begin{align*}
\norm{\bx+\by}_2 \le\norm{\bx}_2\!+\!\norm{\by}_2, ~~
\norm {\bs \bV}_2  \le \norm\bs_2 \norm\bV_2,
\end{align*}
and $\norm\bs_2 \le \norm\bs_1 =1, \norm\bV_2 \le \norm\bV_F$.

 Next, we characterize the error of score $\norm{\Delta\bs}_2$.
This is not easy as the error propagates through the softmax function. To proceed, we need the relative condition number of the softmax function.
As indicated in \cite{blanchard2021accurately},
\begin{equation*}
    \!\!\frac{ \norm{\sm(\bx\!+\!\Delta\bx) \!-\!\sm(\bx)}_\infty }{\norm{\sm(\bx)}_\infty} 
    \!\!\le \!
    \kappa(\bx) \frac{\norm{\Delta\bx}_\infty}{\norm\bx_\infty}\!,
\end{equation*}
where $\kappa(\bx)=n\norm\bx_\infty$ ($\bx\in\Rbb^n$) is an upper bound of the relative condition number of the softmax function.
Let $\bx = \bq\bK^\top/\sqrt{d}$ and $\Delta\bx = \bq\Delta\bK^\top/\sqrt{d}$, we have
\begin{align*}
    \frac{\norm{\Delta\bs}_\infty}{\norm\bs_\infty} \le n \norm{\Delta\bx}_\infty \le \frac{n}{\sqrt{d}}\norm\bq_2 \normtwoinf{\Delta\bK}.
\end{align*}
Considering that the output of the softmax function is a probability, we have $\norminf\bs\le 1$.
Therefore, we obtain
\[
\norm{\Delta\bs}_2 \le \sqrt{n}\norminf{\Delta\bs} 
\le \frac{n^{2/3}}{\sqrt{d}}\norm\bq_2 \normtwoinf{\Delta\bK}.
\]
Combining the above ingredients, we derive the main results of the Theorem \ref{theo:error analysis}. 

\end{proof}

\section{System Prompt of Vicuna Models}
\label{apdx-sec:prompt}
\begin{figure}[H]
\centering
\includegraphics[width=\linewidth]{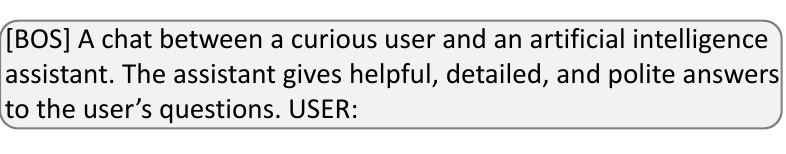}
\caption{System Prompt of Vicuna Models.}
\label{fig:prompt}
\end{figure}

\section{Visualization of Activations and Attention Map}
\label{apdx-sec:viss}
\subsection{Implementation Details}
\label{apdx-subsec:vis_details}
We use ShareGPT dataset for our visualizations, where each sample starts with Vicuna system prompt of length 34. We use a randomly sampled sequence of length 128 to visualize the output activations and plot the corresponding attention map of the first 64 tokens for better visualization. The attention score is mean pooled over different heads.

\subsection{Visualization of LLaMA Models}
\begin{figure*}[t]
\centering
\subfloat[(a) Output activations of LLaMA-7B Layer 0]{
        \includegraphics[width=0.24\linewidth]{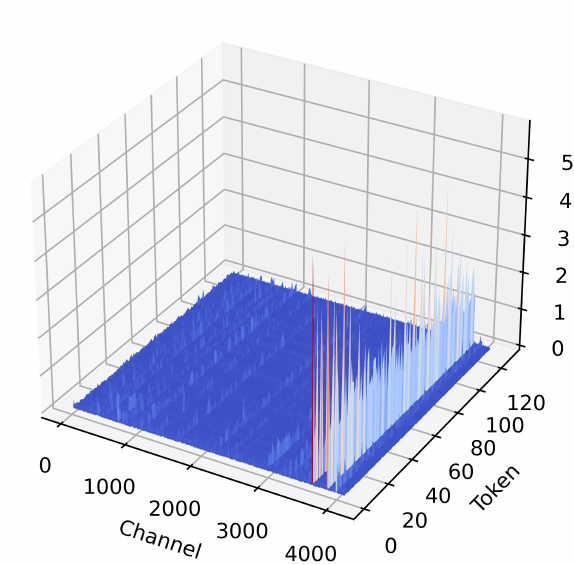}
    }
\subfloat[(b) Output activations of LLaMA-7B Layer 8]{
        \includegraphics[width=0.24\linewidth]{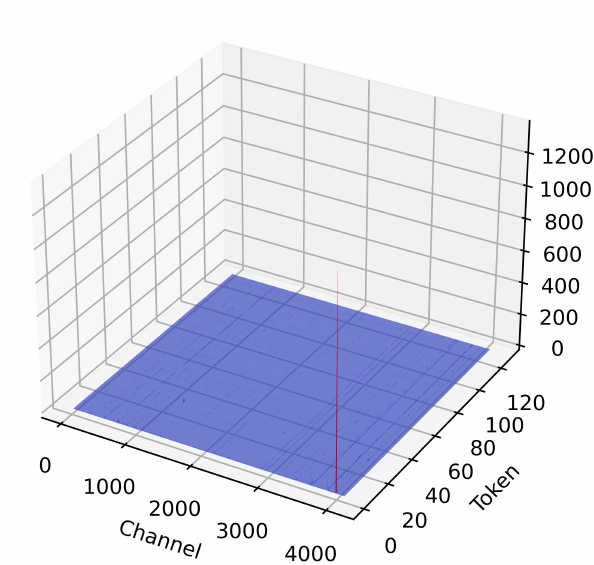}
    }
\subfloat[(c) Output activations of LLaMA-7B Layer 16]{
        \includegraphics[width=0.24\linewidth]{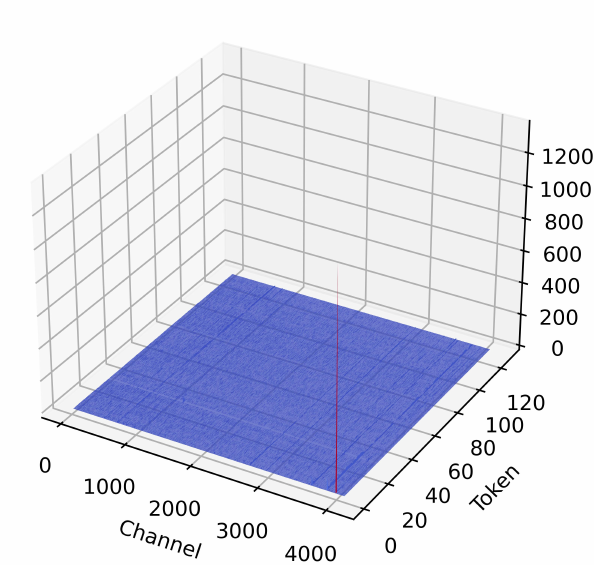}
    }
\subfloat[(d) Output activations of LLaMA-7B Layer 24]{
        \includegraphics[width=0.24\linewidth]{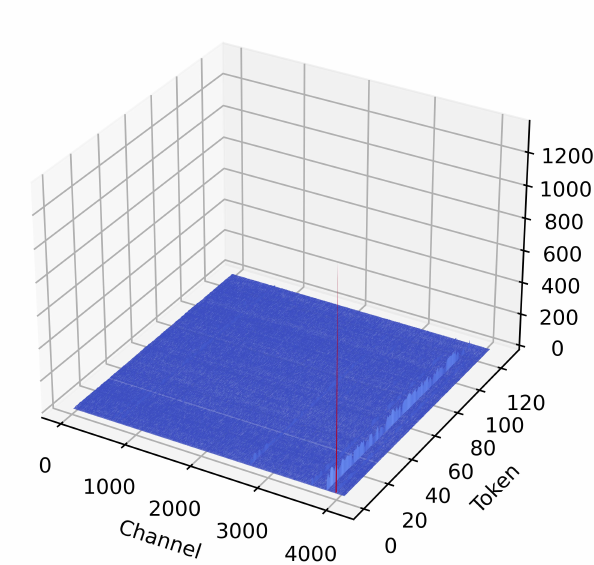}
    }
\vspace{0.25cm}
\subfloat[(e) Attention map of \\LLaMA-7B Layer 0]{
        \includegraphics[width=0.24\linewidth]{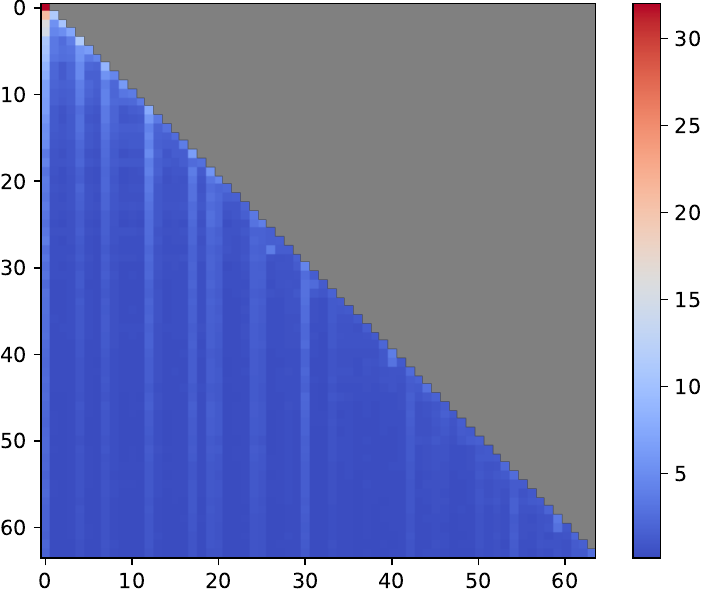}
    }
\subfloat[(f) Attention map of \\LLaMA-7B Layer 8]{
        \includegraphics[width=0.24\linewidth]{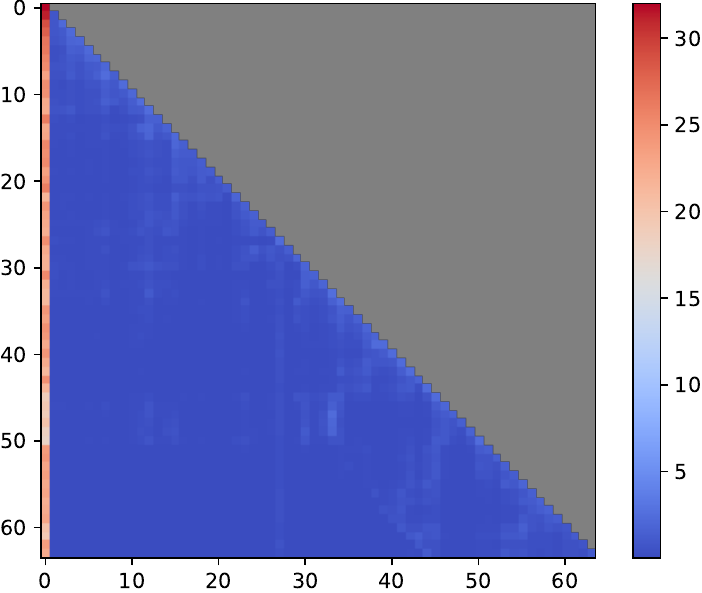}
    }
\subfloat[(g) Attention map of \\LLaMA-7B Layer 16]{
        \includegraphics[width=0.24\linewidth]{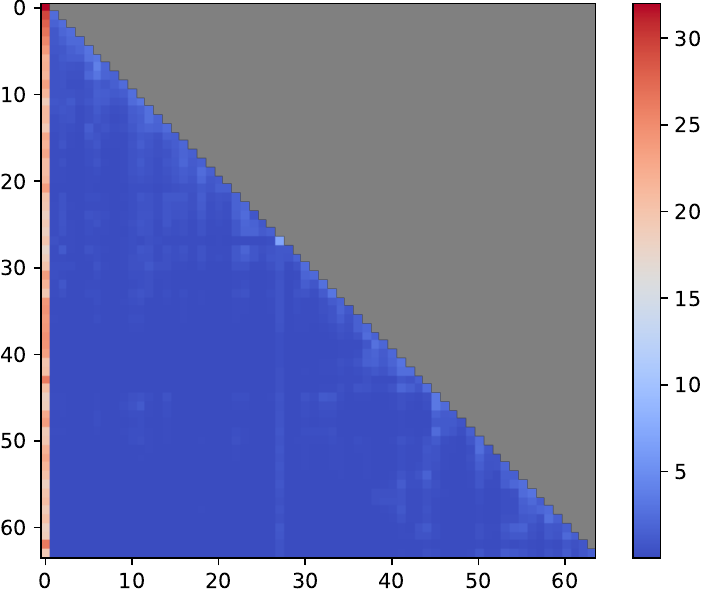}
    }
\subfloat[(h) Attention map of \\LLaMA-7B Layer 24]{
        \includegraphics[width=0.24\linewidth]{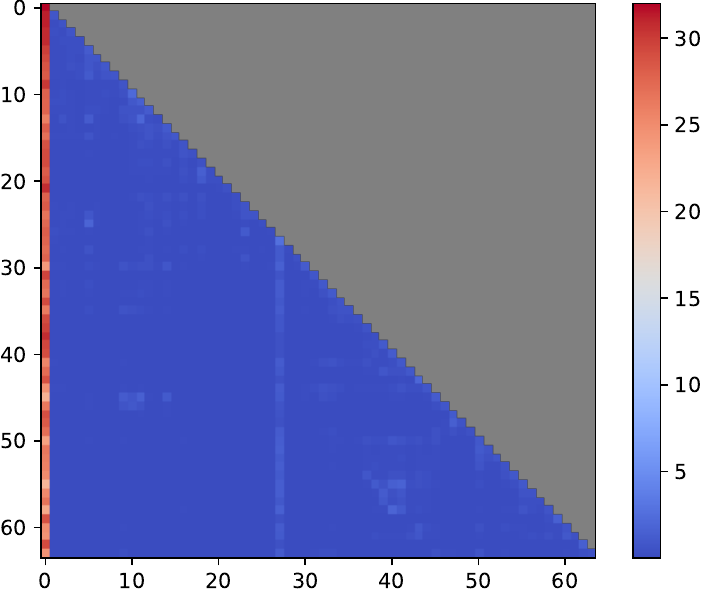}
    }
	\caption{Magnitude of the output activations and attention map in LLaMA-7B.
 }
\label{apdx-fig:llama-7b}
\end{figure*}

\begin{figure*}[t]
\centering
\subfloat[(a) Output activations of LLaMA-2-70B Layer 16]{
        \includegraphics[width=0.24\linewidth]{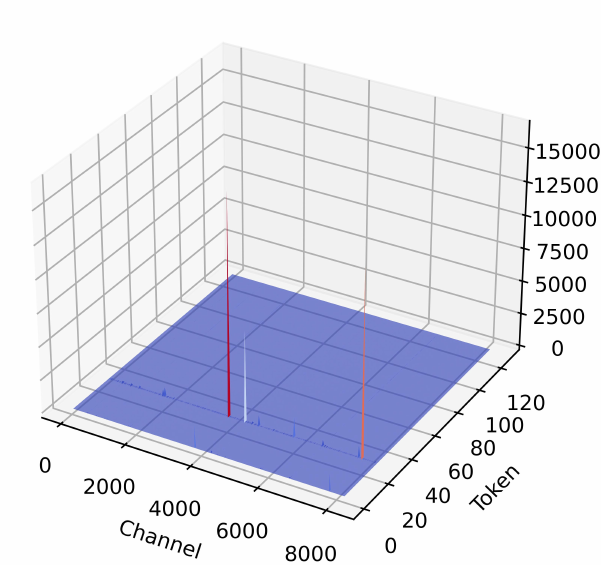}
    }
\subfloat[(b) Output activations of LLaMA-2-70B Layer 32]{
        \includegraphics[width=0.24\linewidth]{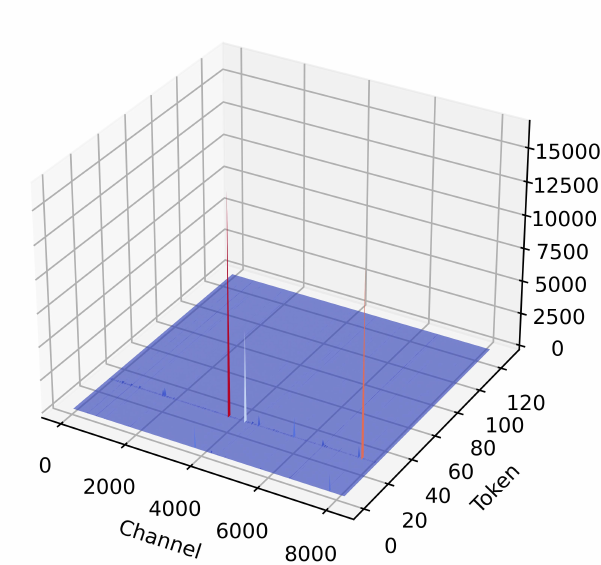}
    }
\subfloat[(c) Output activations of LLaMA-2-70B Layer 48]{
        \includegraphics[width=0.24\linewidth]{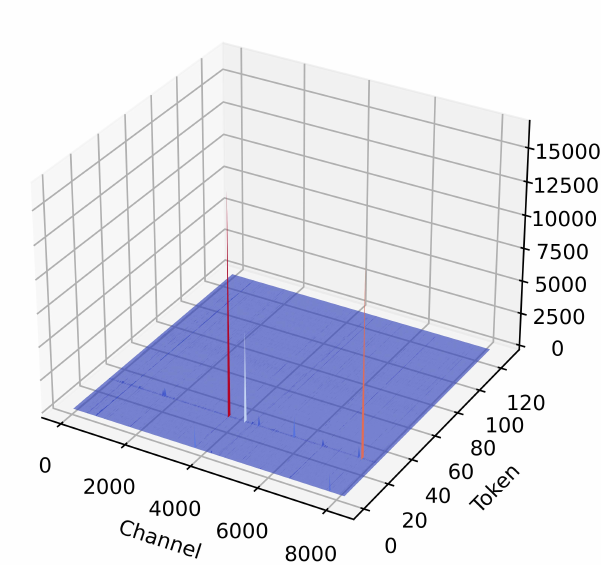}
    }
\subfloat[(d) Output activations of LLaMA-2-70B Layer 64]{
        \includegraphics[width=0.24\linewidth]{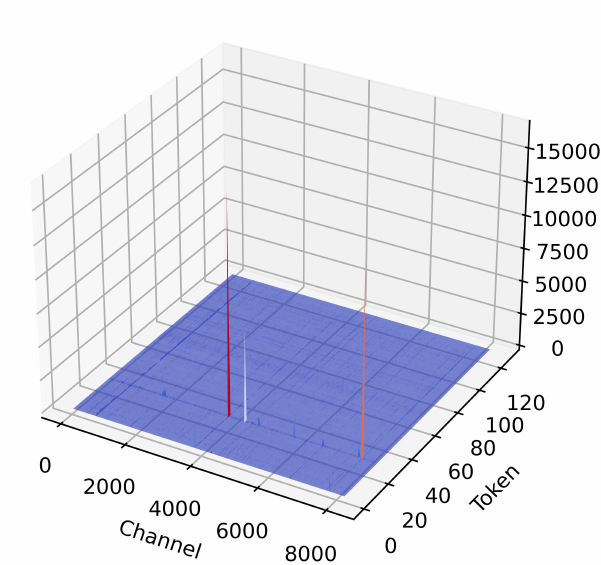}
    }
\vspace{0.25cm}
\subfloat[(e) Attention map of \\LLaMA-2-70B Layer 16]{
        \includegraphics[width=0.24\linewidth]{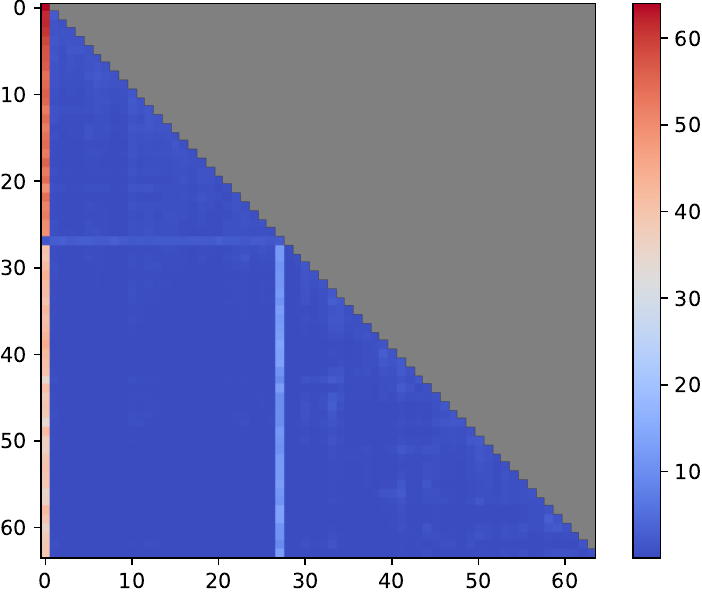}
    }
\subfloat[(f) Attention map of \\LLaMA-2-70B Layer 32]{
        \includegraphics[width=0.24\linewidth]{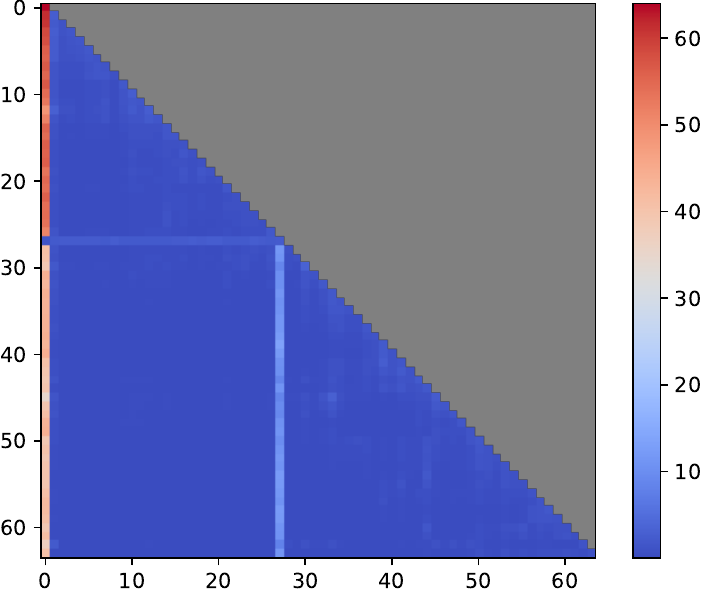}
    }
\subfloat[(g) Attention map of \\LLaMA-2-70B Layer 48]{
        \includegraphics[width=0.24\linewidth]{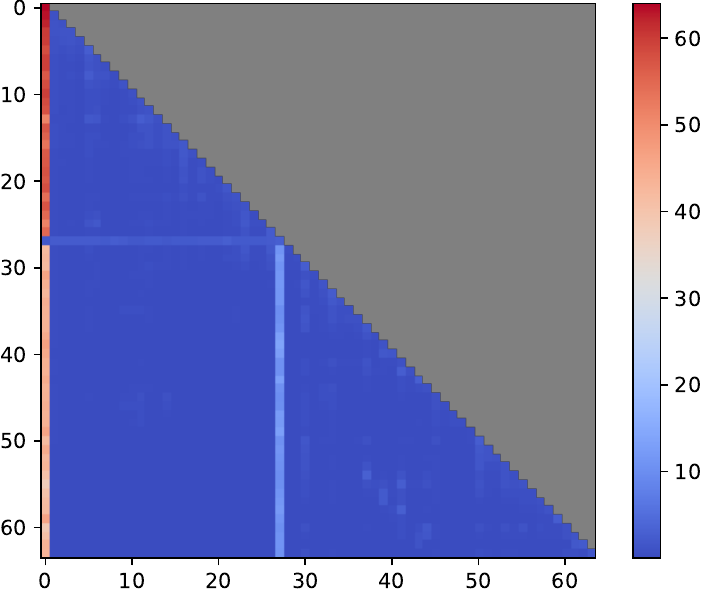}
    }
\subfloat[(h) Attention map of \\LLaMA-2-70B Layer 64]{
        \includegraphics[width=0.24\linewidth]{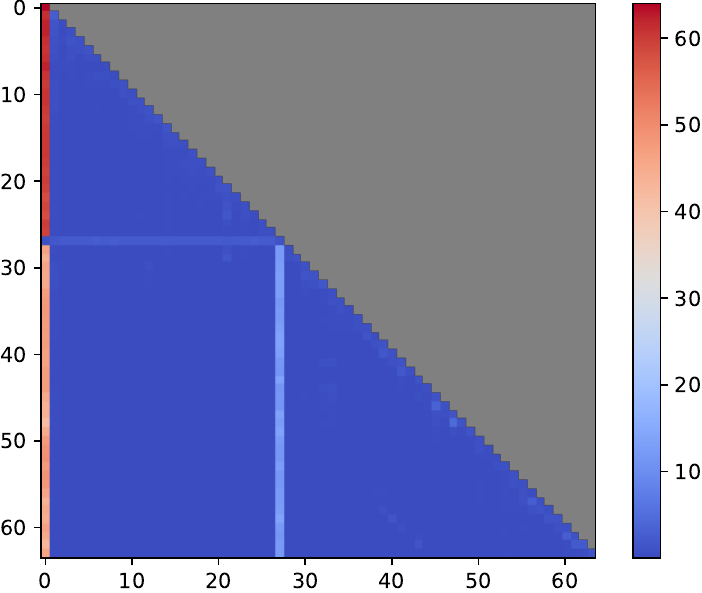}
    }
	\caption{Magnitude of the output activations and attention map in LLaMA-2-70B.
 }
\label{apdx-fig:llama-2-70b}
\end{figure*}

\begin{figure*}[t]
\centering
\subfloat[(a) Output activations of LLaMA-3-70B Layer 16]{
        \includegraphics[width=0.24\linewidth]{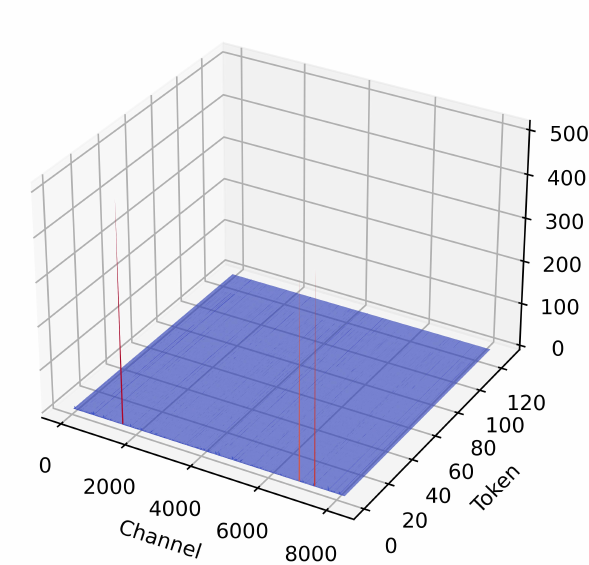}
    }
\subfloat[(b) Output activations of LLaMA-3-70B Layer 32]{
        \includegraphics[width=0.24\linewidth]{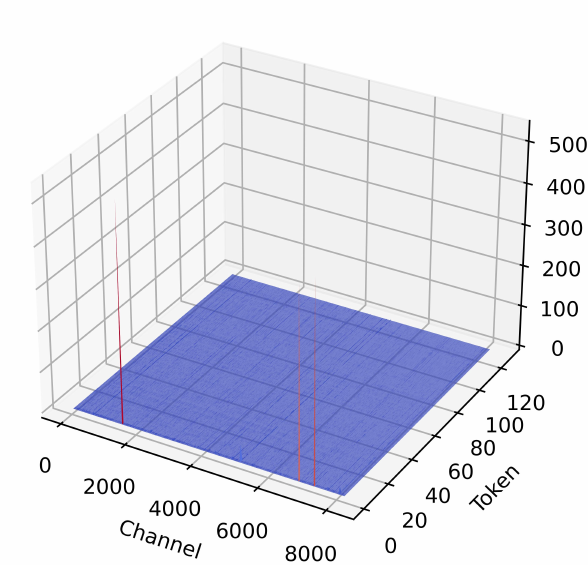}
    }
\subfloat[(c) Output activations of LLaMA-3-70B Layer 48]{
        \includegraphics[width=0.24\linewidth]{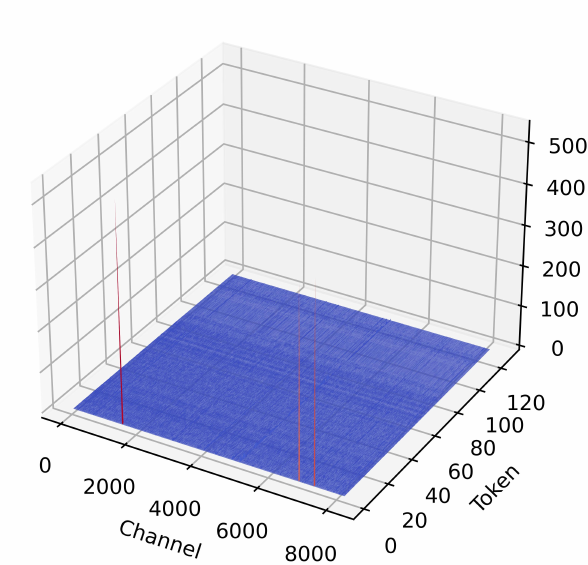}
    }
\subfloat[(d) Output activations of LLaMA-3-70B Layer 64]{
        \includegraphics[width=0.24\linewidth]{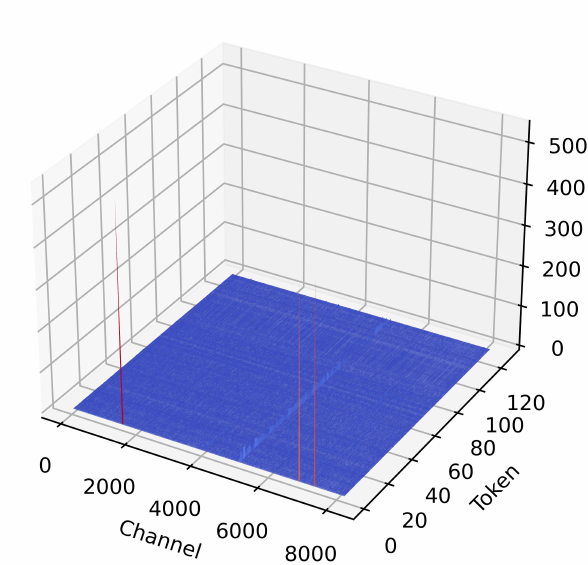}
    }
\vspace{0.25cm}
\subfloat[(e) Attention map of \\LLaMA-3-70B Layer 16]{
        \includegraphics[width=0.24\linewidth]{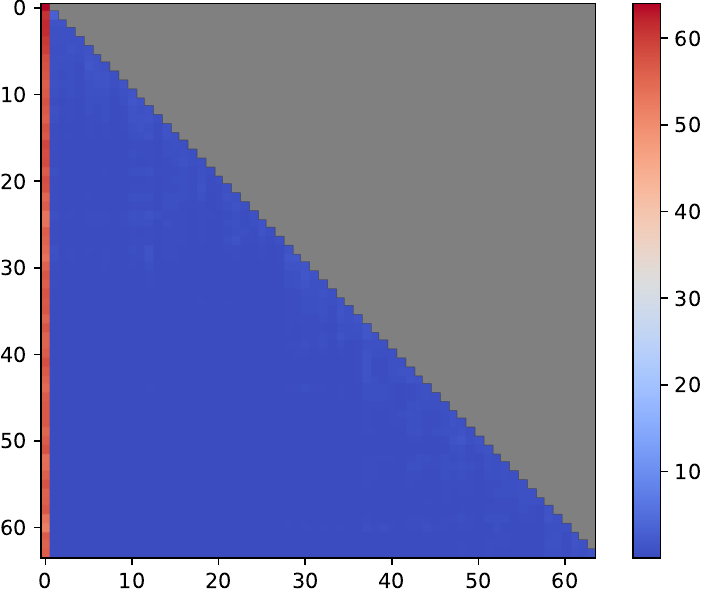}
    }
\subfloat[(f) Attention map of \\LLaMA-3-70B Layer 32]{
        \includegraphics[width=0.24\linewidth]{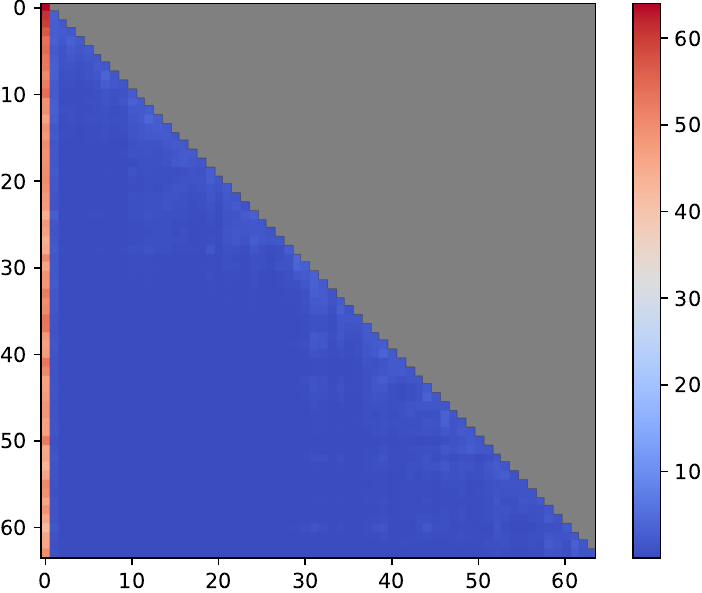}
    }
\subfloat[(g) Attention map of \\LLaMA-3-70B Layer 48]{
        \includegraphics[width=0.24\linewidth]{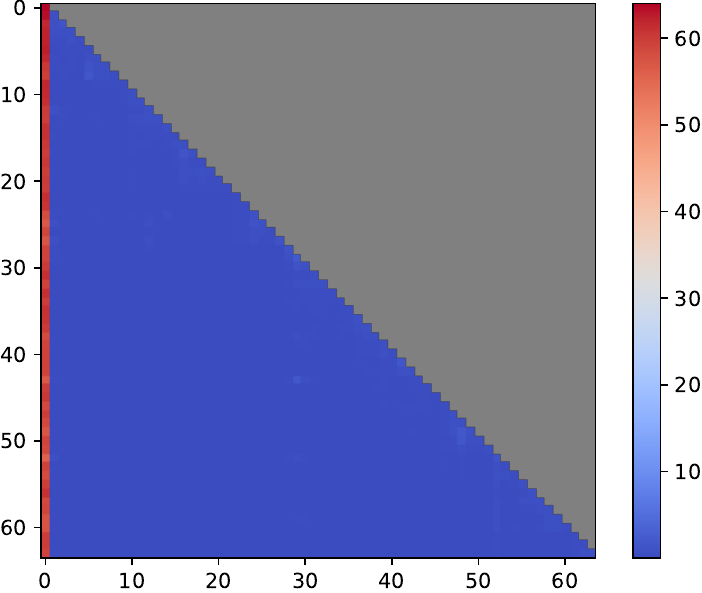}
    }
\subfloat[(h) Attention map of \\LLaMA-3-70B Layer 64]{
        \includegraphics[width=0.24\linewidth]{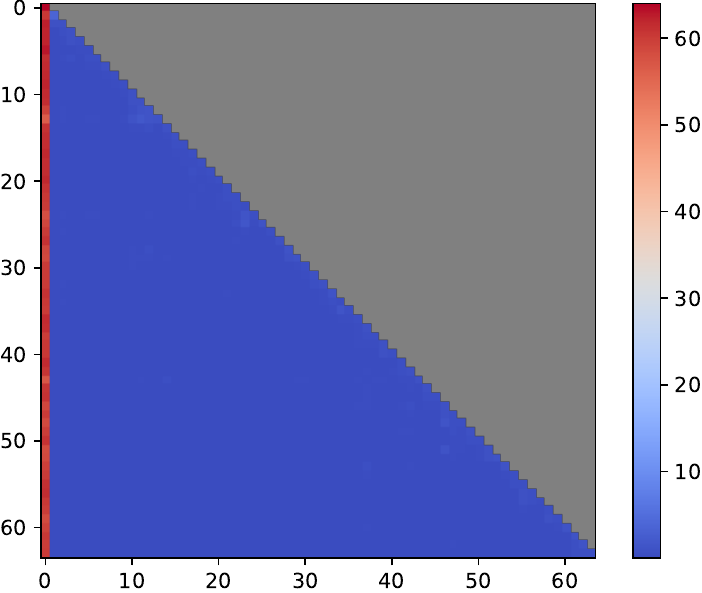}
    }
	\caption{Magnitude of the output activations and attention map in LLaMA-3-70B.
 }
\label{apdx-fig:llama-3-70b}
\end{figure*}

We provide more visualizations of the output activations and attention map of LLaMA models in Figure.~\ref{apdx-fig:llama-7b}--\ref{apdx-fig:llama-3-70b}. Similar to our observations in Section~\ref{sec:motivation}, we find that pivot tokens only appear at the very beginning of the input sequence, and \bos always serves as a pivot token.

\subsection{Visualization of Vicuna Models}
\label{apdx-subsec:vicuna_vis}
\begin{figure*}[t]
\centering
\subfloat[(a) Output activations of Vicuna-v1.3-7B Layer 0]{
        \includegraphics[width=0.24\linewidth]{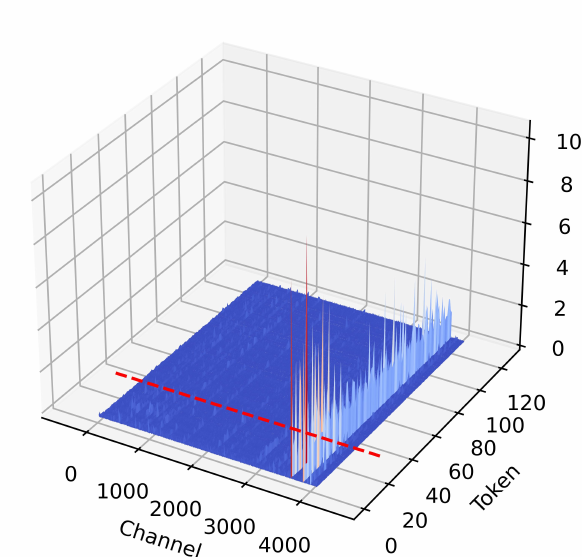}
    }
\subfloat[(b) Output activations of Vicuna-v1.3-7B Layer 8]{
        \includegraphics[width=0.24\linewidth]{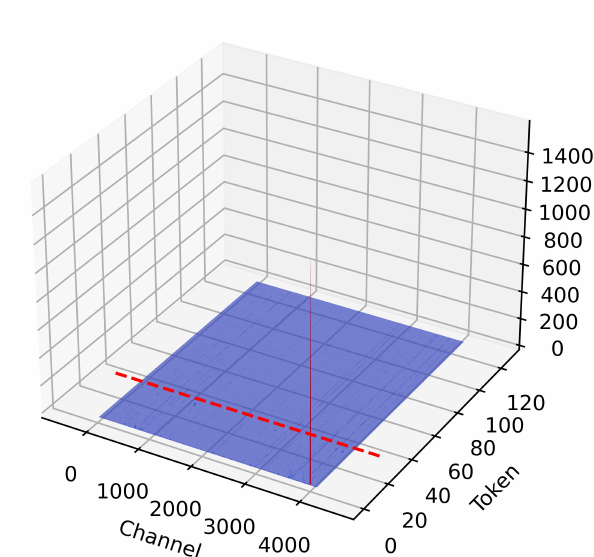}
    }
\subfloat[(c) Output activations of Vicuna-v1.3-7B Layer 16]{
        \includegraphics[width=0.24\linewidth]{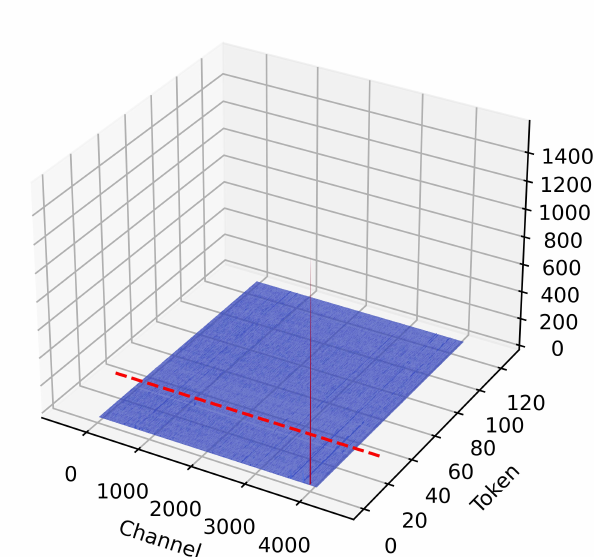}
    }
\subfloat[(d) Output activations of Vicuna-v1.3-7B Layer 24]{
        \includegraphics[width=0.24\linewidth]{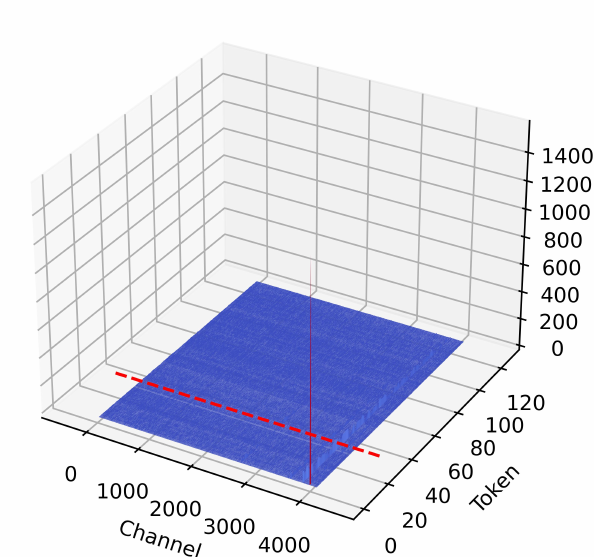}
    }
\vspace{0.25cm}
\subfloat[(e) Attention map of \\Vicuna-v1.3-7B Layer 0]{
        \includegraphics[width=0.24\linewidth]{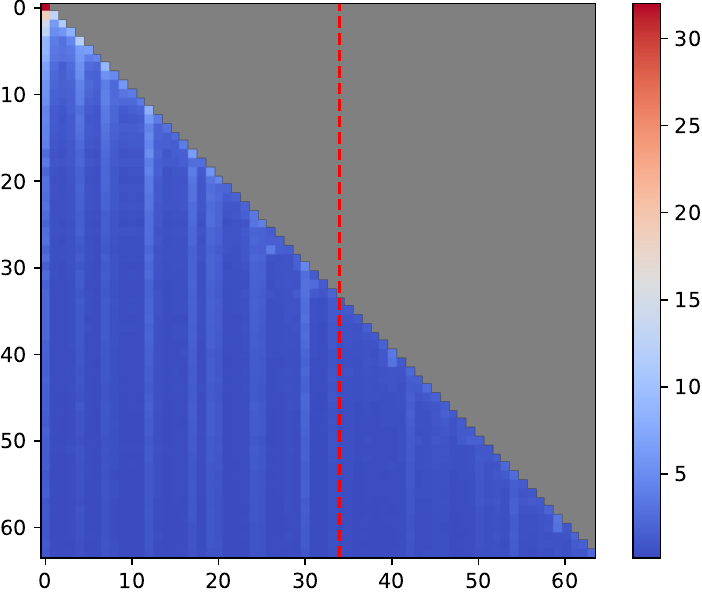}
    }
\subfloat[(f) Attention map of \\Vicuna-v1.3-7B Layer 8]{
        \includegraphics[width=0.24\linewidth]{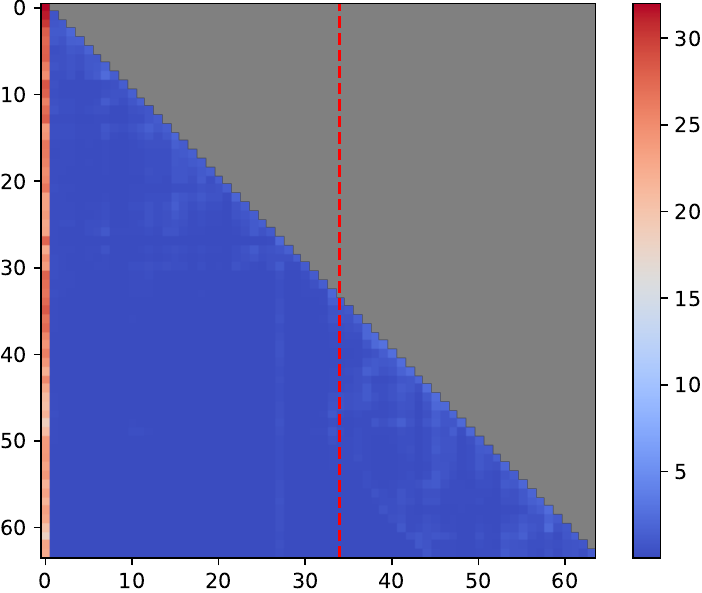}
    }
\subfloat[(g) Attention map of \\Vicuna-v1.3-7B Layer 16]{
        \includegraphics[width=0.24\linewidth]{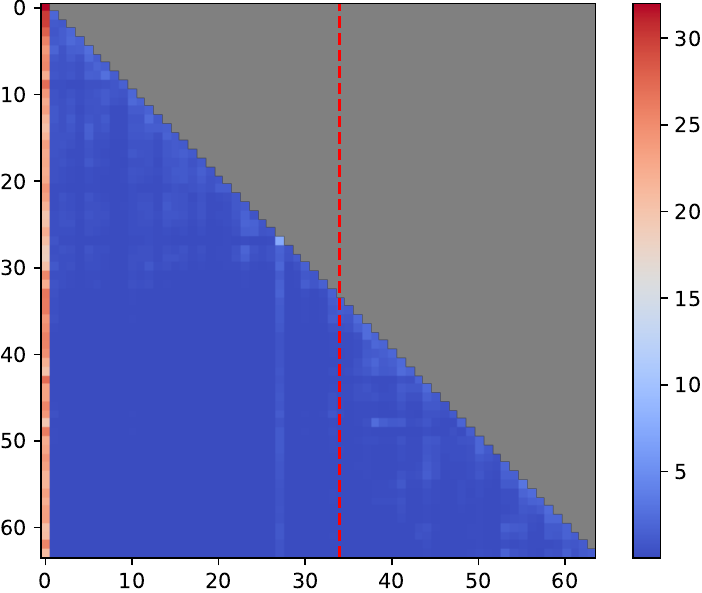}
    }
\subfloat[(h) Attention map of \\Vicuna-v1.3-7B Layer 24]{
        \includegraphics[width=0.24\linewidth]{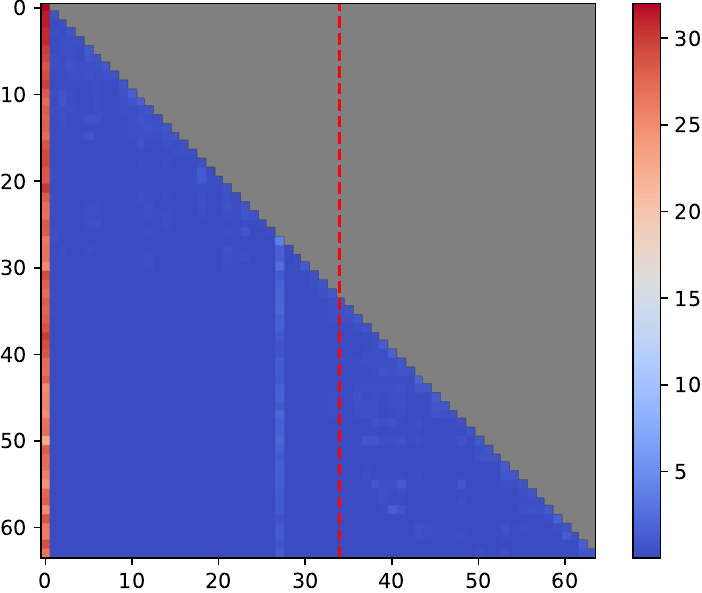}
    }
	\caption{Magnitude of the output activations and attention map in Vicuna-v1.3-7B. The tokens before the red dashed line correspond to the Vicuna system prompt.
 }
\label{apdx-fig:vicuna-v1.3-7b}
\end{figure*}

\begin{figure*}[t]
\centering
\subfloat[(a) Output activations of Vicuna-v1.5-13B Layer 8]{
        \includegraphics[width=0.24\linewidth]{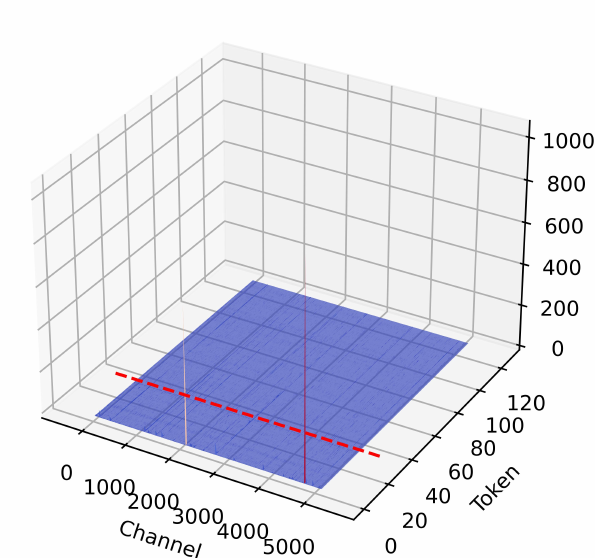}
    }
\subfloat[(b) Output activations of Vicuna-v1.5-13B Layer 16]{
        \includegraphics[width=0.24\linewidth]{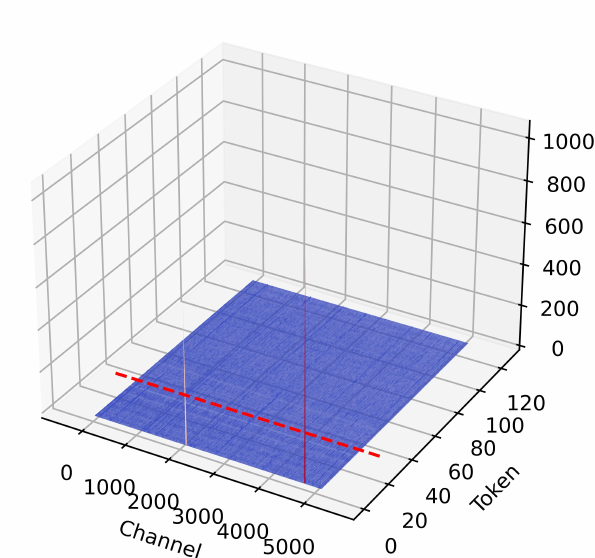}
    }
\subfloat[(c) Output activations of Vicuna-v1.5-13B Layer 24]{
        \includegraphics[width=0.24\linewidth]{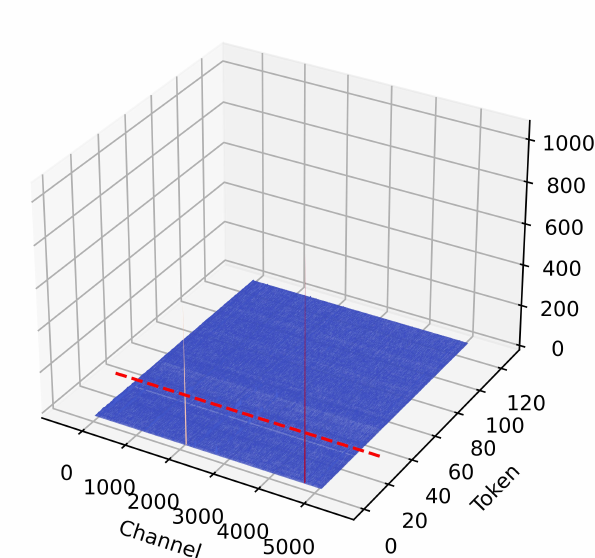}
    }
\subfloat[(d) Output activations of Vicuna-v1.5-13B Layer 32]{
        \includegraphics[width=0.24\linewidth]{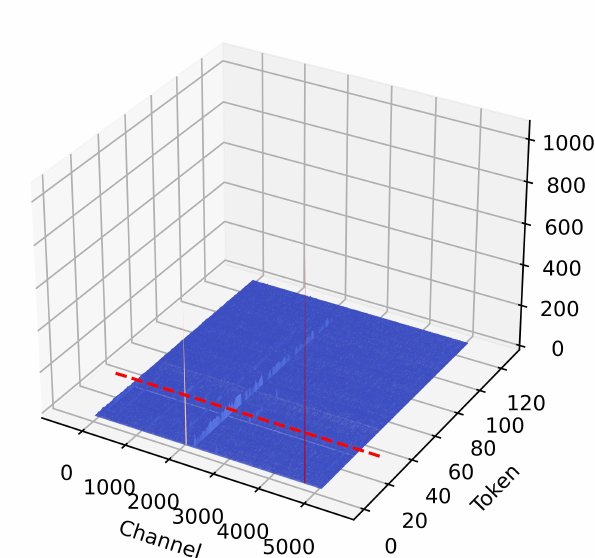}
    }
\vspace{0.25cm}
\subfloat[(e) Attention map of \\Vicuna-v1.5-13B Layer 8]{
        \includegraphics[width=0.24\linewidth]{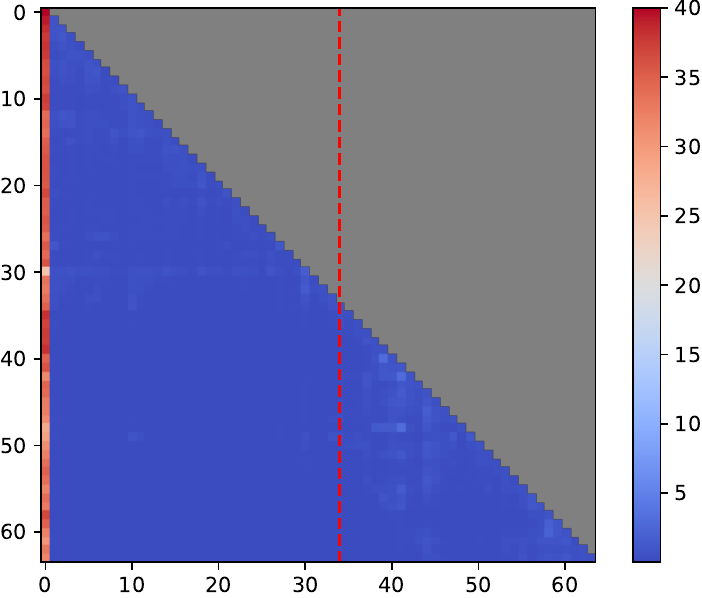}
    }
\subfloat[(f) Attention map of \\Vicuna-v1.5-13B Layer 16]{
        \includegraphics[width=0.24\linewidth]{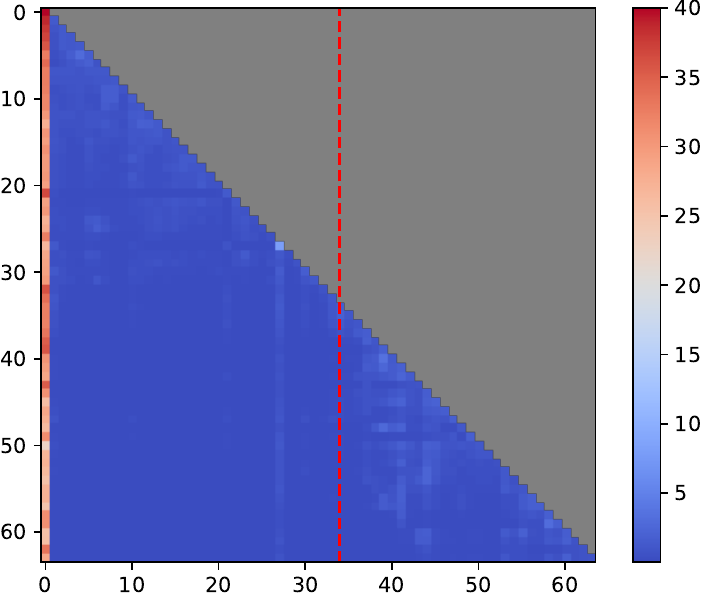}
    }
\subfloat[(g) Attention map of \\Vicuna-v1.5-13B Layer 24]{
        \includegraphics[width=0.24\linewidth]{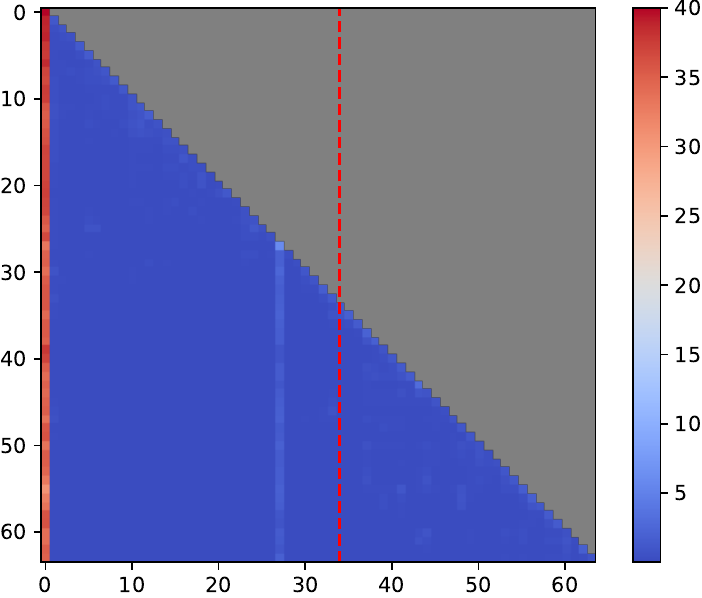}
    }
\subfloat[(h) Attention map of \\Vicuna-v1.5-13B Layer 32]{
        \includegraphics[width=0.24\linewidth]{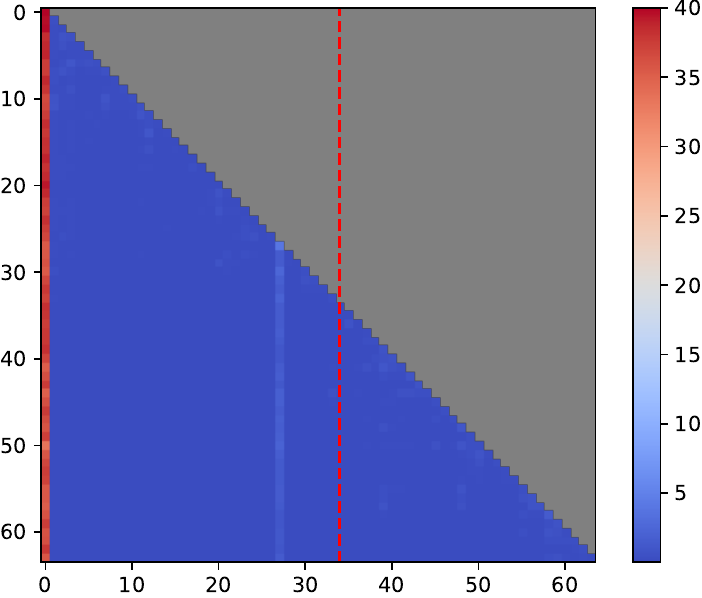}
    }
	\caption{Magnitude of the output activations and attention map in Vicuna-v1.5-13B. The tokens before the red dashed line correspond to the Vicuna system prompt.
 }
\label{apdx-fig:vicuna-v1.5-13b}
\end{figure*}

We provide more visualizations of the output activations and attention map of Vicuna models in Figure.~\ref{apdx-fig:vicuna-v1.3-7b}--\ref{apdx-fig:vicuna-v1.5-13b}. Although Vicuna models demonstrate stronger performance than LLaMA models of the same size, we are surprised to find that the position of pivot tokens remains unchanged for Vicuna and LLaMA models of the same size. Besides, as shown in Figure.~\ref{apdx-fig:vicuna-v1.3-7b}--\ref{apdx-fig:vicuna-v1.5-13b}, we find that the Vicuna system prompt is enough to cover all the pivot tokens in all Vicuna models.

\subsection{Visualization of OPT and Mistral Models}
\label{apdx-subsec:opt_mistral_vis}
\begin{figure*}[t]
\centering
\subfloat[(a) Output activations of OPT-6.7B Layer 0]{
        \includegraphics[width=0.24\linewidth]{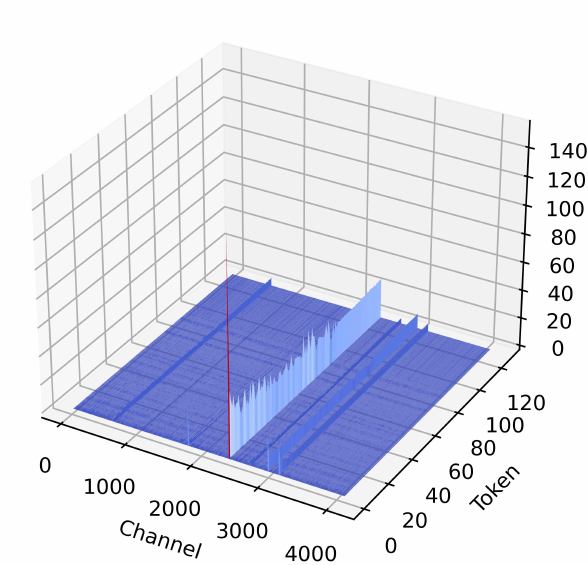}
    }
\subfloat[(b) Output activations of OPT-6.7B Layer 8]{
        \includegraphics[width=0.24\linewidth]{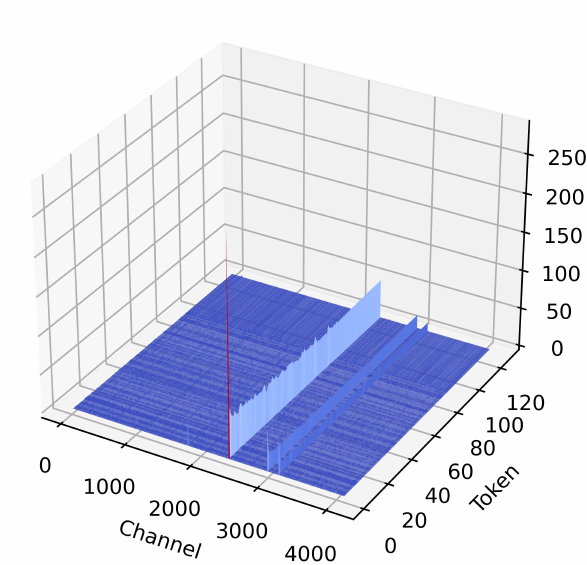}
    }
\subfloat[(c) Output activations of OPT-6.7B Layer 16]{
        \includegraphics[width=0.24\linewidth]{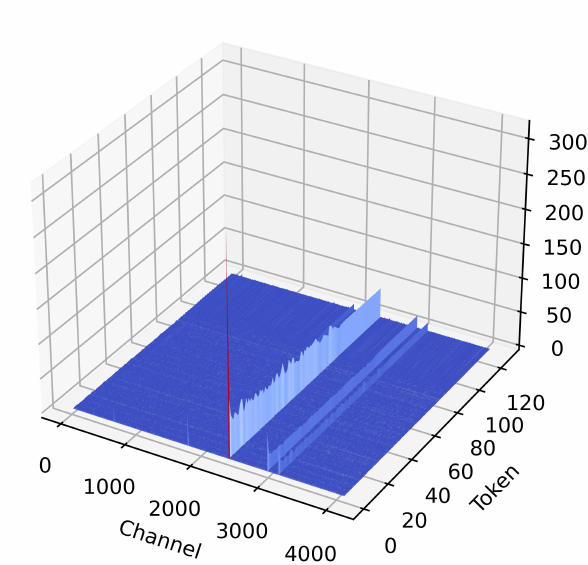}
    }
\subfloat[(d) Output activations of OPT-6.7B Layer 24]{
        \includegraphics[width=0.24\linewidth]{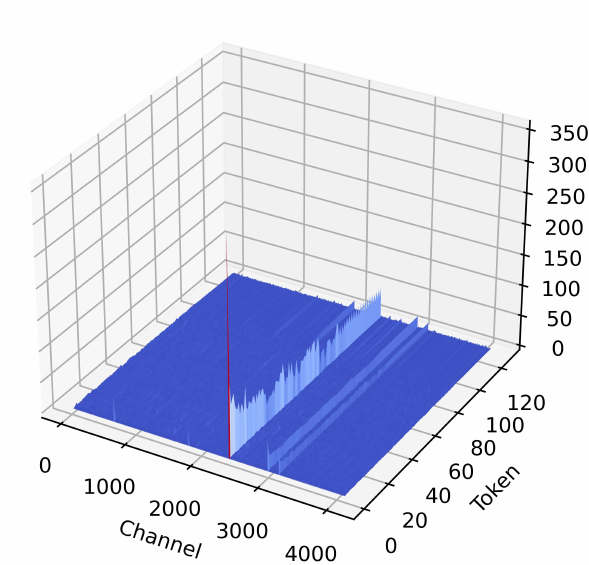}
    }
\vspace{0.25cm}
\subfloat[(e) Attention map of \\OPT-6.7B Layer 0]{
        \includegraphics[width=0.24\linewidth]{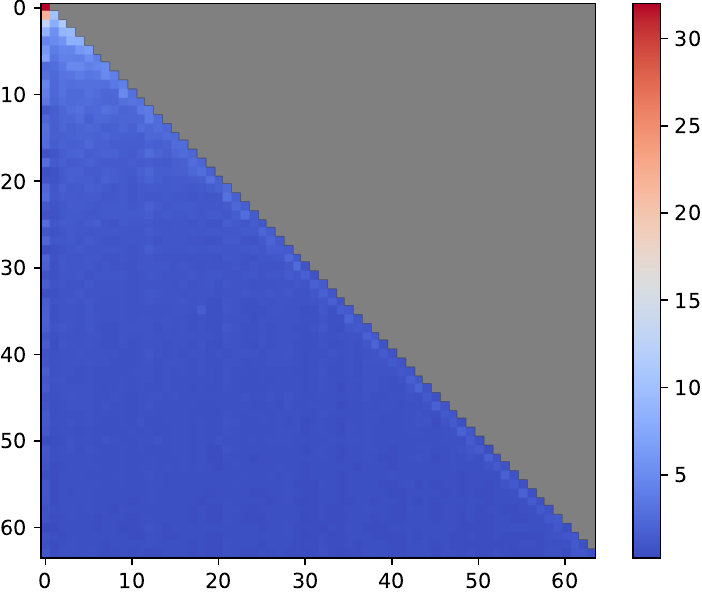}
    }
\subfloat[(f) Attention map of \\OPT-6.7B Layer 8]{
        \includegraphics[width=0.24\linewidth]{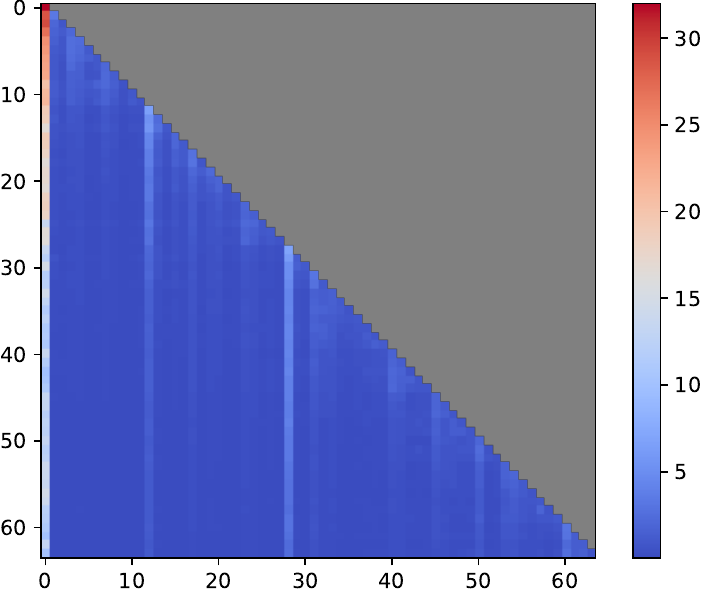}
    }
\subfloat[(g) Attention map of \\OPT-6.7B Layer 16]{
        \includegraphics[width=0.24\linewidth]{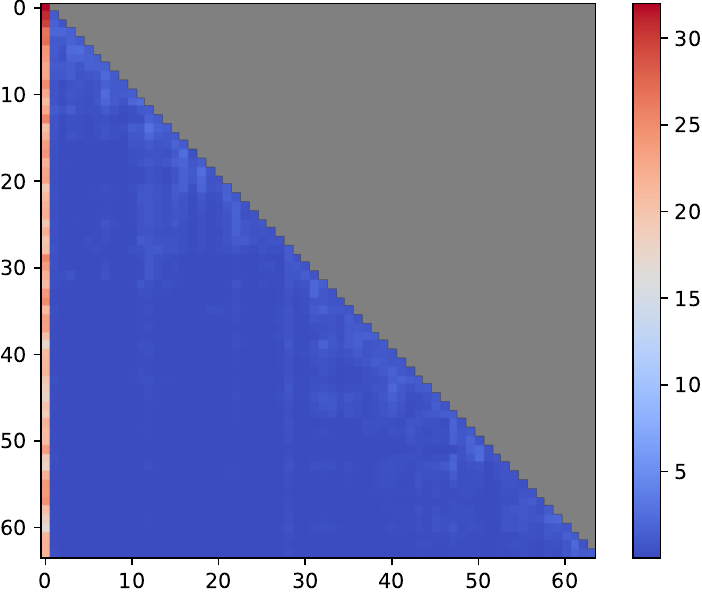}
    }
\subfloat[(h) Attention map of \\OPT-6.7B Layer 24]{
        \includegraphics[width=0.24\linewidth]{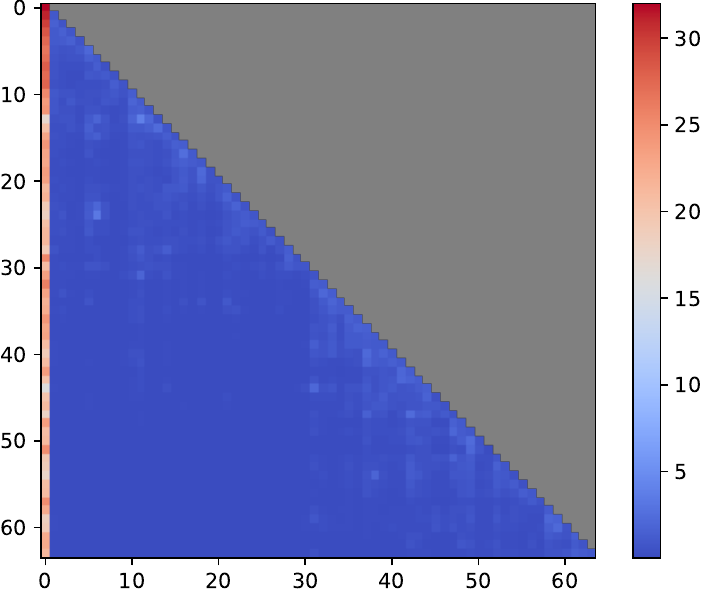}
    }
	\caption{Magnitude of the output activations and attention map in OPT-6.7B.
 }
\label{apdx-fig:opt-6.7b}
\end{figure*}

\begin{figure*}[t]
\centering
\subfloat[(a) Output activations of Mistral-7B Layer 0]{
        \includegraphics[width=0.24\linewidth]{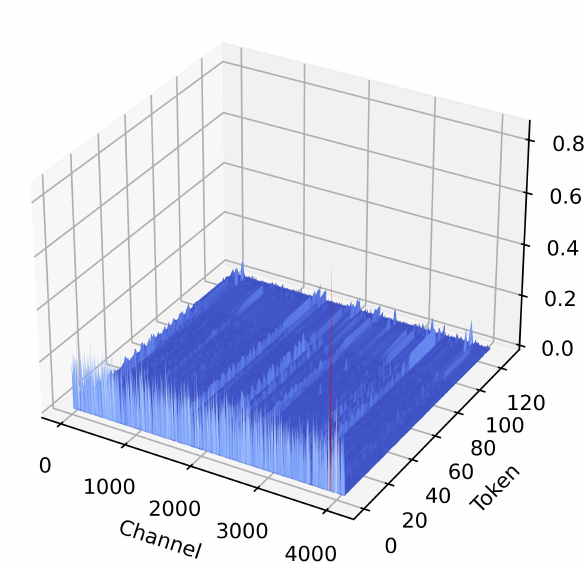}
    }
\subfloat[(b) Output activations of Mistral-7B Layer 8]{
        \includegraphics[width=0.24\linewidth]{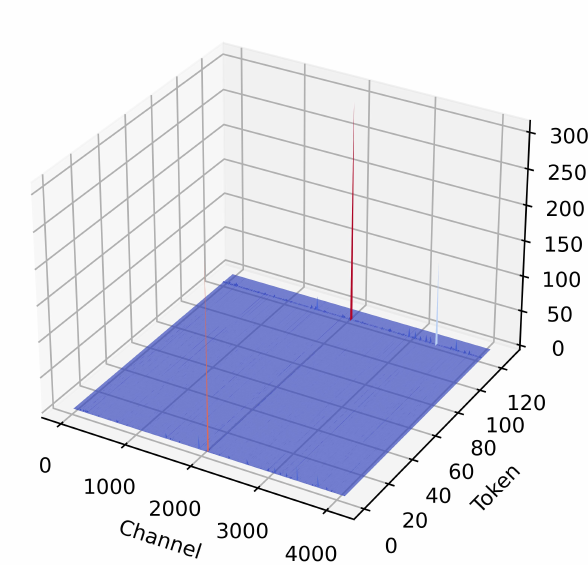}
    }
\subfloat[(c) Output activations of Mistral-7B Layer 16]{
        \includegraphics[width=0.24\linewidth]{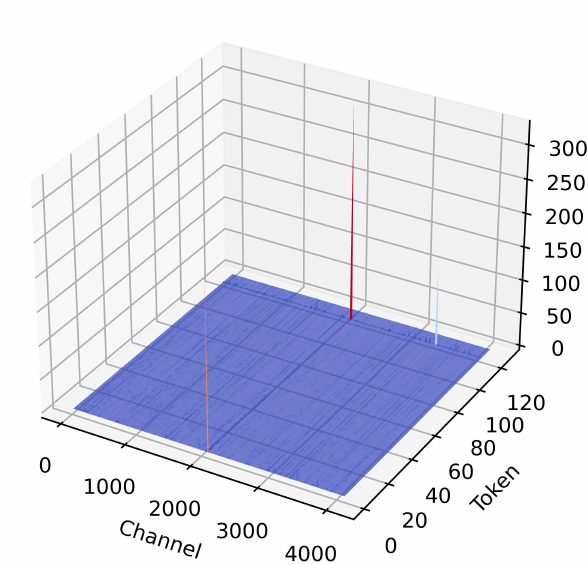}
    }
\subfloat[(d) Output activations of Mistral-7B Layer 24]{
        \includegraphics[width=0.24\linewidth]{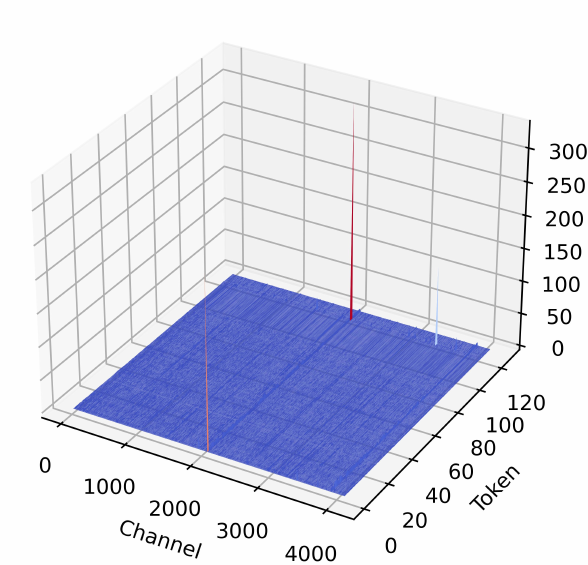}
    }
\vspace{0.25cm}
\subfloat[(e) Attention map of \\Mistral-7B Layer 0]{
        \includegraphics[width=0.24\linewidth]{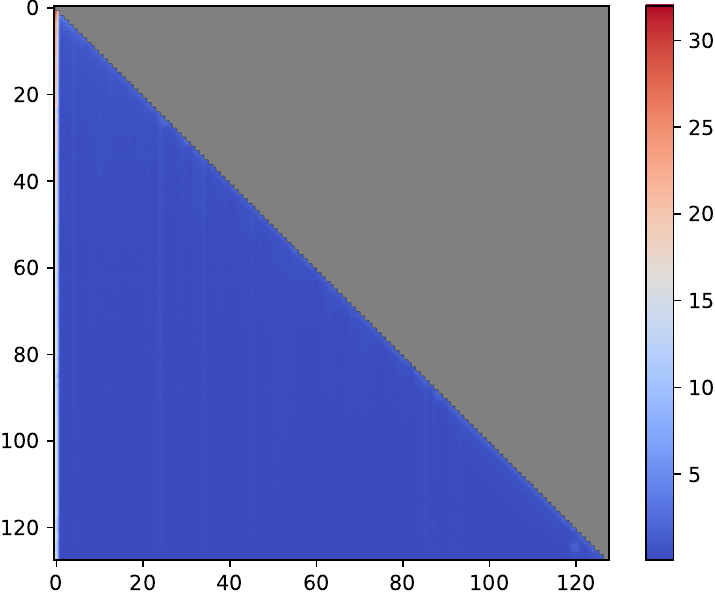}
    }
\subfloat[(f) Attention map of \\Mistral-7B Layer 8]{
        \includegraphics[width=0.24\linewidth]{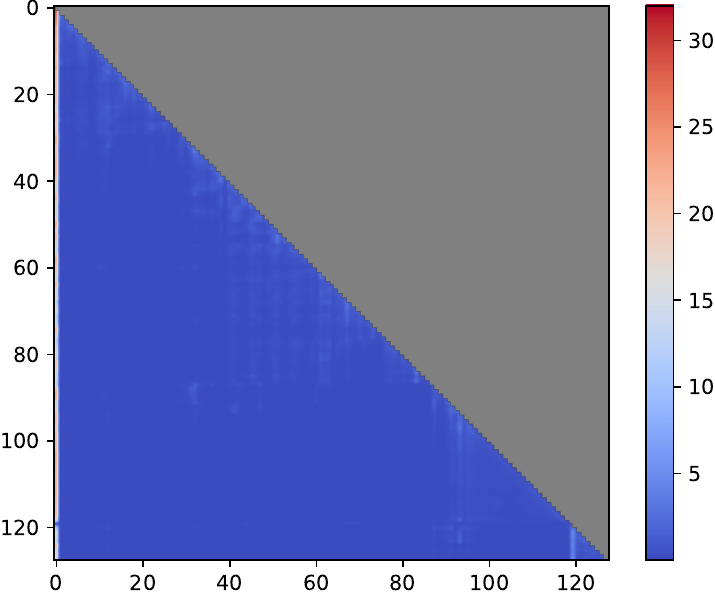}
    }
\subfloat[(g) Attention map of \\Mistral-7B Layer 16]{
        \includegraphics[width=0.24\linewidth]{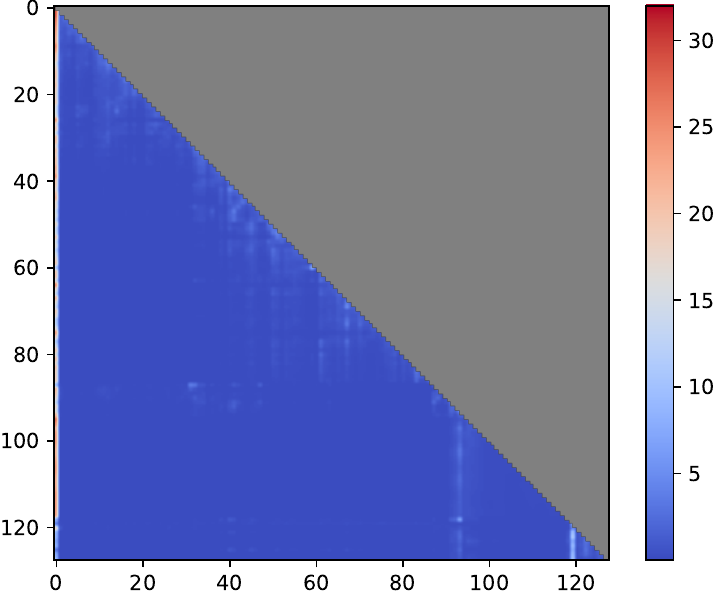}
    }
\subfloat[(h) Attention map of \\Mistral-7B Layer 24]{
        \includegraphics[width=0.24\linewidth]{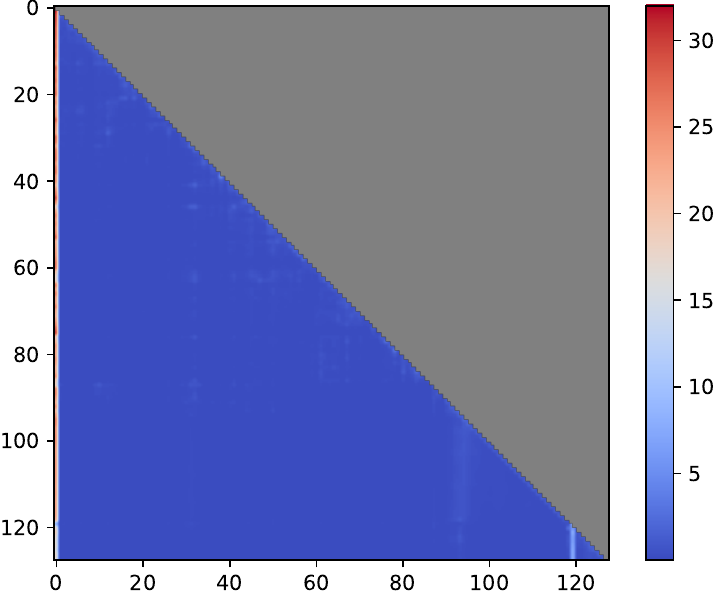}
    }
	\caption{Magnitude of the output activations and attention map in Mistral-7B.
 }
\label{apdx-fig:mistral-v0.1-7b}
\end{figure*}

To demonstrate the prevalence of pivot tokens in LLMs, we provide more visualizations on OPT and Mistral models in Figure.~\ref{apdx-fig:opt-6.7b}--\ref{apdx-fig:mistral-v0.1-7b}. The results show that the pivot tokens with extreme outliers are ubiquitous in various LLMs.

\section{Experiment Details of Figure~\ref{fig:kv-size}}
\label{apdx-sec:kv-size-details}
We plot the quantization loss of the last Transformer layer as well as the total quantization loss of all attention layers with respect to the size of \name\ on four different models, i.e., LLaMA-13B, LLaMA-30B, LLaMA-2-7B, and LLaMA-2-70B, covering different model types and model sizes. We use lossless \name\ generated by the full-precision model to quantify the effect of \name\ on the quantized model. \name\ of size $s$ can ensure that the KV cache of the first $s$ tokens of the input sequence are generated by the full-precision model and thus lossless. Quantization loss is computed with MSE loss between the output activations of the quantized model and the full-precision model. We sample 128 sequences from the ShareGPT dataset to construct the validation set, each with a common prompt prefix of length 34. MSE loss is calculated on the tokens after the common prompt prefix. We quantize the model weights to 3 bits using round-to-nearest quantization with a group size of 128.

\section{Quantization Method Details}
\label{apdx-sec:quant-method-details}
We carefully reproduce the results of various quantization methods with their official code or released checkpoint.
\paragraph{Weight-only Quantization.} For GPTQ, we use AutoGPTQ\footnote{\url{https://github.com/AutoGPTQ/AutoGPTQ}} with C4 calibration set following \citep{frantar2022gptq} to reproduce all results. We turn the quantization option "--desc\_act" on to quantize weight columns in order of decreasing activation size, which is a heuristic rule empirically found to be effective for GPTQ. For AWQ~\cite{lin2023awq}, we directly load the officially released quantization parameters of LLaMA models for evaluation and reproduce results on Vicuna models with their official code\footnote{\url{https://github.com/mit-han-lab/llm-awq}} using Pile \citep{gao2020pile} calibration set. For weight-only quantization of OmniQuant, we reproduce results with their official code\footnote{\url{https://github.com/OpenGVLab/OmniQuant}} using WikiText2 \citep{merity2016wiki} calibration set. We only activate the option "--lwc" to learn the weight clipping parameters for both LLaMA and Vicuna models, following \citep{shao2023omniquant}. Additionally, for OmniQuant+\namebos, we directly integrate \namebos\ into the training process of OmniQuant to adapt the weight clipping parameters to \namebos, which is found to be effective and introduces no extra training costs.
\begin{table}[!t]
\begin{center}
\resizebox{\linewidth}{!}{
\begin{tabular}{l|cc|cc}
\toprule
\multicolumn{1}{c|}{\multirow{2}{*}{\textbf{Method}}} & \multicolumn{2}{c|}{\textbf{OPT-6.7B}}                       & \multicolumn{2}{c}{\textbf{Mistral-7B}}                        \\  
\cmidrule{2-5} \multicolumn{1}{c|}{}                       & \textbf{C4}    & \textbf{WikiText2}           & \textbf{C4}    & \textbf{WikiText2}\\ \midrule
FP16                                         & 12.75 & 10.83           & 8.39 & 5.30           \\\midrule
RTN                                          & 36.18 & 23.91          & 9.65 & 6.20           \\
AWQ                                          & 13.39 & 11.38          & 9.29 & 5.95           \\
\rowcolor{orange!10} \ \ +\namebos           &\textbf{13.37}& \textbf{11.32}         & \textbf{9.25} & \textbf{5.93} \\
\bottomrule
\end{tabular}
}
\end{center}
\vspace{-10pt}
\caption{INT3-group128 weight-only quantization results of OPT and Mistral on C4 and WikiText2 datasets.
}
\label{apdx-tab:ppl_backbone}
\end{table}
\begin{table*}[!h]
\begin{center}
\resizebox{\linewidth}{!}{
\begin{tabular}{l|l|ccccc|ccccc}
\toprule
\multicolumn{1}{c|}{\multirow{2}{*}{\textbf{Model}}} & \multicolumn{1}{c|}{\multirow{2}{*}{\textbf{Method}}} & \multicolumn{5}{c|}{\textbf{MMLU (0 shot)}}                       & \multicolumn{5}{c}{\textbf{MMLU (5 shot)}}                        \\ \cmidrule{3-12} 
\multicolumn{1}{c|}{}                       & \multicolumn{1}{c|}{}                        & \textbf{Hums}    & \textbf{STEM}    & \textbf{Social}  & \textbf{Others}  &\textbf{ Avg}           & \textbf{Hums}    & \textbf{STEM}    & \textbf{Social}  & \textbf{Others}  & \textbf{Avg}             \\ \midrule
\multirow{6}{*}{Vicuna-v1.5-7B}             & FP16                                         & 45.40\% & 38.67\% & 56.16\% & 55.92\% & 48.74\%          & 45.78\% & 39.50\% & 58.14\% & 57.46\% & 49.84\%          \\ \cmidrule{2-12}
                                            & RTN                                          & 42.06\% & 34.16\% & 50.47\% & 50.59\% & 44.17\%          & 40.68\% & 38.60\% & 50.31\% & 50.56\% & 44.62\%          \\
                                            & GPTQ                                         & 39.89\% & 33.00\% & 48.10\% & 48.46\% & 42.19\%          & 40.30\% & 36.28\% & 50.76\% & 50.09\% & 43.99\%          \\
                                            & OmniQuant                                    & 42.72\% & 36.38\% & 51.93\% & 53.55\% & \textbf{45.88\%} & 42.70\% & 37.97\% & 54.31\% & 53.08\% & 46.62\%          \\
                                            & AWQ                                          & 42.08\% & 35.55\% & 51.61\% & 51.54\% & 44.95\%          & 42.55\% & 38.93\% & 53.10\% & 52.78\% & 46.45\%          \\
                                            \rowcolor{orange!10} \cellcolor{white!10}&
                                            \ \ +\namebos                     & 42.42\% & 35.42\% & 51.71\% & 51.57\% & 45.06\%          & 42.95\% & 38.60\% & 54.37\% & 53.15\% & \textbf{46.87\%} \\ \midrule
\multirow{6}{*}{Vicuna-v1.5-13B}            & FP16          & 50.48\%          & 43.70\%          & 62.72\%          & 62.74\%          & 54.54\%          & 51.97\%          & 44.96\%          & 65.26\%          & 62.40\%          & 55.78\%          \\ \cmidrule{2-12}  
                                            & RTN           & 46.61\%          & 41.32\%          & 58.92\%          & 57.53\%          & 50.69\%          & 47.14\%          & 42.81\%          & 59.38\%          & 58.17\%          & 51.44\%          \\
                                            & GPTQ          & 48.35\%          & 40.99\%          & 59.25\%          & 57.99\%          & 51.38\%          & 49.63\%          & 43.04\%          & 60.22\%          & 60.09\%          & 52.95\%          \\
                                            & OmniQuant     & 49.73\% & 41.02\%          & 59.31\%          & 58.33\%          & 51.94\%          & 49.18\%          & 44.17\% & 60.45\%          & 58.91\%          & 52.82\%          \\
                                            & AWQ           & 48.82\%          & 41.72\%          & 61.03\%          & 58.30\%          & 52.16\%          & 49.52\%          & 43.01\%          & 61.72\%          & 58.73\%          & 52.92\%          \\
                                            \rowcolor{orange!10} \cellcolor{white!10}&
                                            \ \ +\namebos & 49.31\%          & 42.18\% & 61.20\% & 59.28\% & \textbf{52.68\%} & 50.31\% & 43.37\%          & 61.91\% & 59.93\% & \textbf{53.58\%} \\ \midrule
\multirow{6}{*}{Vicuna-v1.3-7B}             & FP16                                         & 44.31\% & 36.28\% & 53.23\% & 53.70\% & 46.71\%          & 44.23\% & 38.34\% & 53.82\% & 53.15\% & 47.12\%          \\ \cmidrule{2-12}
                                            & RTN                                          & 38.09\% & 31.58\% & 42.35\% & 44.32\% & 39.06\%          & 36.81\% & 32.77\% & 43.87\% & 44.79\% & 39.33\%          \\
                                            & GPTQ                                         & 39.09\% & 32.57\% & 44.59\% & 46.73\% & 40.66\%          & 36.94\% & 33.90\% & 45.08\% & 45.81\% & 40.12\%          \\
                                            & OmniQuant                                    & 41.40\% & 34.06\% & 48.07\% & 48.06\% & 42.82\%          & 40.98\% & 35.19\% & 48.23\% & 48.03\% & 42.95\%          \\
                                            & AWQ                                          & 40.49\% & 32.44\% & 47.06\% & 49.57\% & 42.29\%          & 39.64\% & 36.22\% & 48.72\% & 49.11\% & 43.08\%          \\
                                            \rowcolor{orange!10} \cellcolor{white!10}&
                                            \ \ +\namebos                     & 41.76\% & 32.94\% & 47.74\% & 49.72\% & \textbf{43.01\%} & 41.93\% & 36.58\% & 50.37\% & 50.77\% & \textbf{44.67\%} \\ \midrule
\multirow{6}{*}{Vicuna-v1.3-13B}            & FP16                                         & 47.89\% & 39.96\% & 58.86\% & 57.34\% & 50.77\%          & 49.78\% & 40.46\% & 60.61\% & 58.24\% & 52.10\%          \\ \cmidrule{2-12} 
                                            & RTN                                          & 42.06\% & 32.87\% & 47.61\% & 49.51\% & 43.02\%          & 42.42\% & 34.46\% & 50.34\% & 51.57\% & 44.56\%          \\
                                            & GPTQ                                         & 45.06\% & 35.88\% & 52.23\% & 51.26\% & 46.09\%          & 45.82\% & 37.57\% & 54.83\% & 53.64\% & 47.83\%          \\
                                            & OmniQuant                                    & 43.51\% & 36.85\% & 52.16\% & 53.05\% & 46.18\%          & 45.91\% & 37.44\% & 55.31\% & 54.94\% & 48.23\%          \\
                                            & AWQ                                          & 45.14\% & 36.18\% & 52.55\% & 53.79\% & 46.84\%          & 46.65\% & 37.64\% & 55.54\% & 54.87\% & 48.56\%          \\
                                            \rowcolor{orange!10} \cellcolor{white!10}&
                                            \ \ +\namebos                     & 45.91\% & 36.65\% & 53.75\% & 54.60\% & \textbf{47.64\%} & 46.57\% & 38.40\% & 56.03\% & 55.95\% & \textbf{49.05\%} \\ \midrule
\multirow{6}{*}{Vicuna-v1.3-33B}            & FP16                                         & 53.73\% & 44.14\% & 67.63\% & 63.54\% & 56.98\%          & 57.66\% & 46.32\% & 69.32\% & 64.25\% & 59.30\%          \\ \cmidrule{2-12}
                                            & RTN                                          & 49.88\% & 40.13\% & 61.33\% & 58.42\% & 52.26\%          & 51.26\% & 42.54\% & 61.75\% & 57.71\% & 53.18\%          \\
                                            & GPTQ                                         & 51.22\% & 40.03\% & 61.85\% & 59.47\% & 53.05\%          & 54.05\% & 44.04\% & 64.35\% & 61.35\% & 55.84\%          \\
                                            & OmniQuant                                    & 51.22\% & 42.18\% & 64.06\% & 60.39\% & 54.21\%          & 53.94\% & 44.10\% & 63.21\% & 59.81\% & 55.21\%          \\
                                            & AWQ                                          & 51.69\% & 42.74\% & 63.41\% & 61.38\% & 54.57\%          & 54.56\% & 44.10\% & 65.36\% & 60.67\% & 56.09\%          \\
                                            \rowcolor{orange!10} \cellcolor{white!10}&
                                            \ \ +\namebos                     & 52.09\% & 42.68\% & 63.70\% & 62.03\% & \textbf{54.91\%} & 55.79\% & 44.90\% & 65.62\% & 61.47\% & \textbf{56.91\%} \\ 
\bottomrule
\end{tabular}
}
\end{center}
\vspace{-10pt}
\caption{INT3-group128 weight-only quantization results of Vicuna models on MMLU benchmarks.
}
\label{apdx-tab:mmlu_3bit}
\end{table*}
\begin{table*}[!h]
\begin{center}
\resizebox{\linewidth}{!}{
\begin{tabular}{l|l|ccccc|ccccc}
\toprule
\multicolumn{1}{c|}{\multirow{2}{*}{\textbf{Model}}} & \multicolumn{1}{c|}{\multirow{2}{*}{\textbf{Method}}} & \multicolumn{5}{c|}{\textbf{MMLU (0 shot)}}                       & \multicolumn{5}{c}{\textbf{MMLU (5 shot)}}                        \\ \cmidrule{3-12} 
\multicolumn{1}{c|}{}                       & \multicolumn{1}{c|}{}                        & \textbf{Hums}    & \textbf{STEM}    & \textbf{Social}  & \textbf{Others}  & \textbf{Avg}             & \textbf{Hums}    & \textbf{STEM}    & \textbf{Social}  & \textbf{Others}  & \textbf{Avg}             \\ \midrule
\multirow{6}{*}{Vicuna-v1.5-7B}             & FP16                                         & 45.40\% & 38.67\% & 56.16\% & 55.92\% & 48.74\%          & 45.78\% & 39.50\% & 58.14\% & 57.46\% & 49.84\%          \\ \cmidrule{2-12}
                                            & RTN                                          & 44.65\% & 38.47\% & 53.95\% & 54.41\% & 47.61\%          & 44.87\% & 39.13\% & 56.45\% & 55.34\% & 48.59\%          \\
                                            & GPTQ                                         & 44.87\% & 37.08\% & 54.44\% & 53.86\% & 47.37\%          & 45.44\% & 38.83\% & 57.33\% & 56.14\% & 49.10\%          \\
                                            & OmniQuant                                    & 44.97\% & 38.80\% & 55.57\% & 56.32\% & \textbf{48.59\%} & 45.53\% & 39.40\% & 57.20\% & 57.50\% & \textbf{49.53\%}          \\
                                            & AWQ                                          & 45.08\% & 37.41\% & 55.64\% & 55.31\% & 48.11\%          & 45.44\% & 38.97\% & 56.94\% & 55.74\% & 48.95\%          \\
                                            \rowcolor{orange!10} \cellcolor{white!10}&
                                            \ \ +\namebos                     & 45.25\% & 37.51\% & 55.93\% & 55.58\% & 48.31\%          & 45.33\% & 39.60\% & 57.36\% & 55.74\% & 49.14\% \\ \midrule
\multirow{6}{*}{Vicuna-v1.5-13B}            & FP16          & 50.48\%          & 43.70\%          & 62.72\%          & 62.74\%          & 54.54\%          & 51.97\%          & 44.96\%          & 65.26\%          & 62.40\%          & 55.78\%          \\ \cmidrule{2-12}  
                                            & RTN           & 50.01\%          & 43.41\%          & 62.33\%          & 62.00\%          & 54.06\%          & 51.31\%          & 43.14\%          & 63.54\%          & 61.63\%          & 54.61\%          \\
                                            & GPTQ          & 50.20\%          & 42.31\%          & 61.62\%          & 61.41\%          & 53.60\%          & 50.10\%          & 43.97\%          & 62.72\%          & 61.01\%          & 54.07\%          \\
                                            & OmniQuant     & 49.99\% & 43.97\%          & 62.40\%          & 62.03\%          & \textbf{54.19\%}          & 51.67\%          & 43.90\% & 63.05\%          & 61.81\%          & 54.84\%          \\
                                            & AWQ           & 50.10\%          & 42.94\%          & 61.68\%          & 61.66\%          & 53.77\%          & 52.31\%          & 44.43\%          & 63.18\%          & 61.84\%          & 55.20\%          \\
                                            \rowcolor{orange!10} \cellcolor{white!10}&
                                            \ \ +\namebos & 50.14\%          & 42.84\%          & 61.78\%          & 61.91\%          & 53.84\%          & 52.31\%          & 44.37\%          & 63.67\%          & 61.91\%          & \textbf{55.31\%} \\ \midrule
\multirow{6}{*}{Vicuna-v1.3-7B}             & FP16                                         & 44.31\% & 36.28\% & 53.23\% & 53.70\% & 46.71\%          & 44.23\% & 38.34\% & 53.82\% & 53.15\% & 47.12\%          \\ \cmidrule{2-12}
                                            & RTN                                          & 42.78\% & 36.55\% & 51.74\% & 51.48\% & 45.41\%          & 42.23\% & 37.08\% & 52.10\% & 51.94\% & 45.53\%          \\
                                            & GPTQ                                         & 43.40\% & 34.46\% & 52.06\% & 53.45\% & 45.70\%          & 43.78\% & 36.41\% & 53.49\% & 52.41\% & 46.32\%          \\
                                            & OmniQuant                                    & 43.12\% & 34.59\% & 52.45\% & 52.31\% & 45.46\%          & 43.04\% & 37.67\% & 52.75\% & 53.08\% & 46.33\%          \\
                                            & AWQ                                          & 43.53\% & 36.22\% & 53.01\% & 52.53\% & 46.11\%          & 43.36\% & 37.74\% & 53.46\% & 52.68\% & \textbf{46.52\%}          \\
                                            \rowcolor{orange!10} \cellcolor{white!10}&
                                            \ \ +\namebos                     & 43.57\% & 36.51\% & 52.29\% & 53.27\% & \textbf{46.20\%} & 43.51\% & 37.44\% & 53.17\% & 52.62\% & 46.43\% \\ \midrule
\multirow{6}{*}{Vicuna-v1.3-13B}            & FP16                                         & 47.89\% & 39.96\% & 58.86\% & 57.34\% & 50.77\%          & 49.78\% & 40.46\% & 60.61\% & 58.24\% & 52.10\%          \\ \cmidrule{2-12} 
                                            & RTN                                          & 47.16\% & 39.00\% & 56.52\% & 56.63\% & 49.64\%          & 49.25\% & 39.63\% & 57.85\% & 57.74\% & 51.03\%          \\
                                            & GPTQ                                         & 46.95\% & 39.30\% & 57.39\% & 56.23\% & 49.74\%          & 49.05\% & 39.46\% & 59.02\% & 57.65\% & 51.16\%          \\
                                            & OmniQuant                                    & 47.52\% & 39.40\% & 57.98\% & 57.37\% & \textbf{50.34\%}          & 49.03\% & 40.09\% & 59.34\% & 58.11\% & 51.47\%          \\
                                            & AWQ                                          & 48.03\% & 39.43\% & 56.94\% & 56.76\% & 50.15\%          & 49.44\% & 40.49\% & 59.57\% & 57.65\% & 51.63\%          \\
                                            \rowcolor{orange!10} \cellcolor{white!10}&
                                            \ \ +\namebos                     & 47.91\% & 39.60\% & 57.69\% & 56.79\% & 50.31\% & 49.54\% & 40.23\% & 60.12\% & 57.71\% & \textbf{51.74\%} \\ \midrule
\multirow{6}{*}{Vicuna-v1.3-33B}            & FP16                                         & 53.73\% & 44.14\% & 67.63\% & 63.54\% & 56.98\%          & 57.66\% & 46.32\% & 69.32\% & 64.25\% & 59.30\%          \\ \cmidrule{2-12}
                                            & RTN                                          & 53.18\% & 44.27\% & 66.88\% & 62.95\% & 56.52\%          & 56.73\% & 45.73\% & 68.09\% & 62.49\% & 58.18\%          \\
                                            & GPTQ                                         & 52.92\% & 44.90\% & 67.05\% & 63.66\% & 56.77\%          & 57.13\% & 45.96\% & 67.63\% & 63.11\% & 58.41\%          \\
                                            & OmniQuant                                    & 53.22\% & 44.43\% & 67.73\% & 63.26\% & 56.73\%          & 56.83\% & 45.46\% & 68.67\% & 62.31\% & 58.25\%          \\
                                            & AWQ                                          & 53.22\% & 44.40\% & 67.63\% & 63.54\% & 56.87\%          & 56.85\% & 45.69\% & 68.80\% & 63.66\% & 58.65\%          \\
                                            \rowcolor{orange!10} \cellcolor{white!10}& 
                                            \ \ +\namebos                     & 53.37\% & 44.40\% & 67.50\% & 63.63\% & \textbf{56.91\%} & 57.07\% & 45.96\% & 68.51\% & 63.63\% & \textbf{58.70\%} \\ 
\bottomrule
\end{tabular}
}
\end{center}
\vspace{-10pt}
\caption{INT4-group128 weight-only quantization results of Vicuna models on MMLU benchmarks.
}
\label{apdx-tab:mmlu_4bit}
\end{table*}
\begin{table*}[!t]
\begin{center}
\resizebox{\linewidth}{!}{
\begin{tabular}{l|l|l|cccccccc}
\toprule
\multicolumn{1}{c|}{\textbf{Model}} & \multicolumn{1}{c|}{\textbf{\#bits}} & \multicolumn{1}{c|}{\textbf{Method}} & \textbf{OBQA} & \textbf{WinoGrande} & \textbf{ARC-C} & \textbf{ARC-E} & \textbf{BoolQ} & \textbf{HellaSwag} & \textbf{LAMBADA} & \textbf{Avg}     \\ 
\midrule
\multirow{11}{*}{Vicuna-V1.5-7B}    & FP16                                 & -                                    & 45.00\%       & 69.53\%             & 45.73\%        & 71.25\%        & 80.92\%        & 73.78\%            & 71.12\%          & 65.33\%          \\ 
\cmidrule{2-11} 
                                    & \multirow{5}{*}{w3g128}           & RTN                                  & 40.60\% & 66.22\% & 43.77\% & 67.89\% & 77.86\% & 71.46\% & 61.75\% & 61.36\%          \\
                                    &                                      & GPTQ                                 & 39.40\% & 64.72\% & 40.87\% & 65.07\% & 74.77\% & 66.32\% & 59.09\% & 58.61\%          \\
                                    &                                      & OmniQuant                            & 43.00\% & 66.46\% & 43.69\% & 67.72\% & 78.59\% & 70.53\% & 66.12\% & 62.30\%          \\
                                    &                                      & AWQ                                  & 41.60\% & 67.56\% & 42.66\% & 67.85\% & 78.96\% & 71.32\% & 65.28\% & 62.18\%          \\
                                    &                                      & \cellcolor{orange!10} \ \ +\namebos      & \cellcolor{orange!10}42.20\%       & \cellcolor{orange!10}67.64\%             & \cellcolor{orange!10}41.98\%        & \cellcolor{orange!10}68.52\%        & \cellcolor{orange!10}79.02\%        & \cellcolor{orange!10}71.24\%            & \cellcolor{orange!10}66.82\%          & \cellcolor{orange!10}\textbf{62.49\%} \\ 
\cmidrule{2-11} 
                                    & \multirow{5}{*}{w4g128}           & RTN                                  & 43.40\% & 68.98\% & 44.80\% & 71.09\% & 82.05\% & 73.32\% & 69.28\% & 64.70\%          \\
                                    &                                      & GPTQ                                 & 43.60\% & 69.77\% & 44.62\% & 70.20\% & 74.01\% & 72.61\% & 68.27\% & 63.30\%          \\
                                    &                                      & OmniQuant                            & 43.40\% & 69.06\% & 44.37\% & 71.17\% & 81.83\% & 72.90\% & 70.13\% & 64.69\%          \\
                                    &                                      & AWQ                                  & 43.80\% & 68.59\% & 45.73\% & 71.09\% & 82.02\% & 73.51\% & 69.42\% & 64.88\%          \\
                                    \rowcolor{orange!10} \cellcolor{white!10}&\cellcolor{white!10}& \ \ +\namebos      & 44.00\%       & 68.90\%             & 45.90\%        & 71.63\%        & 82.29\%        & 73.52\%            & 69.61\%          & \textbf{65.12\%} \\ 
\midrule
\multirow{11}{*}{Vicuna-v1.5-13B} & FP16                        & -                           & 45.40\%          & 71.51\%          & 50.68\%          & 74.87\%          & 85.29\%          & 77.50\%          & 73.43\%          & 68.38\%          \\ \cmidrule{2-11} 
                                 & \multirow{5}{*}{w3g128}  
& RTN                         & 43.60\% & 71.27\% & 48.55\% & 72.81\% & 82.91\% & 74.55\% & 69.18\% & 66.12\%          \\
                                 &                             
& GPTQ                        & 43.00\% & 70.09\% & 48.98\% & 72.98\% & 84.43\% & 74.80\% & 70.11\% & 66.34\%          \\
                                 &                             
& OmniQuant                   & 43.60\% & 69.85\% & 47.78\% & 71.17\% & 82.45\% & 74.16\% & 70.04\% & 65.58\%          \\
                                 &                             
& AWQ                         & 45.40\% & 69.38\% & 48.38\% & 71.89\% & 84.46\% & 75.24\% & 70.85\% & 66.51\%          \\
                                 &                             
& \cellcolor{orange!10}
\ \ +\namebos   & \cellcolor{orange!10}45.40\% & \cellcolor{orange!10}70.32\% & \cellcolor{orange!10}48.38\% & \cellcolor{orange!10}72.14\% & \cellcolor{orange!10}85.20\% & \cellcolor{orange!10}75.23\% & \cellcolor{orange!10}71.86\% & \cellcolor{orange!10}\textbf{66.93\%} \\ \cmidrule{2-11} 
                                 & \multirow{5}{*}{w4g128}  
& RTN                         & 44.80\% & 71.51\% & 49.15\% & 73.78\% & 85.20\% & 76.70\% & 72.62\% & 67.68\%          \\
                                 &                             
& GPTQ                        & 45.80\% & 70.96\%          & 50.51\% & 73.99\%          & 85.47\%          & 76.70\%          & 73.43\% & 68.12\%          \\
                                 &                             
& OmniQuant                   & 44.40\% & 70.80\% & 50.09\% & 73.86\% & 85.29\% & 76.79\% & 72.39\% & 67.66\%          \\
                                 &                             
& AWQ                         & 45.60\% & 72.85\% & 49.49\% & 74.07\% & 85.72\% & 77.37\% & 72.37\% & 68.21\%          \\
\rowcolor{orange!10} \cellcolor{white!10}&\cellcolor{white!10}& 
\ \ +\namebos  & 45.40\%          & 73.09\% & 49.57\%          & 74.45\% & 85.66\% & 77.32\% & 72.75\%          & \textbf{68.32\%} \\
\midrule
\multirow{11}{*}{Vicuna-V1.3-7B}    & FP16                                 & -                                    & 43.80\%       & 69.46\%             & 44.54\%        & 71.89\%        & 78.07\%        & 73.93\%            & 69.98\%          & 64.52\%          \\ \cline{2-11} 
                                    & \multirow{5}{*}{w3g128}           & RTN                                  & 41.80\% & 63.38\% & 38.91\% & 63.47\% & 76.57\% & 68.92\% & 60.29\% & 59.05\%          \\
                                    &                                      & GPTQ                                 & 40.00\% & 65.90\% & 41.55\% & 66.16\% & 70.73\% & 69.66\% & 62.95\% & 59.56\%          \\
                                    &                                      & OmniQuant                            & 42.00\% & 66.06\% & 39.68\% & 66.67\% & 75.69\% & 70.45\% & 65.65\% & 60.89\%          \\
                                    &                                      & AWQ                                  & 42.40\% & 66.69\% & 39.51\% & 65.40\% & 77.06\% & 70.53\% & 63.69\% & 60.75\%          \\
                                    &                                      & \cellcolor{orange!10} \ \ +\namebos      & \cellcolor{orange!10}43.60\%       & \cellcolor{orange!10}68.43\%             & \cellcolor{orange!10}39.16\%        & \cellcolor{orange!10}67.30\%        & \cellcolor{orange!10}77.28\%        & \cellcolor{orange!10}71.20\%            & \cellcolor{orange!10}66.54\%          & \cellcolor{orange!10}\textbf{61.93\%} \\ 
\cmidrule{2-11} 
                                    & \multirow{5}{*}{w4g128}           & RTN                                  & 42.20\% & 67.80\% & 43.00\% & 70.66\% & 75.50\% & 73.16\% & 68.37\% & 62.96\%          \\
                                    &                                      & GPTQ                                 & 45.20\% & 68.82\% & 42.41\% & 70.45\% & 67.58\% & 72.50\% & 67.40\% & 62.05\%          \\
                                    &                                      & OmniQuant                            & 43.40\% & 67.96\% & 44.28\% & 71.46\% & 76.42\% & 73.22\% & 68.81\% & 63.65\%          \\
                                    &                                      & AWQ                                  & 43.60\% & 68.03\% & 43.26\% & 71.68\% & 75.87\% & 73.44\% & 68.45\% & 63.48\%          \\
                                    \rowcolor{orange!10} \cellcolor{white!10}&\cellcolor{white!10}& \ \ +\namebos      & 43.80\%       & 68.59\%             & 42.92\%        & 71.84\%        & 76.79\%        & 73.49\%            & 69.57\%          & \textbf{63.86\%} \\ 
\midrule
\multirow{11}{*}{Vicuna-V1.3-13B}   & FP16                                 & -                                    & 45.40\%       & 71.03\%             & 47.70\%        & 73.70\%        & 82.81\%        & 77.00\%            & 72.91\%          & 67.22\%          \\ \cline{2-11} 
                                    & \multirow{5}{*}{w3g128}           & RTN                                  & 44.00\% & 70.96\% & 44.03\% & 67.30\% & 80.40\% & 73.33\% & 64.00\% & 63.43\%          \\
                                    &                                      & GPTQ                                 & 45.20\% & 69.77\% & 46.08\% & 70.33\% & 81.90\% & 74.89\% & 67.59\% & \textbf{65.11\%}          \\
                                    &                                      & OmniQuant                            & 45.20\% & 69.22\% & 45.22\% & 68.90\% & 80.95\% & 74.72\% & 68.15\% & 64.62\%          \\
                                    &                                      & AWQ                                  & 42.80\% & 68.98\% & 46.08\% & 68.98\% & 81.31\% & 74.97\% & 68.78\% & 64.56\%          \\
                                    &                                      & \cellcolor{orange!10} \ \ +\namebos      & \cellcolor{orange!10}43.20\%       & \cellcolor{orange!10}69.46\%             & \cellcolor{orange!10}46.16\%        & \cellcolor{orange!10}69.74\%        & \cellcolor{orange!10}81.80\%        & \cellcolor{orange!10}75.11\%            & \cellcolor{orange!10}69.67\%          & \cellcolor{orange!10}65.02\% \\ \cline{2-11} 
                                    & \multirow{5}{*}{w4g128}           & RTN                                  & 45.20\% & 71.43\% & 48.04\% & 73.15\% & 82.87\% & 76.56\% & 70.62\% & 66.84\%          \\
                                    &                                      & GPTQ                                 & 44.60\%       & 70.01\%             & 47.87\%        & 73.32\%        & 82.23\%        & 76.55\%            & 71.78\%          & 66.62\%          \\
                                    &                                      & OmniQuant                            & 45.60\% & 70.56\% & 46.76\% & 73.02\% & 82.81\% & 76.74\% & 70.41\% & 66.56\%          \\
                                    &                                      & AWQ                                  & 45.20\% & 70.32\% & 47.27\% & 73.91\% & 82.81\% & 76.79\% & 71.32\% & 66.80\%          \\
                                    \rowcolor{orange!10} \cellcolor{white!10}&\cellcolor{white!10}& \ \ +\namebos      & 45.60\%       & 71.19\%             & 47.10\%        & 73.32\%        & 82.72\%        & 76.95\%            & 71.38\%          & \textbf{66.89\%} \\ 
\midrule
\multirow{11}{*}{Vicuna-V1.3-33B}   & FP16                                 & -                                    & 47.80\%       & 74.35\%             & 51.79\%        & 74.71\%        & 83.91\%        & 80.38\%            & 73.74\%          & 69.53\%          \\ \cline{2-11} 
                                    & \multirow{5}{*}{w3g128}           & RTN                                  & 46.60\% & 72.53\% & 49.06\% & 72.18\% & 83.12\% & 78.06\% & 69.73\% & 67.33\%          \\
                                    &                                      & GPTQ                                 & 44.80\% & 71.74\% & 47.01\% & 70.12\% & 83.64\% & 77.79\% & 71.51\% & 66.66\%          \\
                                    &                                      & OmniQuant                            & 45.40\% & 73.64\% & 48.63\% & 72.35\% & 83.55\% & 77.98\% & 71.73\% & 67.61\%          \\
                                    &                                      & AWQ                                  & 45.60\% & 73.32\% & 50.68\% & 71.63\% & 82.39\% & 78.55\% & 71.49\% & 67.67\%          \\
                                    &                                      & \cellcolor{orange!10} \ \ +\namebos      & \cellcolor{orange!10}44.80\%       & \cellcolor{orange!10}73.56\%             & \cellcolor{orange!10}51.11\%        & \cellcolor{orange!10}72.60\%        & \cellcolor{orange!10}82.78\%        & \cellcolor{orange!10}78.55\%            & \cellcolor{orange!10}71.90\%          & \cellcolor{orange!10}\textbf{67.90\%} \\ 
\cmidrule{2-11} 
                                    & \multirow{5}{*}{w4g128}           & RTN                                  & 47.20\% & 73.88\% & 51.62\% & 74.12\% & 83.58\% & 79.86\% & 73.24\% & 69.07\%          \\
                                    &                                      & GPTQ                                 & 47.00\%       & 73.48\%             & 50.85\%        & 73.06\%        & 83.67\%        & 80.31\%            & 72.50\%          & 68.70\%          \\
                                    &                                      & OmniQuant                            & 48.80\% & 74.19\% & 50.68\% & 73.91\% & 83.79\% & 79.83\% & 73.28\% & \textbf{69.21\%} \\
                                    &                                      & AWQ                                  & 47.00\% & 73.16\% & 50.85\% & 73.82\% & 84.19\% & 79.77\% & 73.32\% & 68.87\%          \\
                                    \rowcolor{orange!10} \cellcolor{white!10}&\cellcolor{white!10}& \ \ +\namebos      & 45.60\%       & 73.24\%             & 50.94\%        & 74.12\%        & 84.28\%        & 79.70\%            & 73.14\%          & 68.72\%          \\ 
\bottomrule
\end{tabular}
}
\end{center}
\vspace{-10pt}
\caption{Weight-only quantization results of Vicuna models on seven 0-shot commonsense QA tasks.
}
\label{apdx-tab:qa}
\end{table*}
\begin{table}[t]
\centering
\resizebox{1\linewidth}{!}{
\begin{tabular}{l|cc}
\toprule
\textbf{Method}    & \textbf{Vicuna-v1.5-7B}   & \textbf{Vicuna-v1.5-13B} \\ \midrule
FP16                  & 5.31                      & 5.52            \\ \midrule
RTN                & 5.18                      & 5.47            \\
                           OmniQuant          & 5.09                      & 5.48            \\
                           AWQ                & 5.22                      & 5.28            \\
                           \rowcolor{orange!10}          
                           \ \ +\nameprompt      & 5.32                      & 5.35            \\
                           \rowcolor{orange!10}       
                           \ \ +\nameprompt+Cal   & \textbf{5.36}             & \textbf{5.50}   \\ 
\bottomrule
\end{tabular}
}
\caption{GPT-4 evaluation of INT4-group128 weight-only quantized Vicuna-v1.5 models on MT-Bench. The scores are on a scale of 10.}
\label{apdx-tab:mt_bench_4bit} 
\vspace{-0.5em}
\end{table}
\paragraph{Weight and Activation Quantization.} For weight and activation quantization of OmniQuant, it is difficult to integrate \namebos\ into training with the learnable equivalent transformation, so we reuse the official checkpoint of LLaMA and LLaMA-2 models. When combining \name\ with OmniQuant, we quantize \name\ to lower bits to avoid additional inference overhead. We do not include OmniQuant results on LLaMA-3 models since the option "--let" is not compatible with GQA (Group Query Attention). For QuaRot~\cite{ashkboos2024quarot}, we reproduce all the results with their official code\footnote{\url{https://github.com/spcl/QuaRot}} using WikiText2 \citep{merity2016wiki} calibration set. We do not quantize \name\ to lower bits since QuaRot adopts a mixed-precision self-attention quantization strategy and can not utilize the integer multiplications for self-attention operations. Therefore, maintaining \name\ in FP16 will not bring any extra inference costs for QuaRot.

\section{Evaluation Details}
\label{apdx-sec:eval}
\paragraph{PPL.}
We evaluate PPL following the new evaluation setting in GPTQ official code\footnote{\url{https://github.com/ist-daslab/gptq}}, except that we substitute the first token of each text segment with \bos token to evaluate the performance of \name.
\paragraph{MMLU.}
We evaluate MMLU following the original MMLU implementation\footnote{\url{https://github.com/hendrycks/test/pull/13}} for 0-shot and 5-shot tasks. We note that when using Vicuna, it is considered more appropriate to fit the input sequences into the Vicuna system prompt. However, the original MMLU implementation does not use the Vicuna system prompt for Vicuna models. In our experiments on Vicuna models, we find that naively fitting the original MMLU prompt into the Vicuna system prompt will harm the final accuracy. Since prompt engineering is out of scope for this paper, we choose to follow the original evaluation setting that does not use the Vicuna system prompt for MMLU evaluation on Vicuna models.

\paragraph{Common Sense Reasoning Tasks.}
For the seven zero-shot common sense reasoning tasks, we adopt the open-sourced lm-evaluation-harness\footnote{\url{https://github.com/EleutherAI/lm-evaluation-harness}} library for evaluation. Similar to PPL evaluation, to assess the performance of \name, we prepend \bos token to the beginning of each input sequence. For the evaluation of Vicuna models, we also follow the evaluation protocol in lm-evaluation-harness and do not use a system prompt.

\paragraph{MT-bench.}
MT-bench employs a GPT-4 model to score the generated content. In our experiments, we find that the scores given by GPT-4 can vary for the same generated content even when the generation temperature of GPT-4 is set to 0. Besides, content generation for the writing and roleplay categories has a relatively high generation temperature of 0.7, which also results in variations in the final score. To faithfully assess the performance of the quantized model and decrease the variations in the final score, we run the content generation process of each model 3 times with random seeds 42, 43, and 44. We report the mean score of three trials as the final score in Table~\ref{tab:mt_bench_3bit} and Table~\ref{apdx-tab:mt_bench_4bit}. Also, we note that GPT-4-Turbo has been shown to be smarter than GPT-4\footnote{\url{https://huggingface.co/spaces/lmsys/chatbot-arena-leaderboard}}, and in our experiments, we find that GPT-4-Turbo can give more stable scores than GPT-4 while having a much lower price. Therefore, we evaluate the generation results on MT-bench with the latest gpt-4-0125-preview API (i.e., GPT-4-Turbo) provided by OpenAI to further reduce variations in the final score.

\section{More Experiment Results}
\label{apdx-sec:results}

\subsection{PPL Results}
\label{apdx-subsec:ppl}
We provide PPL results of LLaMA and LLaMA-2 models on WikiText2 in 
Table~\ref{apdx-tab:ppl}, and PPL results of LLaMA-3 models in Table~\ref{apdx-tab:ppl_llama3}. These results affirm \name's effectiveness in restoring the capabilities of quantized models. Moreover, in Table~\ref{apdx-tab:ppl_backbone}, we conduct experiments on more heterogeneous backbones like OPT and Mistral, which further proves the compatibility of our \name\ with various LLM backbones.

\subsection{MMLU Results}
\label{apdx-subsec:mmlu}
We provide INT3-group128 weight-only quantization results on MMLU in Table~\ref{apdx-tab:mmlu_3bit}, and INT4-group128 weight-only quantization results on MMLU in Table~\ref{apdx-tab:mmlu_4bit}. For INT3-group128 quantization, AWQ+\name\ consistently improves AWQ in every experiment setting and outperforms OmniQuant for nine out of ten settings. For INT4-group128 quantization, AWQ+\name\ leads to relatively less improvement over AWQ compared with INT3-group128 quantization, but still outperforms AWQ in nine out of ten experiment settings, and performs on par with OmniQuant.

\subsection{Commonsense QA Results}
\label{apdx-subsec:qa}
We conduct experiments on seven zero-shot commonsense QA tasks for the Vicuna family with both INT3-group128 and INT4-group128 weight-only quantization. The results are shown in Table~\ref{apdx-tab:qa}. For INT3-group128 quantization, AWQ+\name\ significantly surpasses all baselines in four out of five experiment settings. For INT4-group128 quantization, AWQ+\name\ improves AWQ and outperforms OmniQuant in four out of five experiment settings, demonstrating the superiority of \name.


\subsection{MT-Bench Results}
\label{apdx-subsec:mt-bench}
We provide INT4-group128 quantization results on MT-bench in Table~\ref{apdx-tab:mt_bench_4bit}. As can be seen, \name\ leads to an average increase of 0.09 in the final score. Remarkably, with trainable \name, AWQ even matches the full-precision model under INT4 quantization, while all other methods clearly lag behind the full-precision model.

\section{Effectiveness of Calibrating \name}
\label{apdx-sec:gbias}
We conduct more experiments on MT-Bench to further demonstrate the effectiveness of calibrating \name. We adopt a commonly used fine-tuning method for quantized models that tunes the quantization bias term (used in non-symmetric quantization) in every quantization group as a baseline method, termed "group bias tuning". Both "group bias tuning" and "calibrating \name" are further tuned based on "AWQ+\nameprompt". We use the same calibration set containing 128 samples and train 20 epochs for a fair comparison. As shown in Table~\ref{apdx-tab:gbias}, although calibrating \name\ uses fewer trainable parameters, it still achieves better or comparable results compared with group bias tuning, demonstrating the effectiveness of calibrating \name. Also, we note that "calibrating \name" can be adopted for any quantization setting, while "group bias tuning" is only suitable for non-symmetric and group-wise quantization, making our proposed method a more versatile calibration strategy for quantized models.

\section{Adapting \name\ for Activation Quantization}
\label{apdx-sec:wa}
It is non-trivial to integrate \name\ into activation quantization. 
For activation quantization, the whole KV cache needs to be quantized to low bits to exploit integer multiplications in self-attention, which contradicts our idea of keeping pivot tokens' KV cache intact. However, as shown in Table~\ref{apdx-tab:kv_stat}, the distribution of the pivot tokens' KV cache is much smoother than that of the non-pivot tokens' KV cache, which implies that \name\ is amenable to quantization. Therefore, we adopt a straightforward solution to adapt \name\ for activation quantization that directly quantizes \name\ to lower bits with RTN. As shown in Table~\ref{apdx-tab:wa_16vs4}, quantizing \name\ incurs minimal accuracy loss. For example, quantizing \name\ to 4 bits only results in an average PPL increase of 0.05 on WikiText2 compared with full-precision \name, which is negligible.

\begin{table}[t]
\centering
\resizebox{1\linewidth}{!}{
\begin{tabular}{l|c|c}
\toprule
\textbf{Method}    & \textbf{INT3-group128}   & \textbf{INT4-group128} \\ \midrule
AWQ                      & 5.17 & 5.28            \\
\ \ +\nameprompt                      & 5.34 & 5.35            \\
\rowcolor{orange!10}\ \ +\nameprompt+gbias      & 5.31 & 5.47            \\
\rowcolor{orange!10}\ \ +\nameprompt+Cal   & \textbf{5.44} & \textbf{5.50}   \\
\bottomrule
\end{tabular}
}
\caption{Evaluation of different calibration methods on MT-bench. "gbias" denotes group bias tuning and "Cal" denotes calibrating \name.}
\label{apdx-tab:gbias} 
\vspace{-0.5em}
\end{table}
\begin{table*}[!h]
\begin{center}
\resizebox{\linewidth}{!}{
\begin{tabular}{l|cc|cc|cc|cc}
\toprule
\multicolumn{1}{c|}{\multirow{2}{*}{\textbf{Method}}} & \multicolumn{2}{c|}{\textbf{Pivot K Cache}} & \multicolumn{2}{c|}{\textbf{Pivot V Cache}} & \multicolumn{2}{c|}{\textbf{Non-pivot K Cache}} & \multicolumn{2}{c}{\textbf{Non-pivot V Cache}}                        \\  
\cmidrule{2-9} \multicolumn{1}{c|}{}                       & \textbf{AbsMax}    & \textbf{Std}           & \textbf{AbsMax}    & \textbf{Std} & \textbf{AbsMax}    & \textbf{Std} & \textbf{AbsMax}    & \textbf{Std}\\ \midrule
LLaMA-7B & 3.15 & 0.38 & 0.63 & 0.04 & 13.91 & 1.58 & 2.34 & 0.46           \\
LLaMA-13B & 3.02 & 0.35 & 0.73 & 0.05 & 13.69 & 1.56 & 2.62 & 0.49           \\
LLaMA-2-7B & 2.76 & 0.30 & 0.79 & 0.05 & 14.28 & 1.65 & 2.23 & 0.42           \\
LLaMA-2-13B & 2.73 & 0.27 & 0.75 & 0.05 & 14.60 & 1.62 & 2.57 & 0.44           \\
LLaMA-3-8B & 3.30 & 0.37 & 0.57 & 0.03 & 15.86 & 2.19 & 1.54 & 0.27           \\
\bottomrule
\end{tabular}
}
\end{center}
\vspace{-10pt}
\caption{The statistical results of pivot tokens' and non-pivot tokens' KV cache. The maximum absolute value and standard deviation are calculated on a sequence of length 1024 and averaged over all layers.}
\label{apdx-tab:kv_stat}
\end{table*}
\begin{table*}[!h]
\begin{center}
\resizebox{\linewidth}{!}{
\begin{tabular}{l|cc|cc|cc|cc}
\toprule
\multicolumn{1}{c|}{\multirow{2}{*}{\textbf{Method}}} & \multicolumn{2}{c|}{\textbf{LLaMA-7B}} & \multicolumn{2}{c|}{\textbf{LLaMA-13B}} & \multicolumn{2}{c|}{\textbf{LLaMA-2-7B}} & \multicolumn{2}{c}{\textbf{LLaMA-2-13B}}                       \\  
\cmidrule{2-9} \multicolumn{1}{c|}{}                       & \textbf{C4}    & \textbf{WikiText2}           & \textbf{C4}    & \textbf{WikiText2} & \textbf{C4}    & \textbf{WikiText2} & \textbf{C4}    & \textbf{WikiText2}\\ \midrule
OmniQuant & 17.03 & 12.17 & 15.65 & 11.16 & 21.40 & 14.74 & 16.24 & 12.28\\
\rowcolor{orange!10} \ \ +\namebos\ (FP16) & 16.26 & 11.30 & 13.89 & 10.00 & 19.97 & 13.61 & 15.77 & 10.94\\
\rowcolor{orange!10} \ \ +\namebos & 16.24 & 11.32 & 13.87 & 10.04 & 20.01 & 13.70 & 15.91 & 11.00\\
\bottomrule
\end{tabular}
}
\end{center}
\vspace{-10pt}
\caption{The effect of quantizing \name\ to lower bits. We show the INT4 weight and activation quantization results of LLaMA models on C4 and WikiText2 datasets. \namebos\ (FP16) indicates keeping \name\ in 16 bits, which incurs extra inference costs. \namebos\ indicates quantizing \name\ to lower bits (i.e., 4 bits).
}
\label{apdx-tab:wa_16vs4}
\end{table*}
\begin{table*}[t]
\centering
\begin{tabular}{c|c}
\toprule
Model & Download URL \\
\midrule
LLaMA-2-7B & \url{https://huggingface.co/meta-llama/Llama-2-7b} \\
LLaMA-2-13B & \url{https://huggingface.co/meta-llama/Llama-2-13b} \\
LLaMA-2-70B & \url{https://huggingface.co/meta-llama/Llama-2-70b} \\
LLaMA-3-8B & \url{https://huggingface.co/meta-llama/Meta-Llama-3-8B} \\
LLaMA-3-70B & \url{https://huggingface.co/meta-llama/Meta-Llama-3-70B} \\
Vicuna-v1.3-7B & \url{https://huggingface.co/lmsys/vicuna-7b-v1.3} \\
Vicuna-v1.3-13B & \url{https://huggingface.co/lmsys/vicuna-13b-v1.3} \\
Vicuna-v1.3-33B & \url{https://huggingface.co/lmsys/vicuna-33b-v1.3} \\
Vicuna-v1.5-7B & \url{https://huggingface.co/lmsys/vicuna-7b-v1.5} \\
Vicuna-v1.5-13B & \url{https://huggingface.co/lmsys/vicuna-13b-v1.5} \\
\bottomrule
\end{tabular}
\caption{Download links to officially released LLMs.}
\label{apdx-tab:links}
\end{table*}

\section{Links to Officially Released LLMs}
\label{apdx-sec:links}
We provide download links to some officially released LLMs used in our experiments in Table~\ref{apdx-tab:links}.

\newpage
\captionsetup[subfloat]{labelsep=none,format=plain,labelformat=empty,justification=centering}
\captionsetup[subfloat]{labelsep=none,format=plain,labelformat=empty,justification=centering}
\begin{figure*}[t]
\centering
\subfloat[(a) Output activations of LLaMA-13B Layer 8]{
        \includegraphics[width=0.24\linewidth]{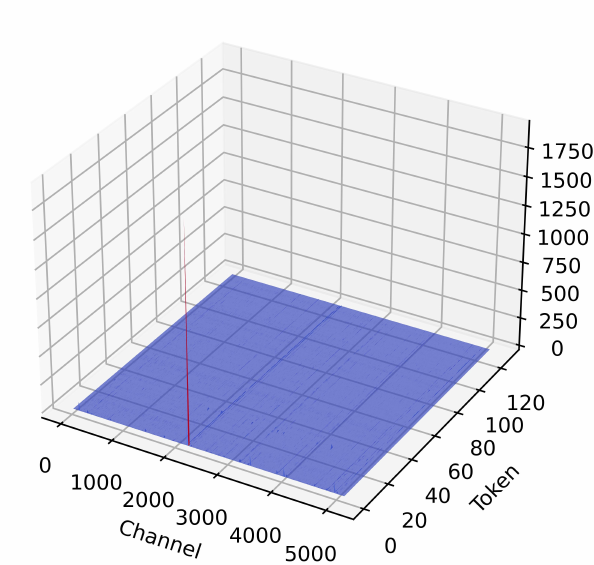}
    }
\subfloat[(b) Output activations of LLaMA-13B Layer 16]{
        \includegraphics[width=0.24\linewidth]{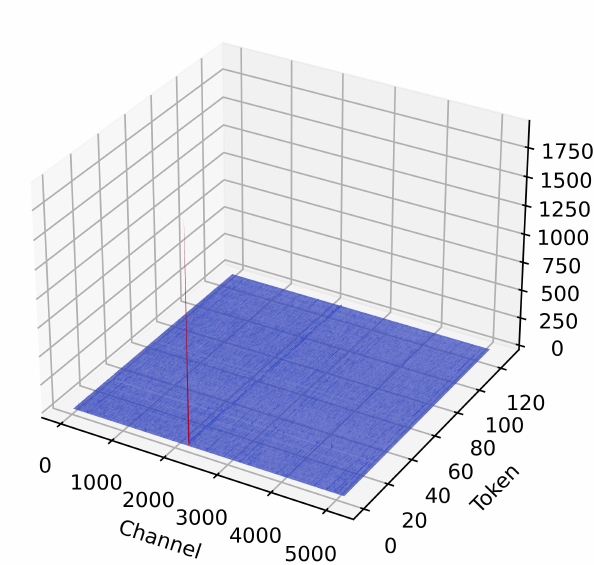}
    }
\subfloat[(c) Output activations of LLaMA-13B Layer 24]{
        \includegraphics[width=0.24\linewidth]{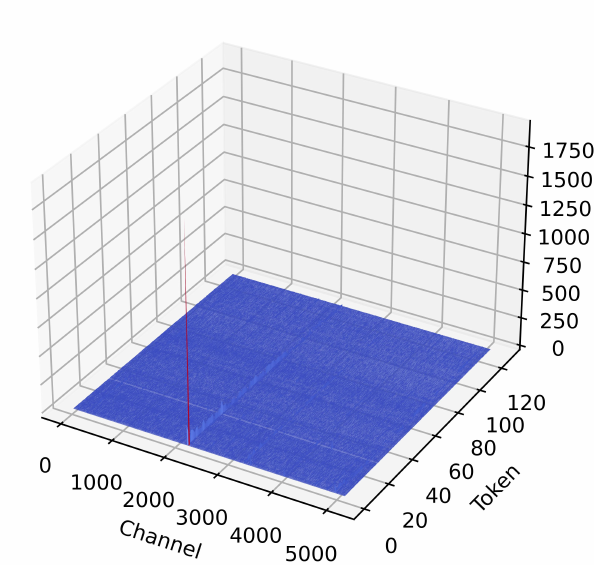}
    }
\subfloat[(d) Output activations of LLaMA-13B Layer 32]{
        \includegraphics[width=0.24\linewidth]{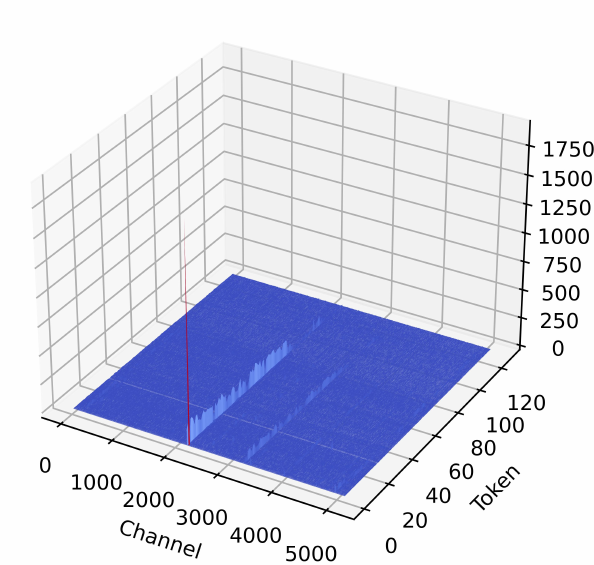}
    }
\vspace{0.25cm}
\subfloat[(e) Attention map of \\LLaMA-13B Layer 8]{
        \includegraphics[width=0.24\linewidth]{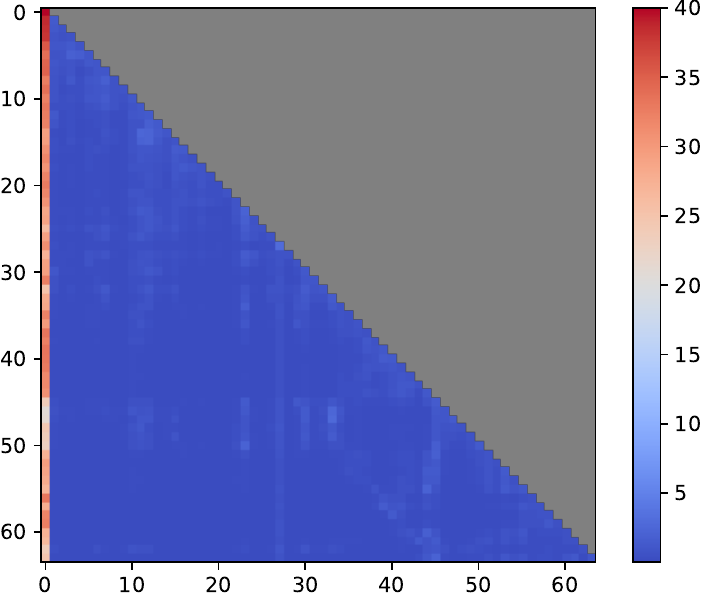}
    }
\subfloat[(f) Attention map of \\LLaMA-13B Layer 16]{
        \includegraphics[width=0.24\linewidth]{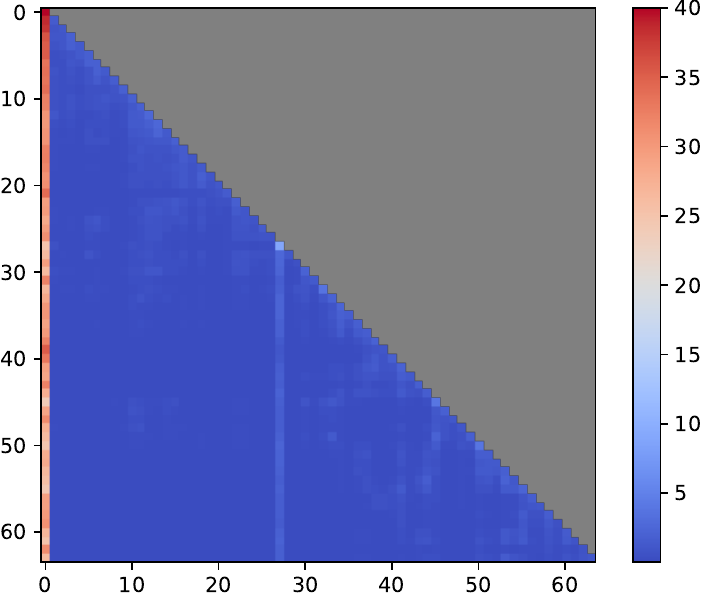}
    }
\subfloat[(g) Attention map of \\LLaMA-13B Layer 24]{
        \includegraphics[width=0.24\linewidth]{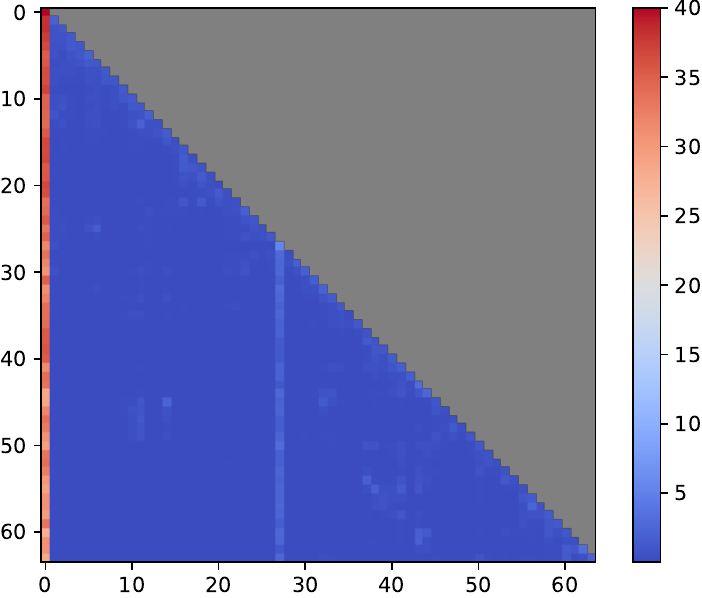}
    }
\subfloat[(h) Attention map of \\LLaMA-13B Layer 32]{
        \includegraphics[width=0.24\linewidth]{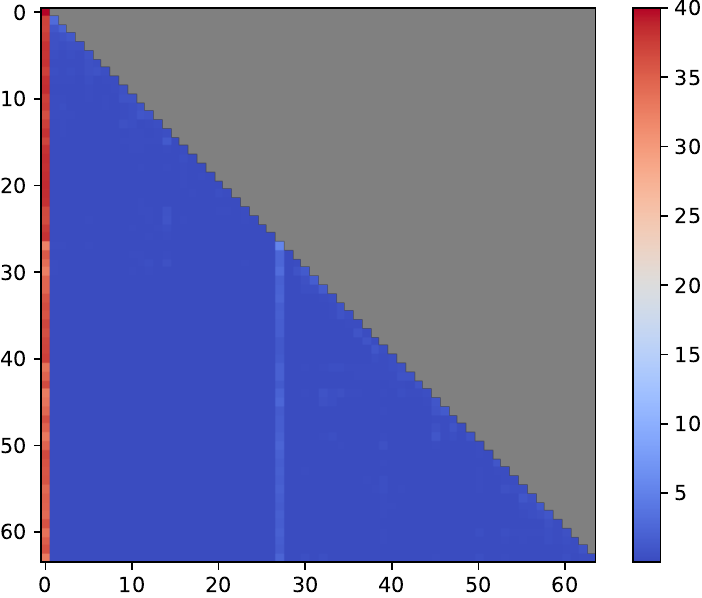}
    }
	\caption{Magnitude of the output activations and attention map in LLaMA-13B.
 }
\end{figure*}
\captionsetup[subfloat]{labelsep=none,format=plain,labelformat=empty,justification=centering}
\begin{figure*}[t]
\centering
\subfloat[(a) Output activations of LLaMA-30B Layer 8]{
        \includegraphics[width=0.24\linewidth]{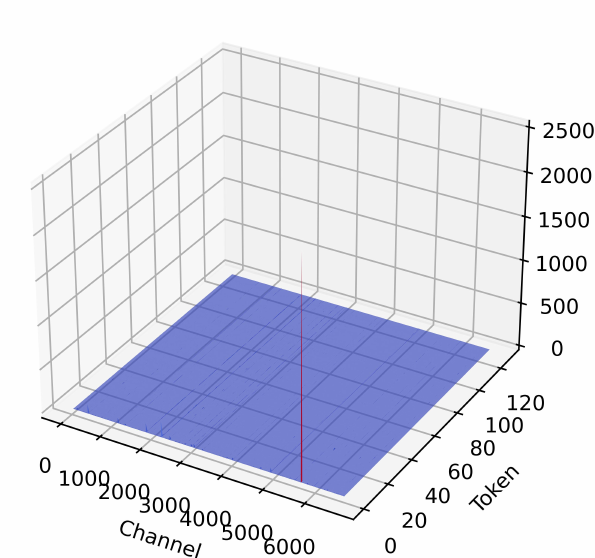}
    }
\subfloat[(b) Output activations of LLaMA-30B Layer 24]{
        \includegraphics[width=0.24\linewidth]{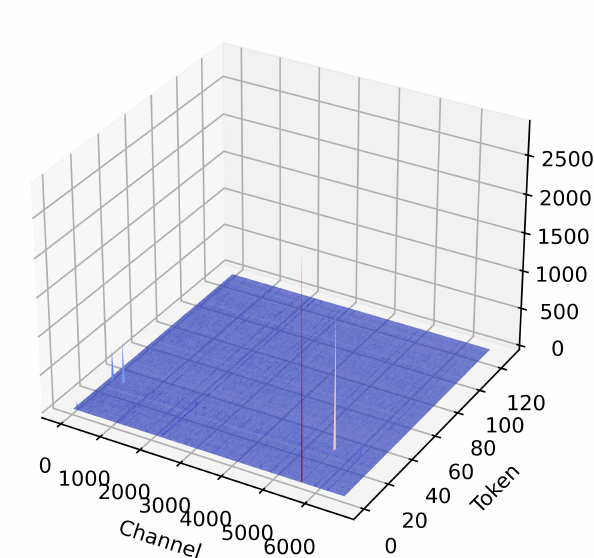}
    }
\subfloat[(c) Output activations of LLaMA-30B Layer 40]{
        \includegraphics[width=0.24\linewidth]{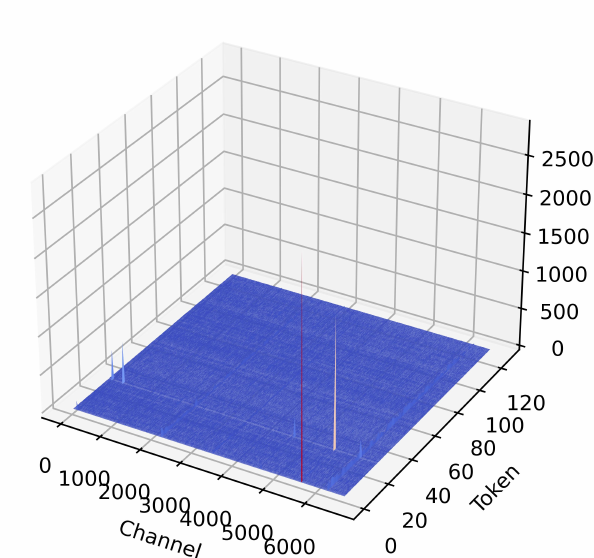}
    }
\subfloat[(d) Output activations of LLaMA-30B Layer 56]{
        \includegraphics[width=0.24\linewidth]{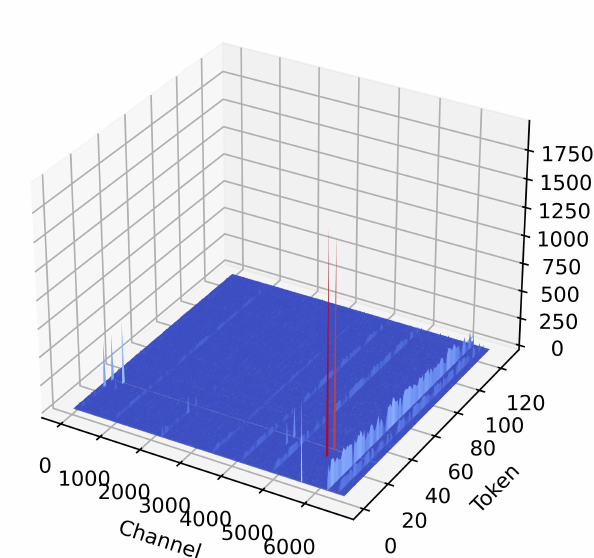}
    }
\vspace{0.25cm}
\subfloat[(e) Attention map of \\LLaMA-30B Layer 8]{
        \includegraphics[width=0.24\linewidth]{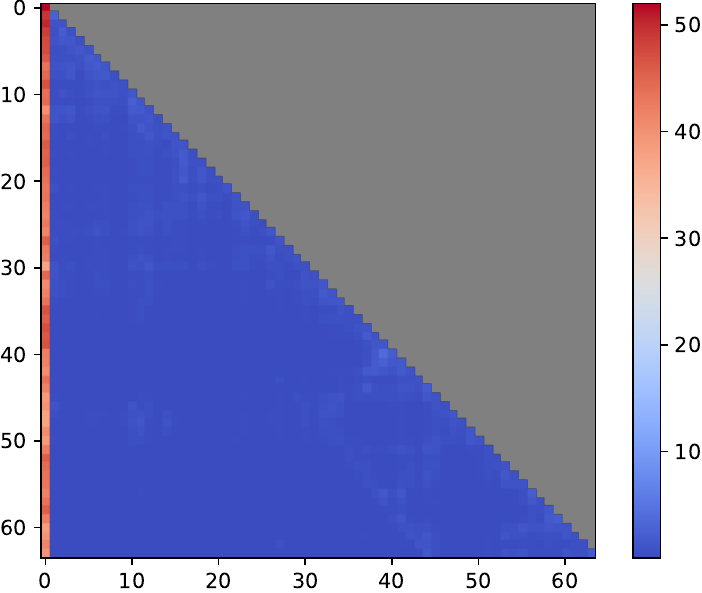}
    }
\subfloat[(f) Attention map of \\LLaMA-30B Layer 24]{
        \includegraphics[width=0.24\linewidth]{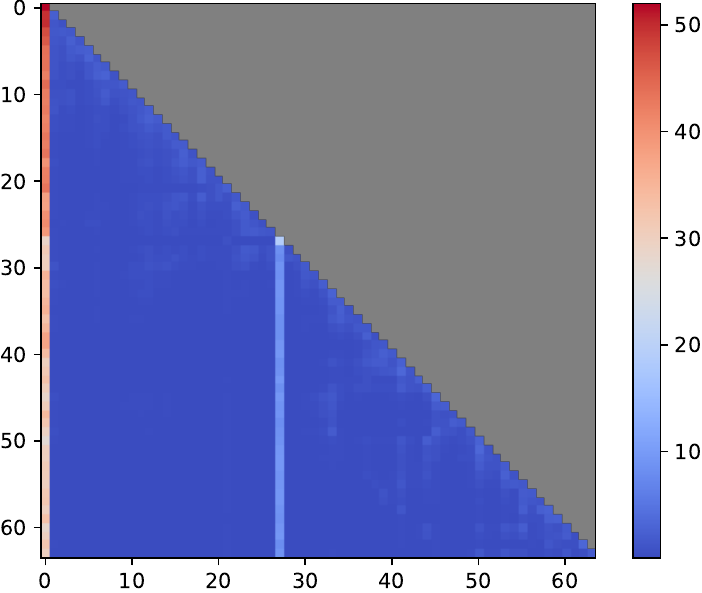}
    }
\subfloat[(g) Attention map of \\LLaMA-30B Layer 40]{
        \includegraphics[width=0.24\linewidth]{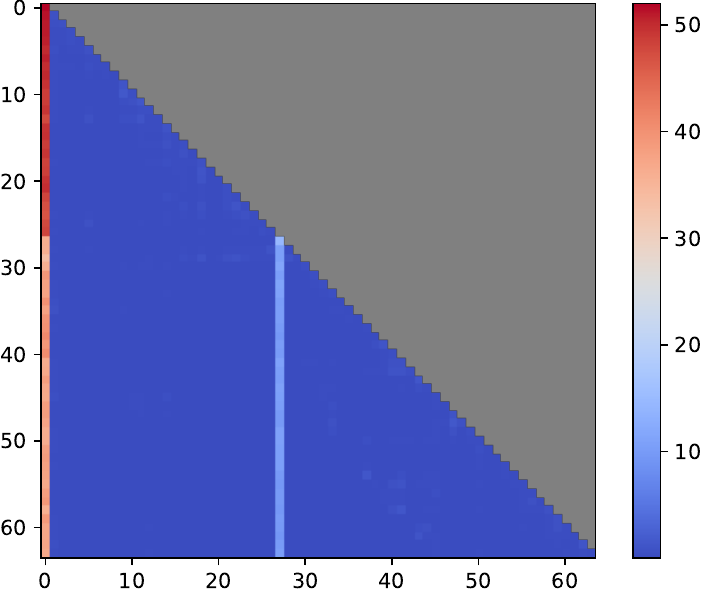}
    }
\subfloat[(h) Attention map of \\LLaMA-30B Layer 56]{
        \includegraphics[width=0.24\linewidth]{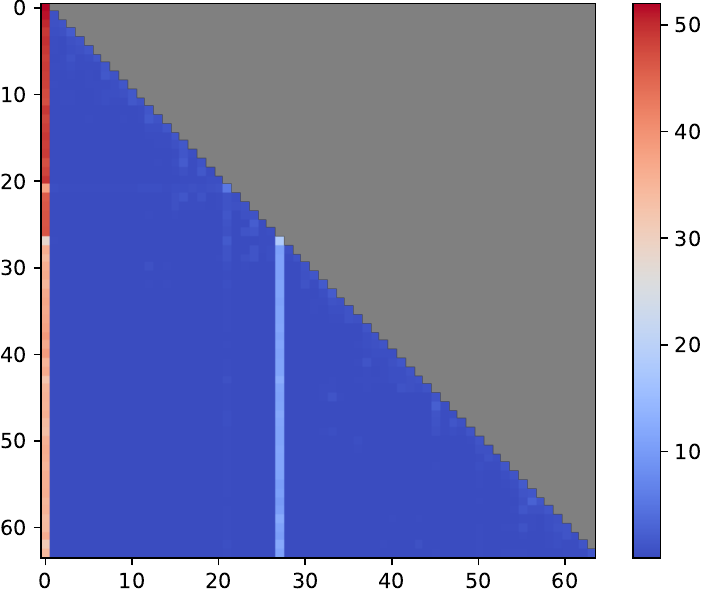}
    }
	\caption{Magnitude of the output activations and attention map in LLaMA-30B.
 }
\end{figure*}
\captionsetup[subfloat]{labelsep=none,format=plain,labelformat=empty,justification=centering}
\begin{figure*}[t]
\centering
\subfloat[(a) Output activations of LLaMA-65B Layer 16]{
        \includegraphics[width=0.24\linewidth]{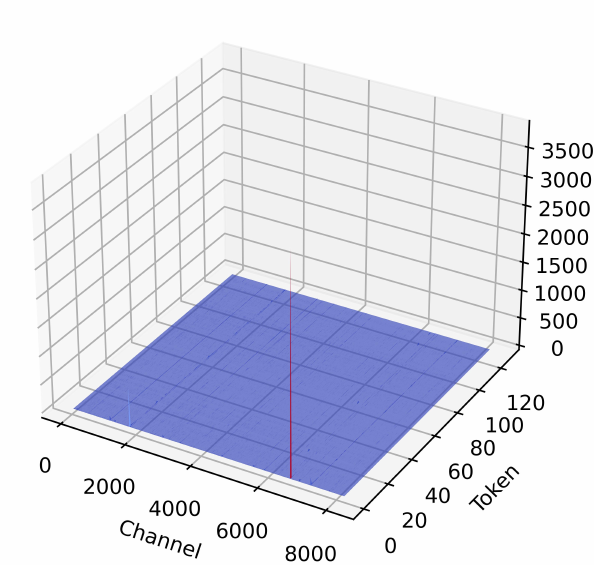}
    }
\subfloat[(b) Output activations of LLaMA-65B Layer 32]{
        \includegraphics[width=0.24\linewidth]{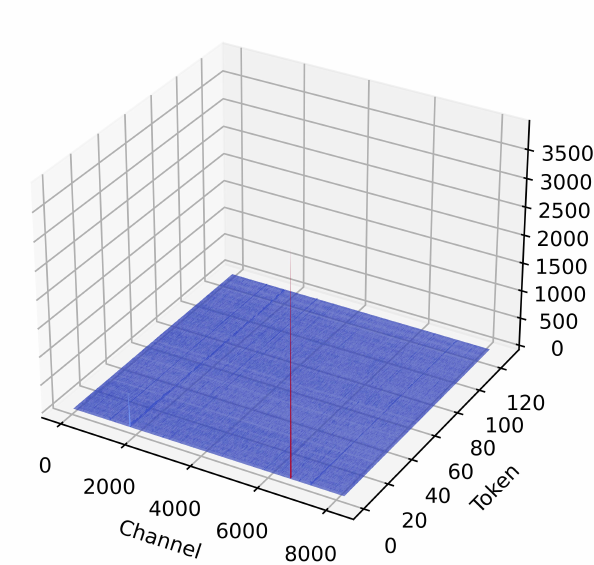}
    }
\subfloat[(c) Output activations of LLaMA-65B Layer 48]{
        \includegraphics[width=0.24\linewidth]{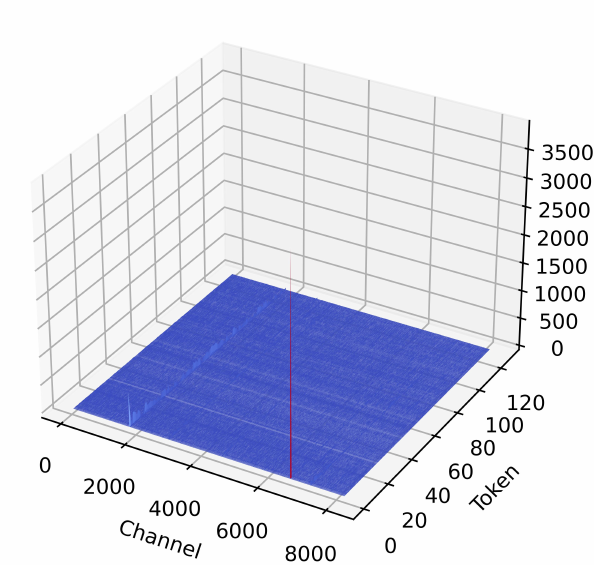}
    }
\subfloat[(d) Output activations of LLaMA-65B Layer 64]{
        \includegraphics[width=0.24\linewidth]{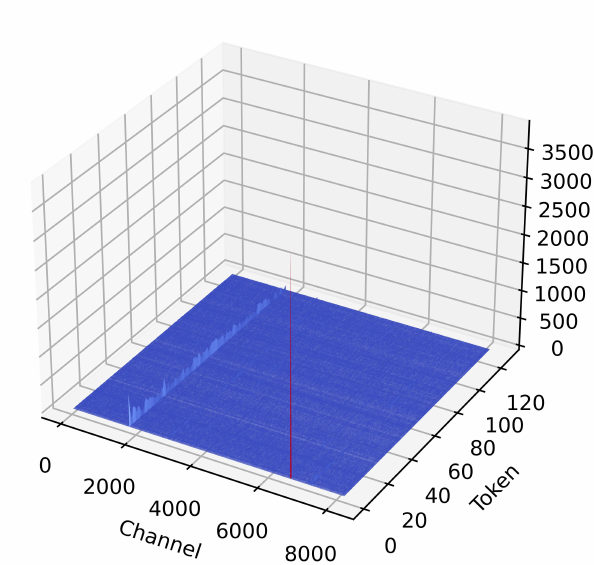}
    }
\vspace{0.25cm}
\subfloat[(e) Attention map of \\LLaMA-65B Layer 16]{
        \includegraphics[width=0.24\linewidth]{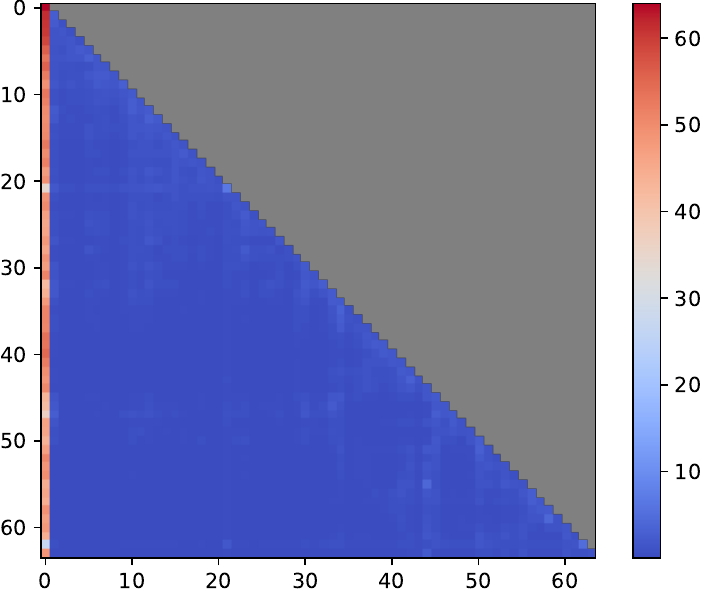}
    }
\subfloat[(f) Attention map of \\LLaMA-65B Layer 32]{
        \includegraphics[width=0.24\linewidth]{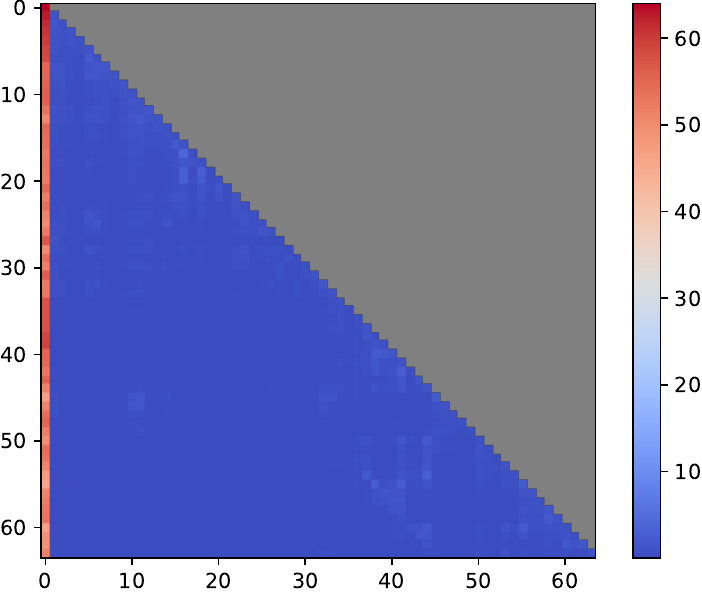}
    }
\subfloat[(g) Attention map of \\LLaMA-65B Layer 48]{
        \includegraphics[width=0.24\linewidth]{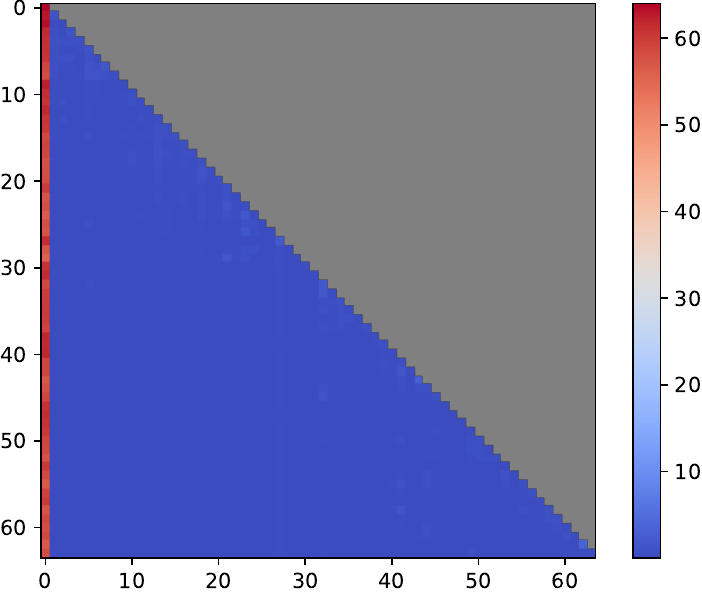}
    }
\subfloat[(h) Attention map of \\LLaMA-65B Layer 64]{
        \includegraphics[width=0.24\linewidth]{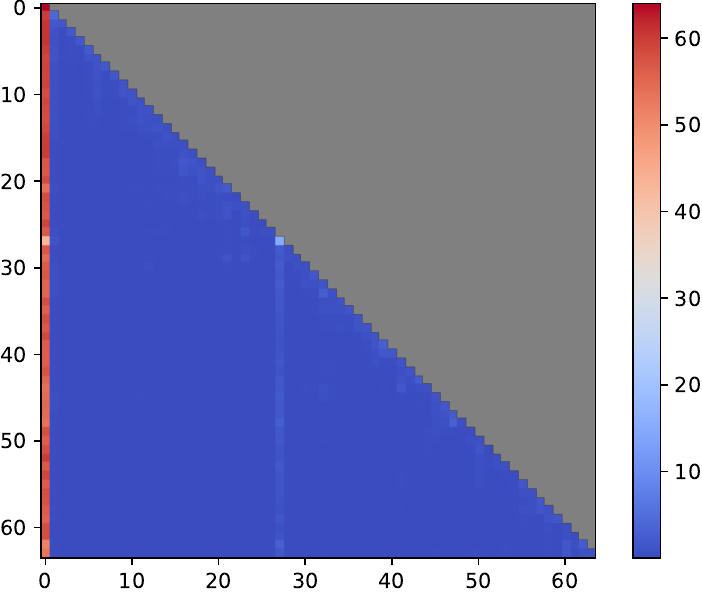}
    }
	\caption{Magnitude of the output activations and attention map in LLaMA-65B.
 }
\end{figure*}
\captionsetup[subfloat]{labelsep=none,format=plain,labelformat=empty,justification=centering}
\begin{figure*}[t]
\centering
\subfloat[(a) Output activations of LLaMA-2-7B Layer 0]{
        \includegraphics[width=0.24\linewidth]{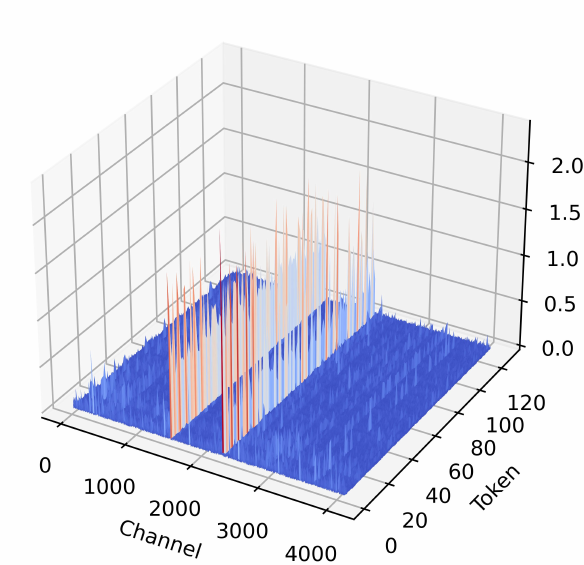}
    }
\subfloat[(b) Output activations of LLaMA-2-7B Layer 8]{
        \includegraphics[width=0.24\linewidth]{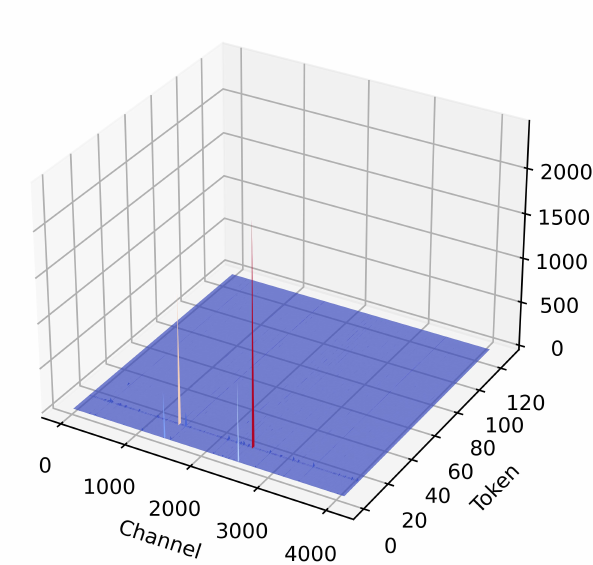}
    }
\subfloat[(c) Output activations of LLaMA-2-7B Layer 16]{
        \includegraphics[width=0.24\linewidth]{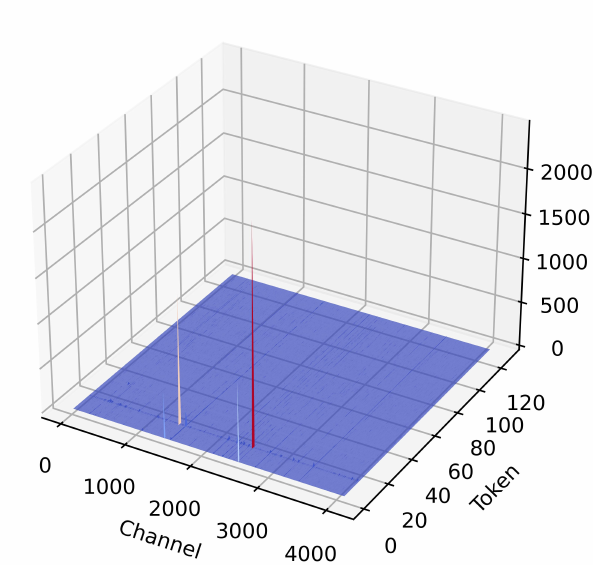}
    }
\subfloat[(d) Output activations of LLaMA-2-7B Layer 24]{
        \includegraphics[width=0.24\linewidth]{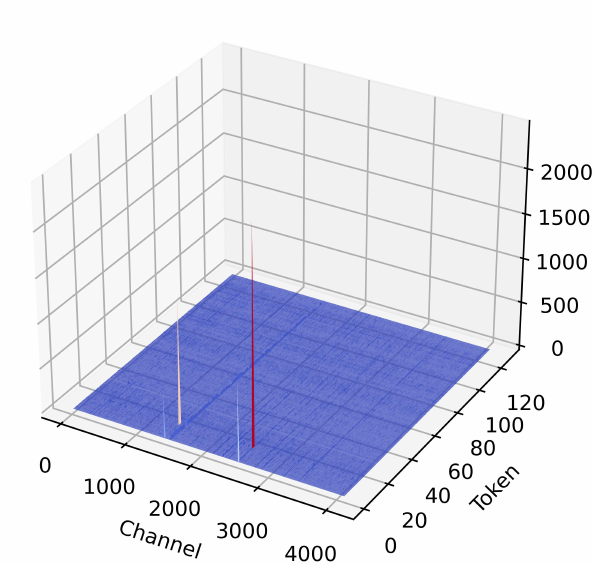}
    }
\vspace{0.25cm}
\subfloat[(e) Attention map of \\LLaMA-2-7B Layer 0]{
        \includegraphics[width=0.24\linewidth]{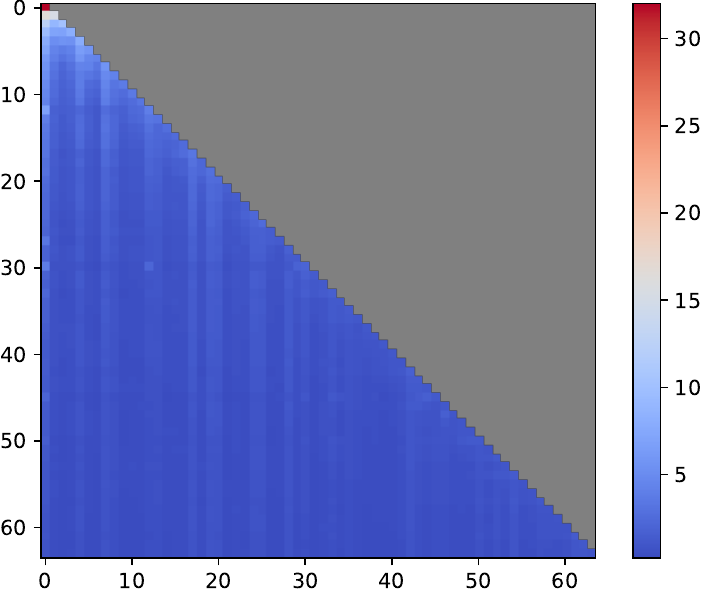}
    }
\subfloat[(f) Attention map of \\LLaMA-2-7B Layer 8]{
        \includegraphics[width=0.24\linewidth]{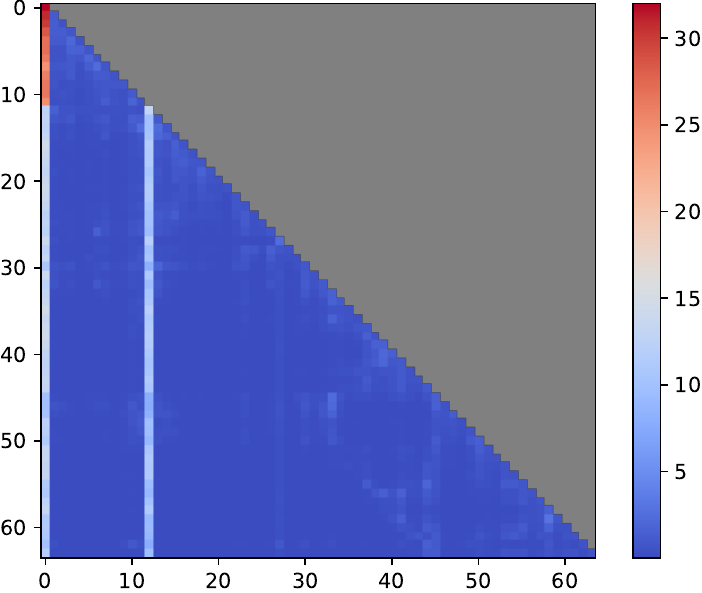}
    }
\subfloat[(g) Attention map of \\LLaMA-2-7B Layer 16]{
        \includegraphics[width=0.24\linewidth]{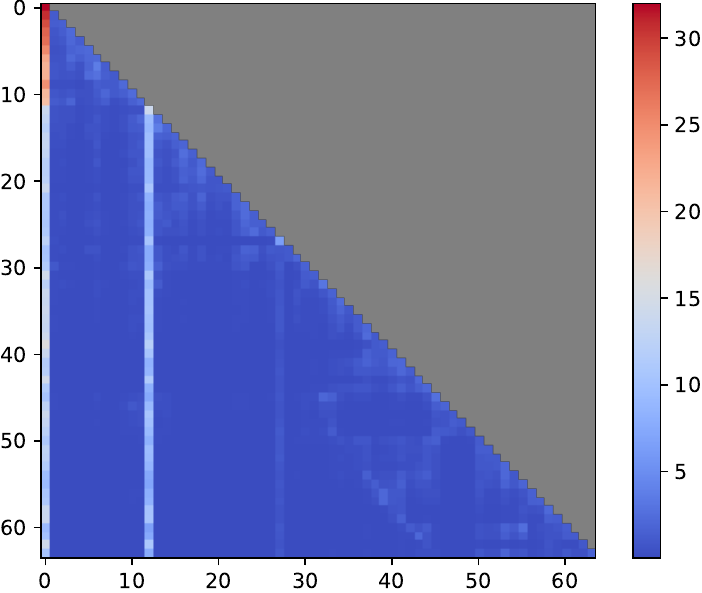}
    }
\subfloat[(h) Attention map of \\LLaMA-2-7B Layer 24]{
        \includegraphics[width=0.24\linewidth]{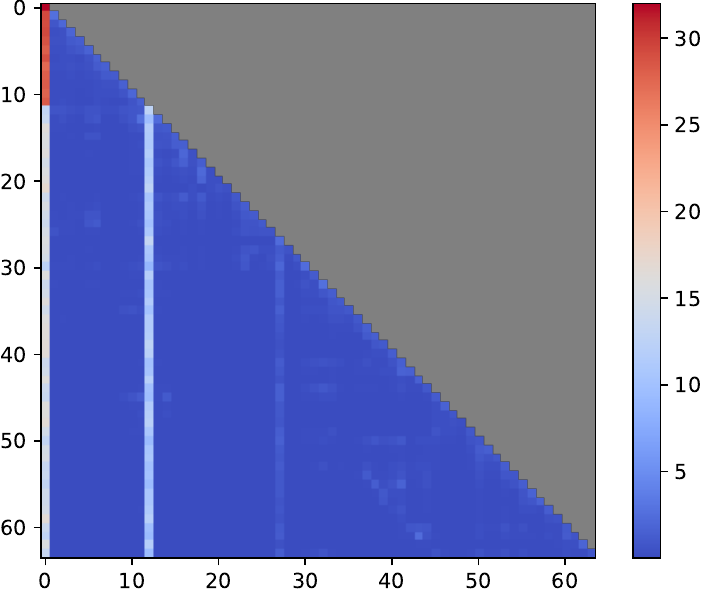}
    }
	\caption{Magnitude of the output activations and attention map in LLaMA-2-7B.
 }
\end{figure*}
\captionsetup[subfloat]{labelsep=none,format=plain,labelformat=empty,justification=centering}
\begin{figure*}[t]
\centering
\subfloat[(a) Output activations of LLaMA-2-13B Layer 8]{
        \includegraphics[width=0.24\linewidth]{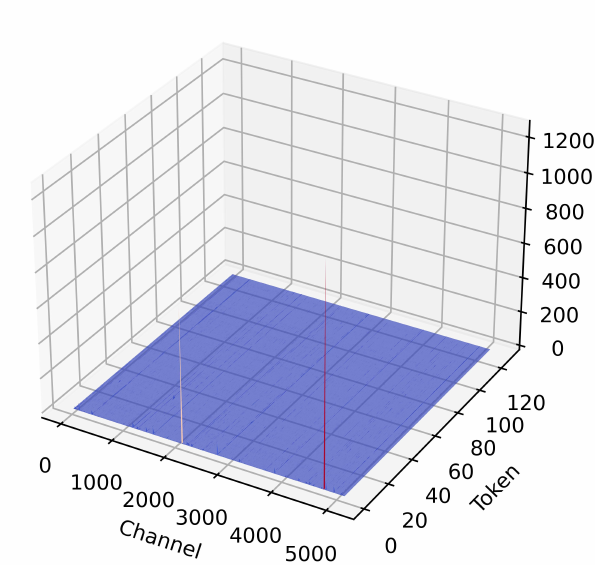}
    }
\subfloat[(b) Output activations of LLaMA-2-13B Layer 16]{
        \includegraphics[width=0.24\linewidth]{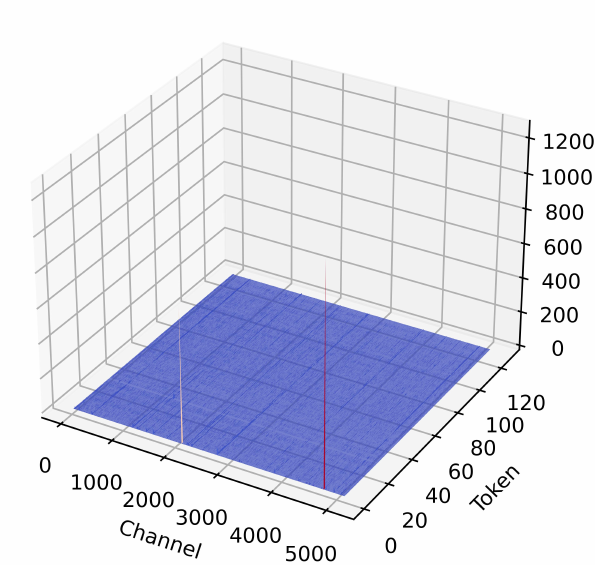}
    }
\subfloat[(c) Output activations of LLaMA-2-13B Layer 24]{
        \includegraphics[width=0.24\linewidth]{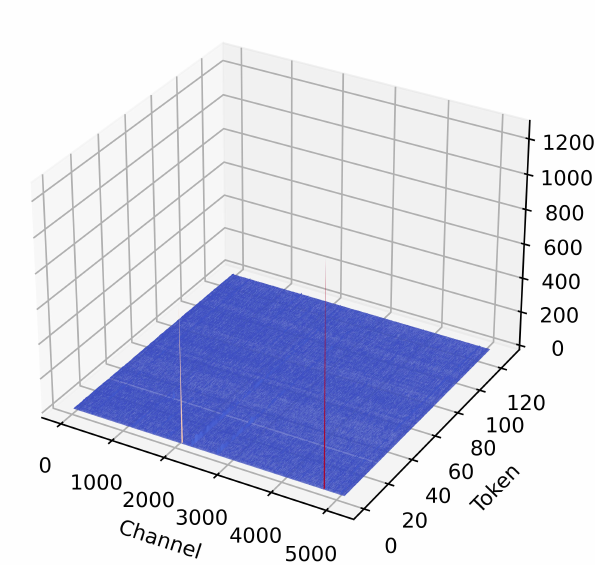}
    }
\subfloat[(d) Output activations of LLaMA-2-13B Layer 32]{
        \includegraphics[width=0.24\linewidth]{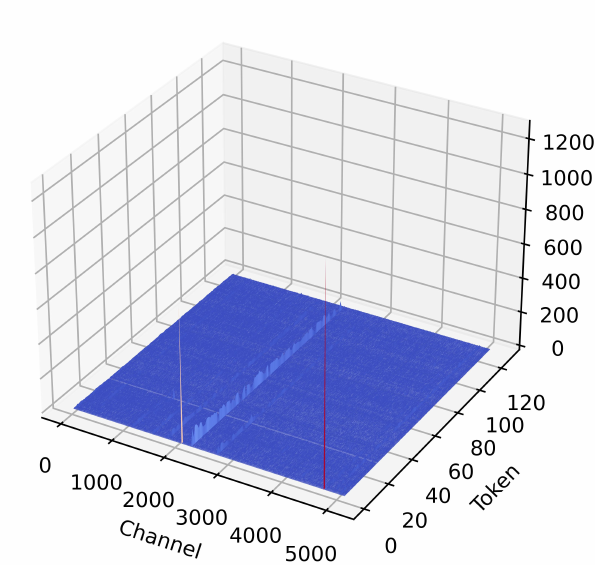}
    }
\vspace{0.25cm}
\subfloat[(e) Attention map of \\LLaMA-2-13B Layer 8]{
        \includegraphics[width=0.24\linewidth]{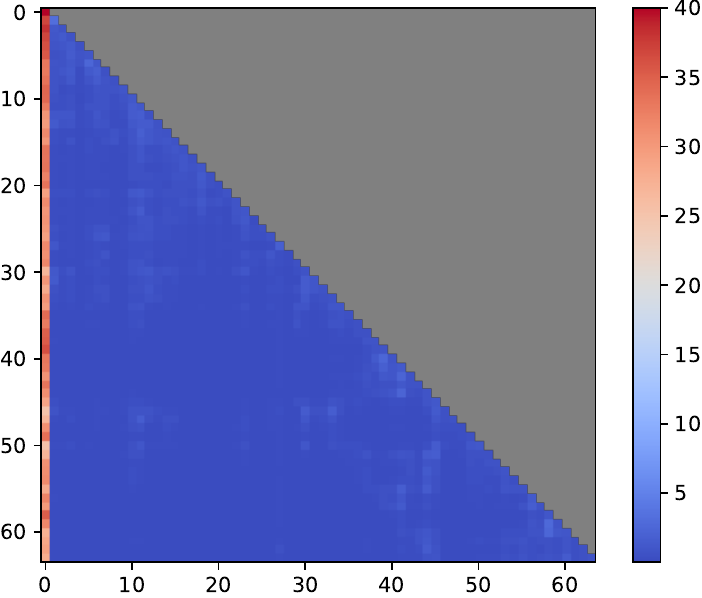}
    }
\subfloat[(f) Attention map of \\LLaMA-2-13B Layer 16]{
        \includegraphics[width=0.24\linewidth]{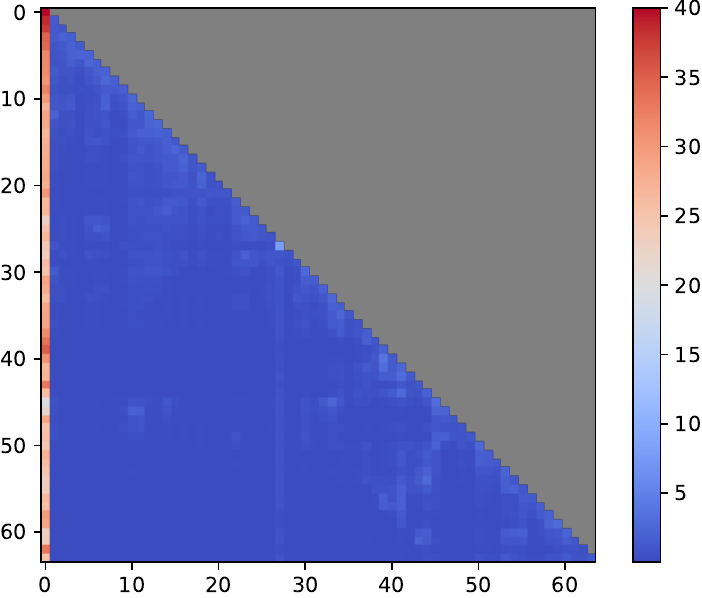}
    }
\subfloat[(g) Attention map of \\LLaMA-2-13B Layer 24]{
        \includegraphics[width=0.24\linewidth]{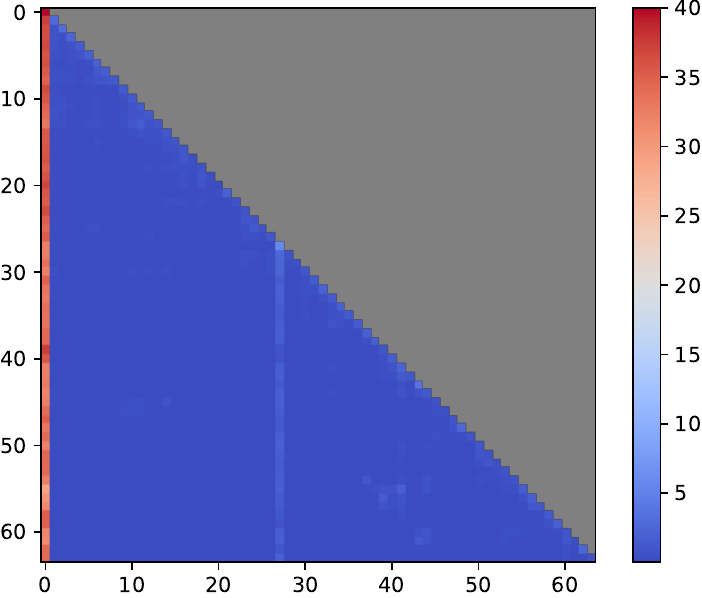}
    }
\subfloat[(h) Attention map of \\LLaMA-2-13B Layer 32]{
        \includegraphics[width=0.24\linewidth]{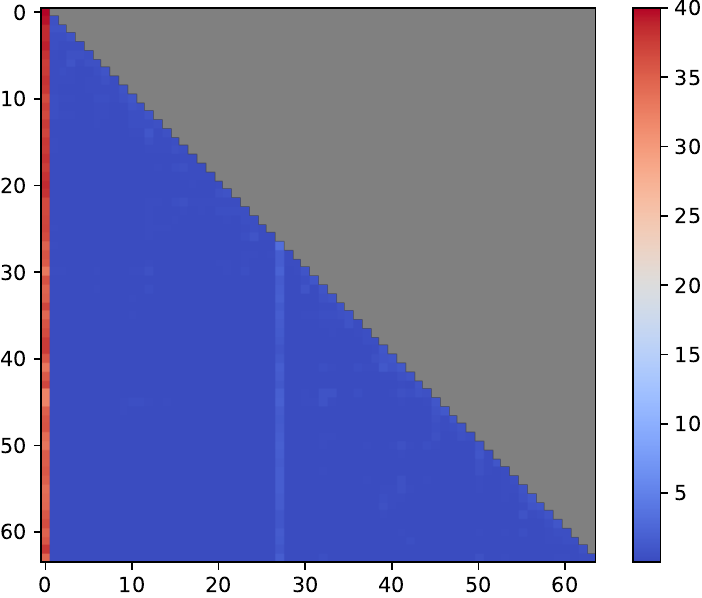}
    }
	\caption{Magnitude of the output activations and attention map in LLaMA-2-13B.
 }
\end{figure*}
\captionsetup[subfloat]{labelsep=none,format=plain,labelformat=empty,justification=centering}
\captionsetup[subfloat]{labelsep=none,format=plain,labelformat=empty,justification=centering}
\begin{figure*}[t]
\centering
\subfloat[(a) Output activations of LLaMA-3-8B Layer 0]{
        \includegraphics[width=0.24\linewidth]{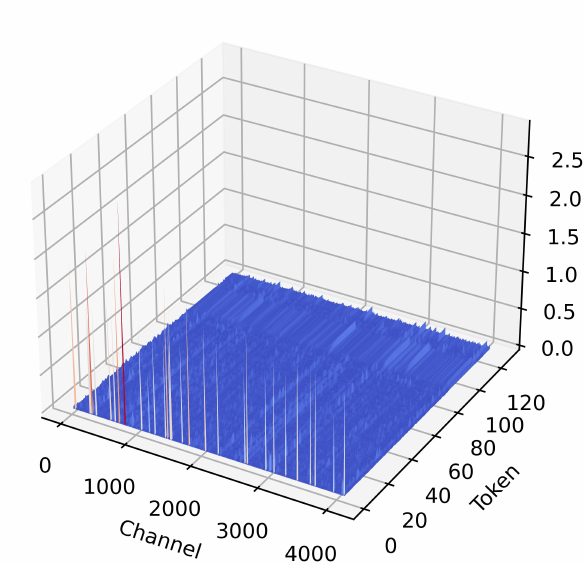}
    }
\subfloat[(b) Output activations of LLaMA-3-8B Layer 8]{
        \includegraphics[width=0.24\linewidth]{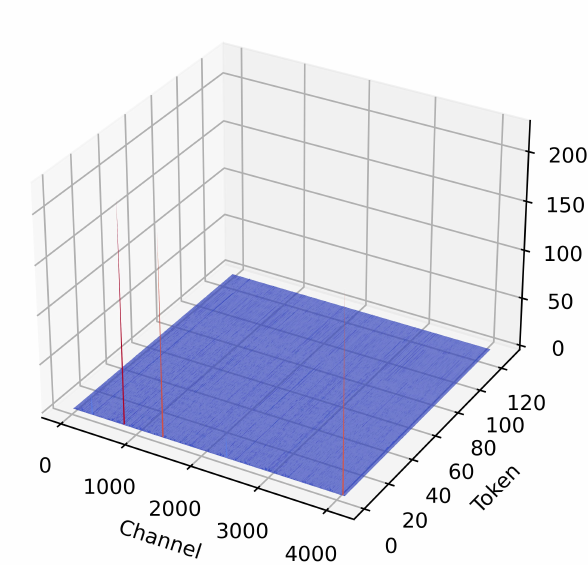}
    }
\subfloat[(c) Output activations of LLaMA-3-8B Layer 16]{
        \includegraphics[width=0.24\linewidth]{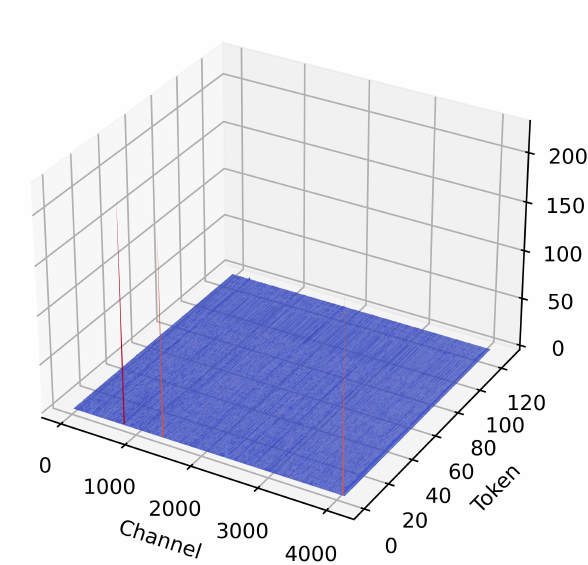}
    }
\subfloat[(d) Output activations of LLaMA-3-8B Layer 24]{
        \includegraphics[width=0.24\linewidth]{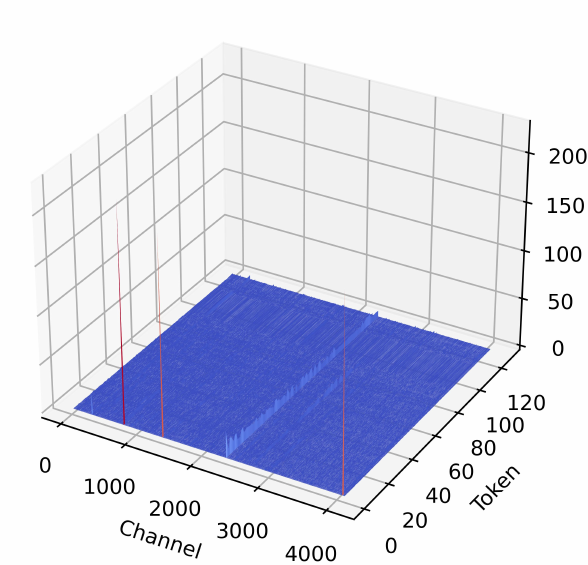}
    }
\vspace{0.25cm}
\subfloat[(e) Attention map of \\LLaMA-3-8B Layer 0]{
        \includegraphics[width=0.24\linewidth]{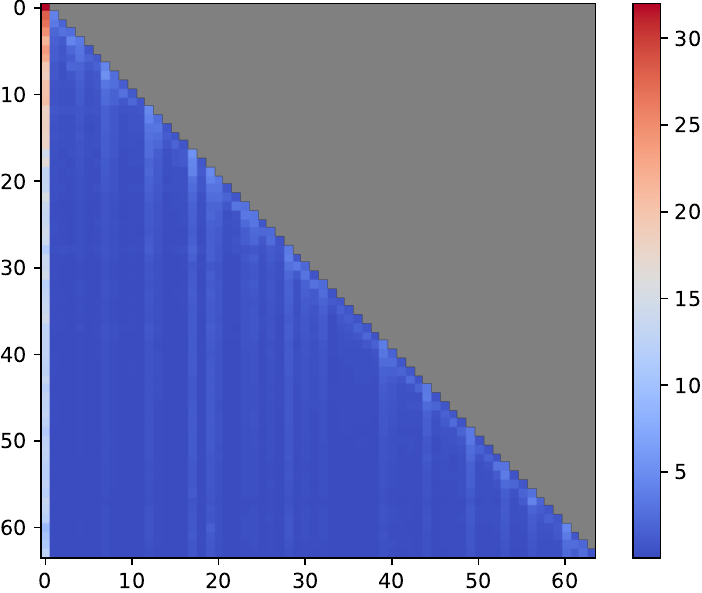}
    }
\subfloat[(f) Attention map of \\LLaMA-3-8B Layer 8]{
        \includegraphics[width=0.24\linewidth]{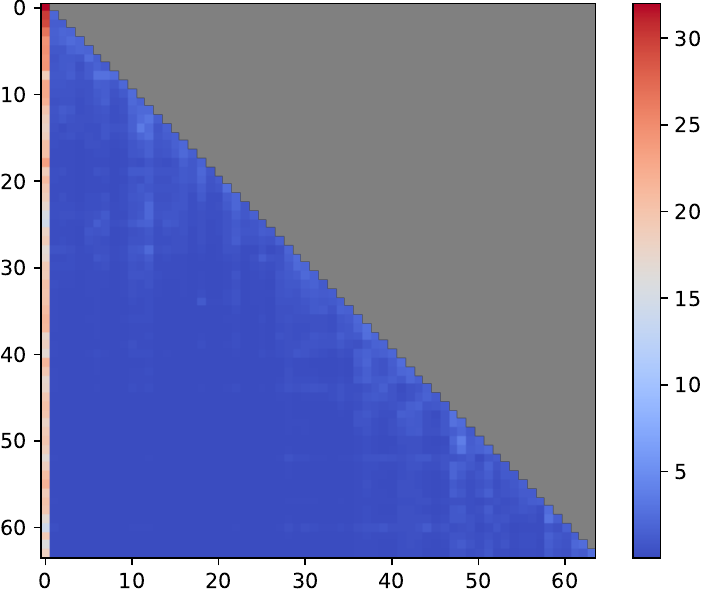}
    }
\subfloat[(g) Attention map of \\LLaMA-3-8B Layer 16]{
        \includegraphics[width=0.24\linewidth]{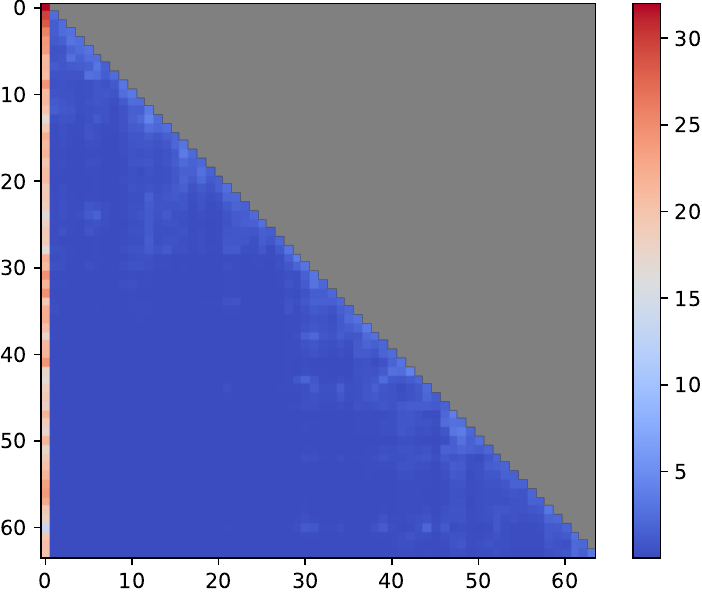}
    }
\subfloat[(h) Attention map of \\LLaMA-3-8B Layer 24]{
        \includegraphics[width=0.24\linewidth]{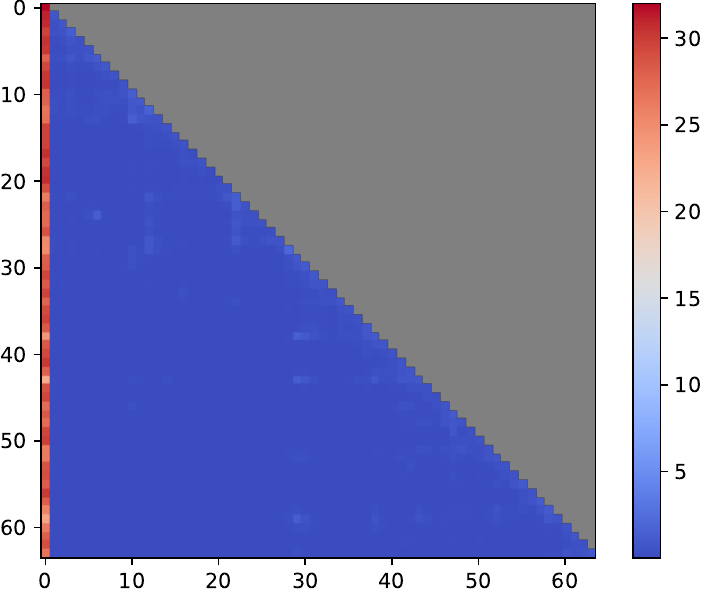}
    }
	\caption{Magnitude of the output activations and attention map in LLaMA-3-8B.
 }
\end{figure*}
\captionsetup[subfloat]{labelsep=none,format=plain,labelformat=empty,justification=centering}
\captionsetup[subfloat]{labelsep=none,format=plain,labelformat=empty,justification=centering}
\captionsetup[subfloat]{labelsep=none,format=plain,labelformat=empty,justification=centering}
\begin{figure*}[t]
\centering
\subfloat[(a) Output activations of Vicuna-v1.3-13B Layer 8]{
        \includegraphics[width=0.24\linewidth]{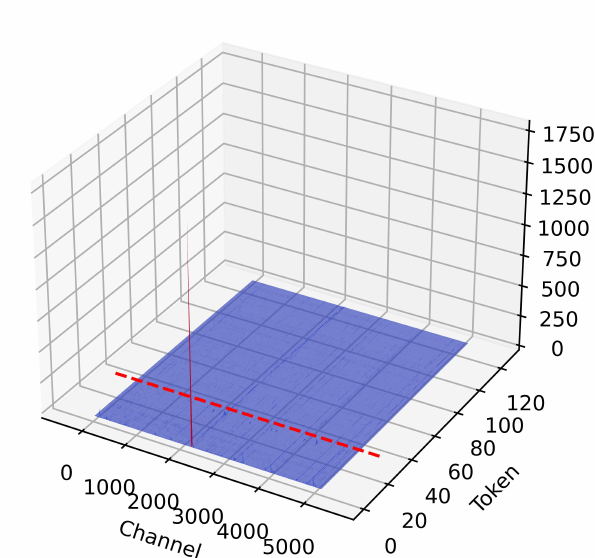}
    }
\subfloat[(b) Output activations of Vicuna-v1.3-13B Layer 16]{
        \includegraphics[width=0.24\linewidth]{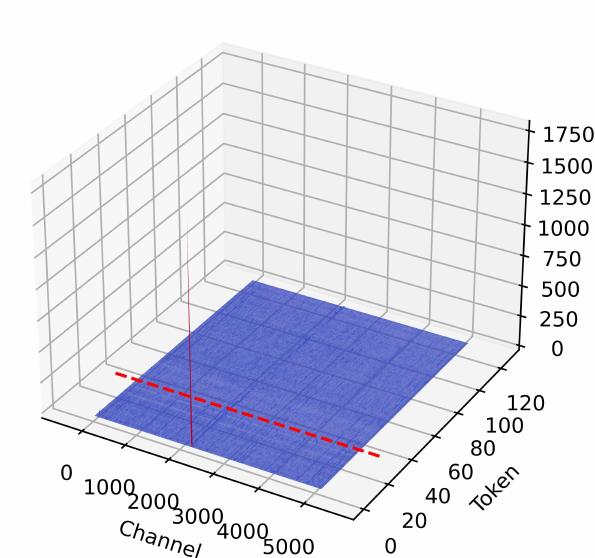}
    }
\subfloat[(c) Output activations of Vicuna-v1.3-13B Layer 24]{
        \includegraphics[width=0.24\linewidth]{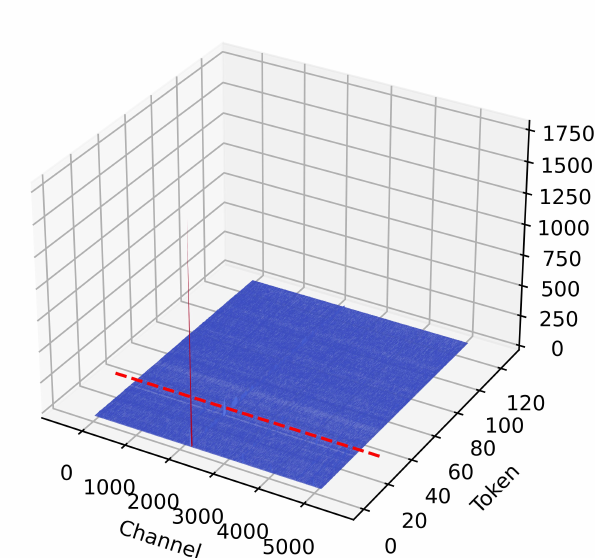}
    }
\subfloat[(d) Output activations of Vicuna-v1.3-13B Layer 32]{
        \includegraphics[width=0.24\linewidth]{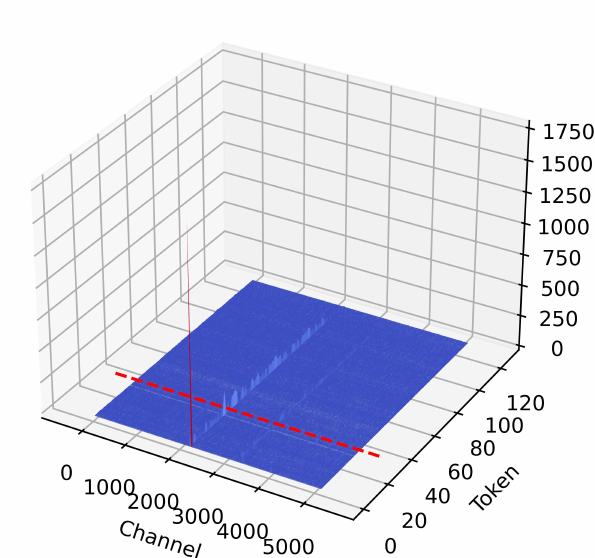}
    }
\vspace{0.25cm}
\subfloat[(e) Attention map of \\Vicuna-v1.3-13B Layer 8]{
        \includegraphics[width=0.24\linewidth]{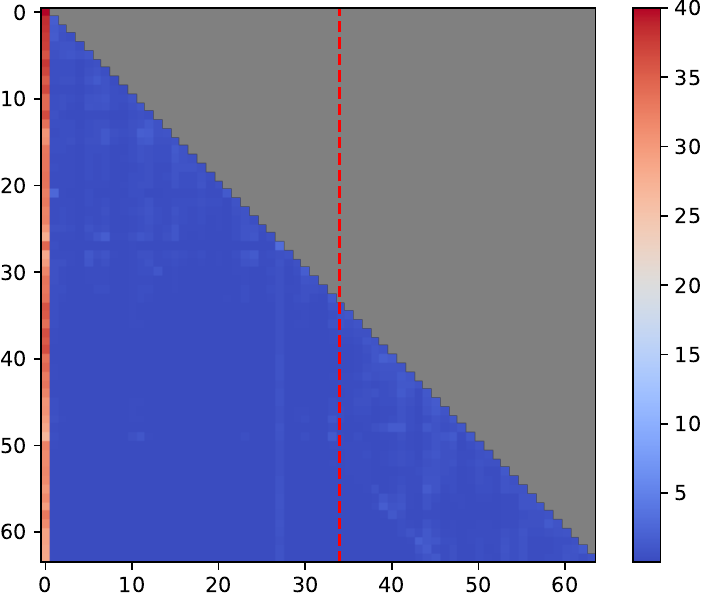}
    }
\subfloat[(f) Attention map of \\Vicuna-v1.3-13B Layer 16]{
        \includegraphics[width=0.24\linewidth]{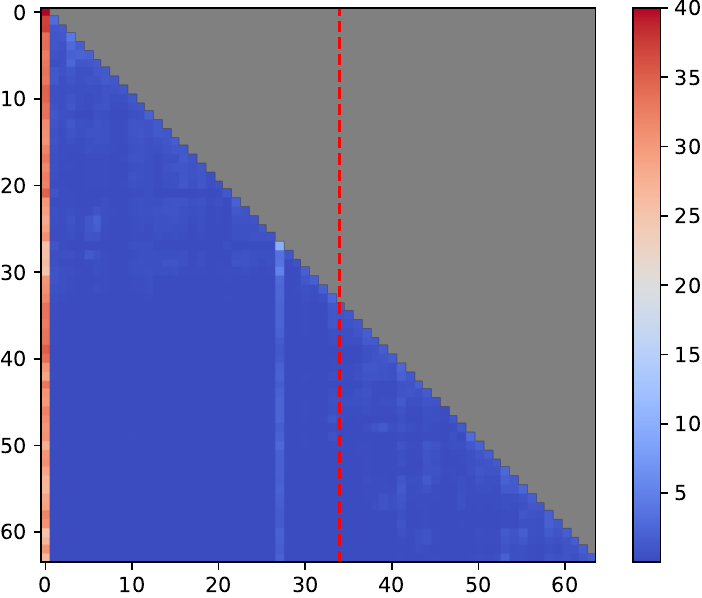}
    }
\subfloat[(g) Attention map of \\Vicuna-v1.3-13B Layer 24]{
        \includegraphics[width=0.24\linewidth]{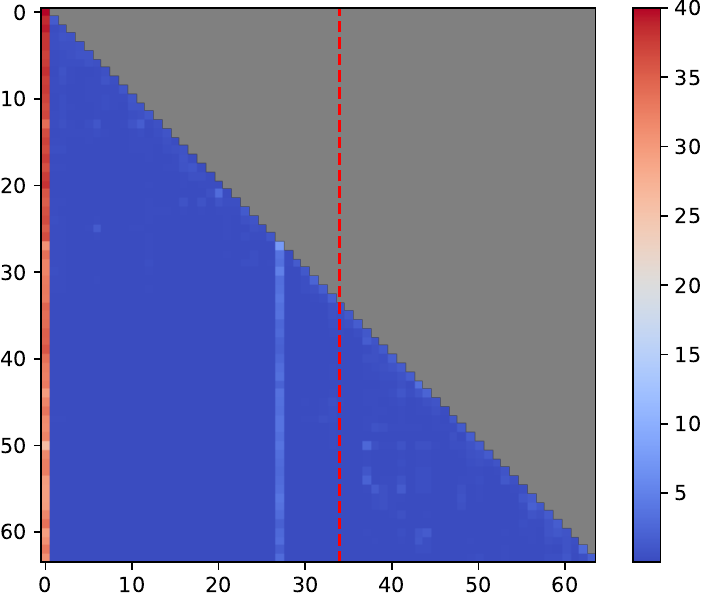}
    }
\subfloat[(h) Attention map of \\Vicuna-v1.3-13B Layer 32]{
        \includegraphics[width=0.24\linewidth]{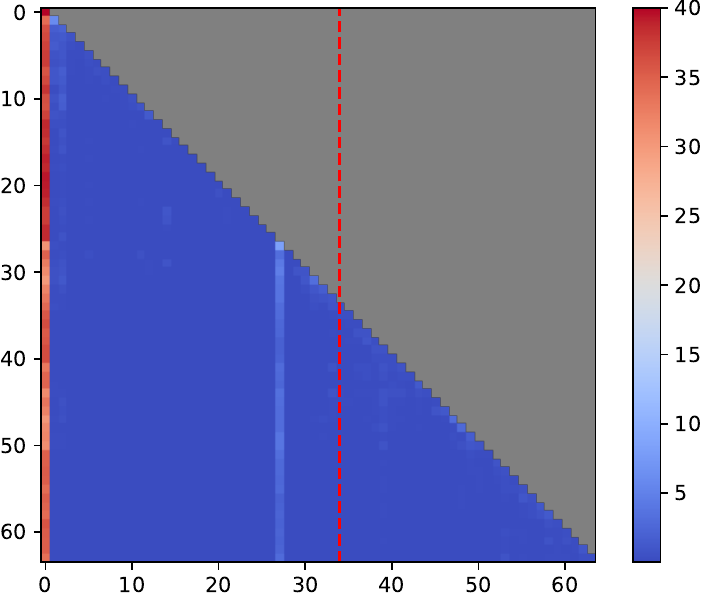}
    }
	\caption{Magnitude of the output activations and attention map in Vicuna-v1.3-13B. The tokens before the red dashed line correspond to the Vicuna system prompt.
 }
\end{figure*}
\captionsetup[subfloat]{labelsep=none,format=plain,labelformat=empty,justification=centering}
\begin{figure*}[t]
\centering
\subfloat[(a) Output activations of Vicuna-v1.3-33B Layer 8]{
        \includegraphics[width=0.24\linewidth]{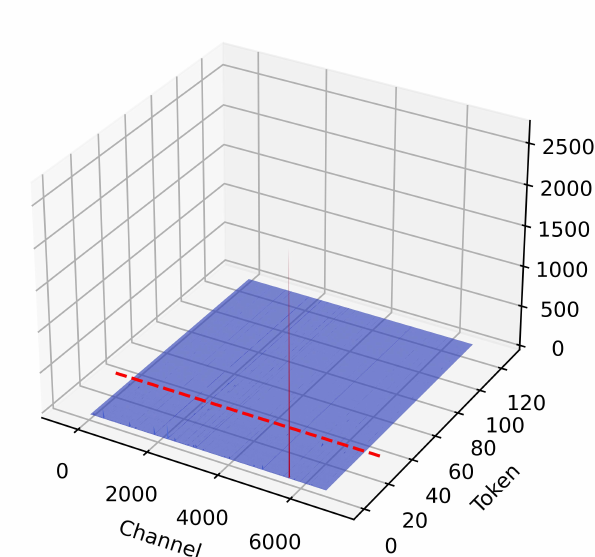}
    }
\subfloat[(b) Output activations of Vicuna-v1.3-33B Layer 24]{
        \includegraphics[width=0.24\linewidth]{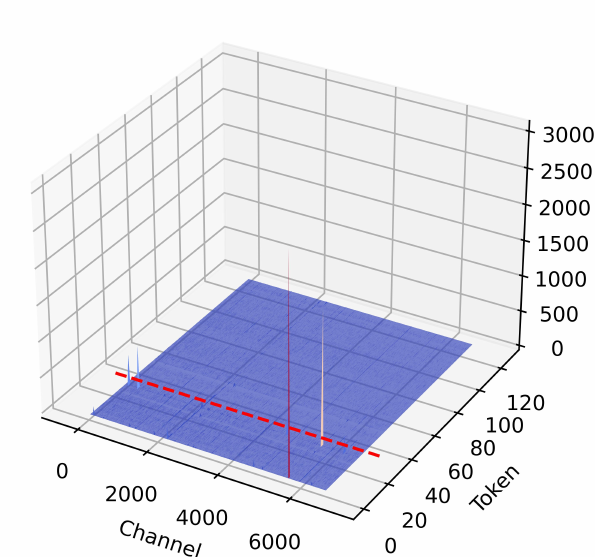}
    }
\subfloat[(c) Output activations of Vicuna-v1.3-33B Layer 40]{
        \includegraphics[width=0.24\linewidth]{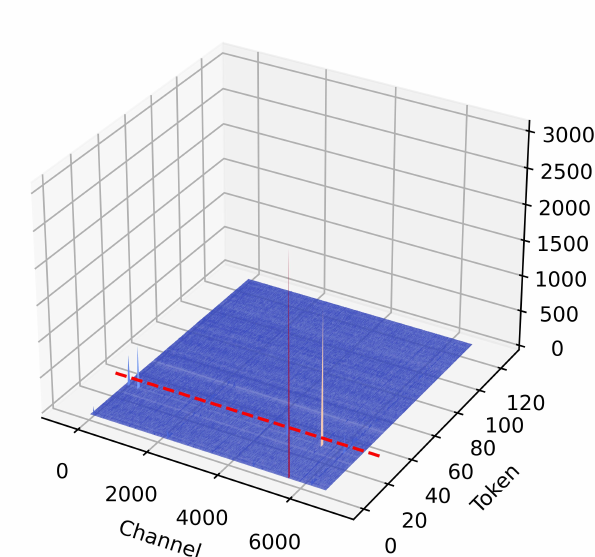}
    }
\subfloat[(d) Output activations of Vicuna-v1.3-33B Layer 56]{
        \includegraphics[width=0.24\linewidth]{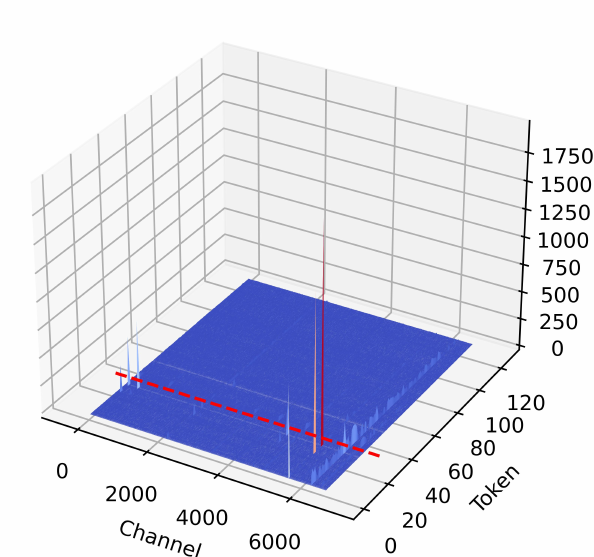}
    }
\vspace{0.25cm}
\subfloat[(e) Attention map of \\Vicuna-v1.3-33B Layer 8]{
        \includegraphics[width=0.24\linewidth]{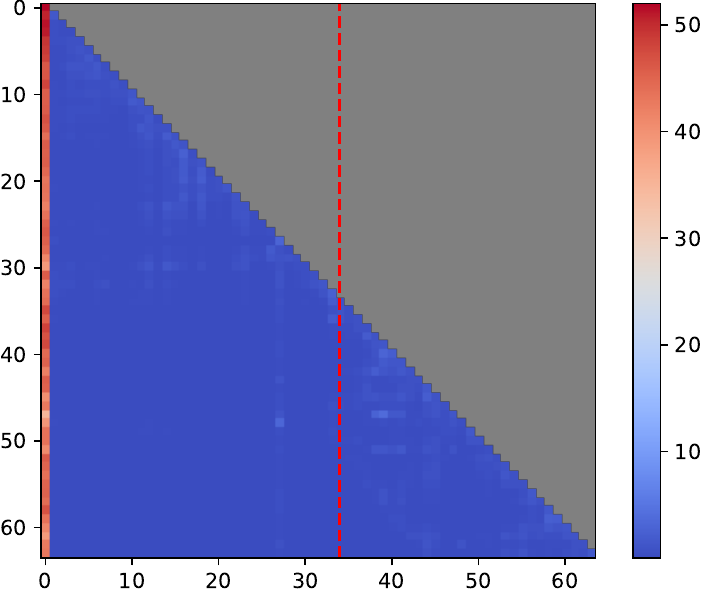}
    }
\subfloat[(f) Attention map of \\Vicuna-v1.3-33B Layer 24]{
        \includegraphics[width=0.24\linewidth]{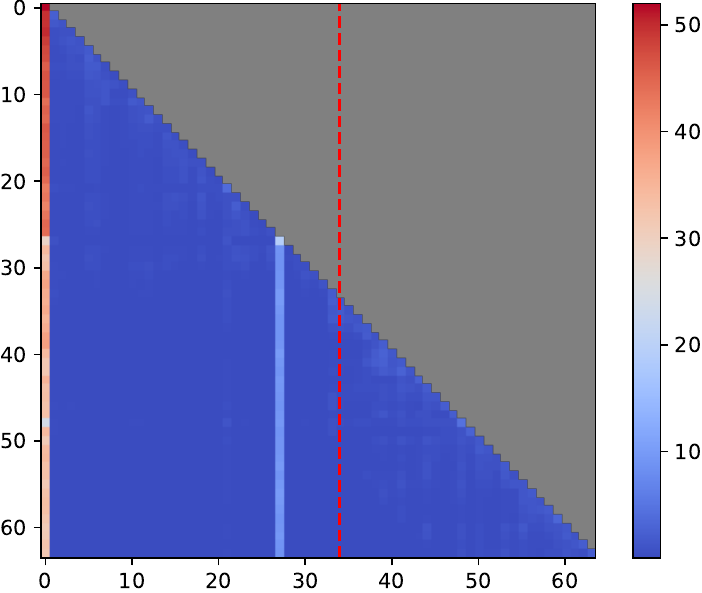}
    }
\subfloat[(g) Attention map of \\Vicuna-v1.3-33B Layer 40]{
        \includegraphics[width=0.24\linewidth]{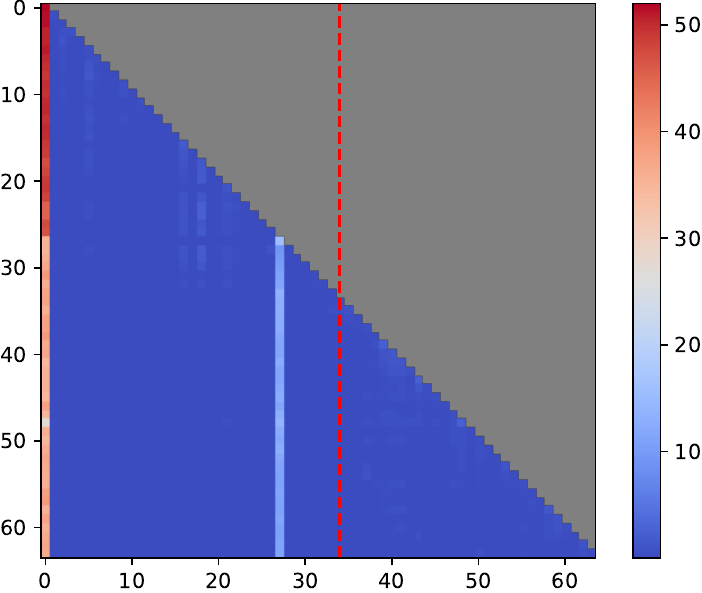}
    }
\subfloat[(h) Attention map of \\Vicuna-v1.3-33B Layer 56]{
        \includegraphics[width=0.24\linewidth]{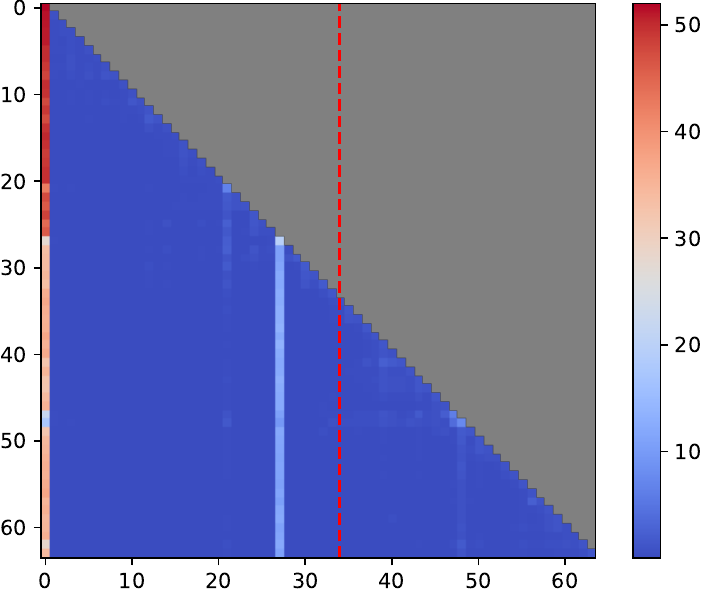}
    }
	\caption{Magnitude of the output activations and attention map in Vicuna-v1.3-33B. The tokens before the red dashed line correspond to the Vicuna system prompt.
 }
\end{figure*}
\captionsetup[subfloat]{labelsep=none,format=plain,labelformat=empty,justification=centering}
\begin{figure*}[t]
\centering
\subfloat[(a) Output activations of Vicuna-v1.5-7B Layer 0]{
        \includegraphics[width=0.24\linewidth]{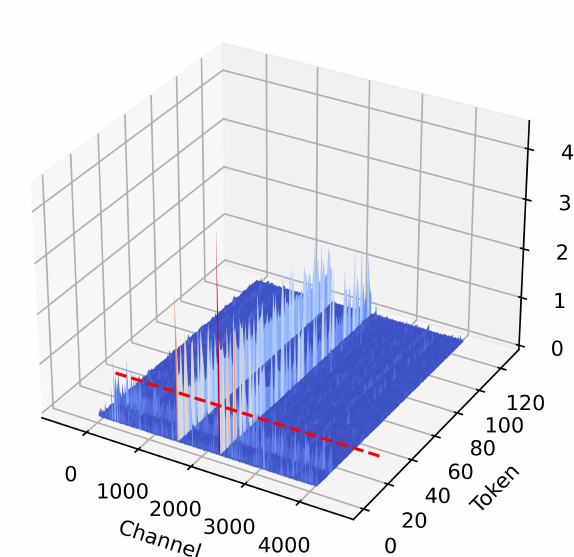}
    }
\subfloat[(b) Output activations of Vicuna-v1.5-7B Layer 8]{
        \includegraphics[width=0.24\linewidth]{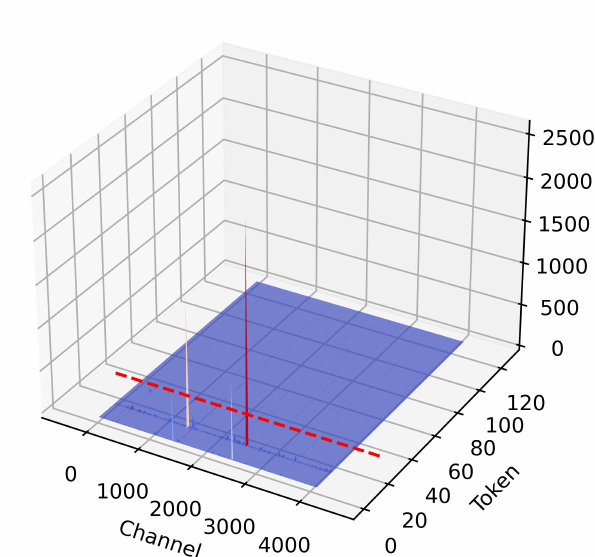}
    }
\subfloat[(c) Output activations of Vicuna-v1.5-7B Layer 16]{
        \includegraphics[width=0.24\linewidth]{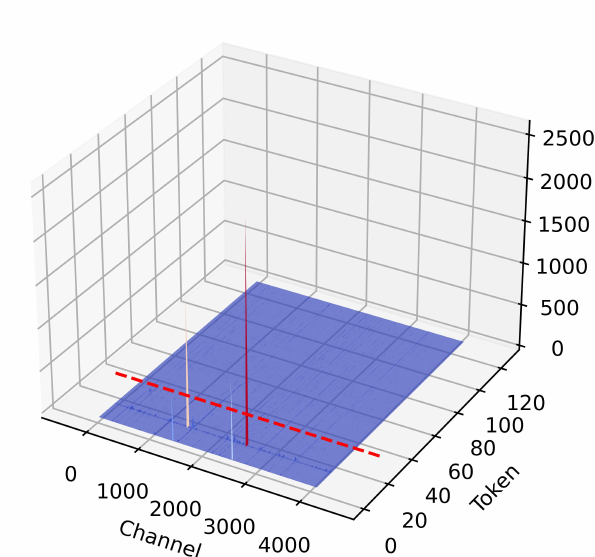}
    }
\subfloat[(d) Output activations of Vicuna-v1.5-7B Layer 24]{
        \includegraphics[width=0.24\linewidth]{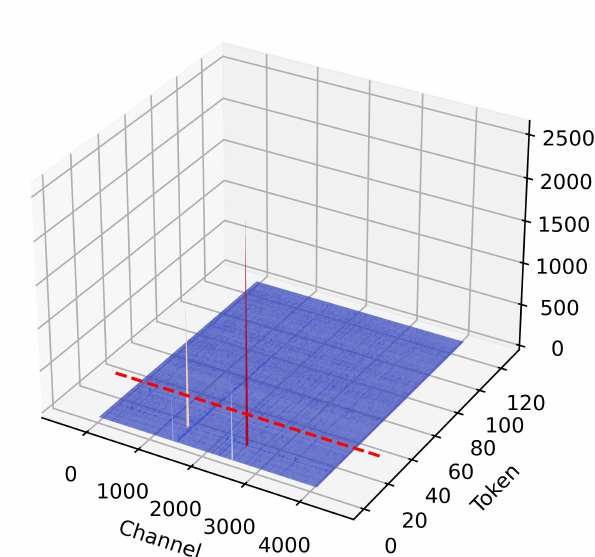}
    }
\vspace{0.25cm}
\subfloat[(e) Attention map of \\Vicuna-v1.5-7B Layer 0]{
        \includegraphics[width=0.24\linewidth]{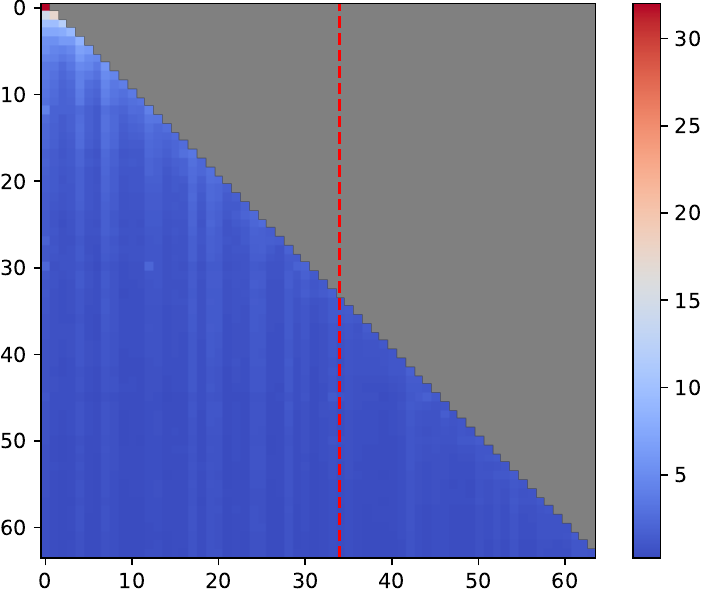}
    }
\subfloat[(f) Attention map of \\Vicuna-v1.5-7B Layer 8]{
        \includegraphics[width=0.24\linewidth]{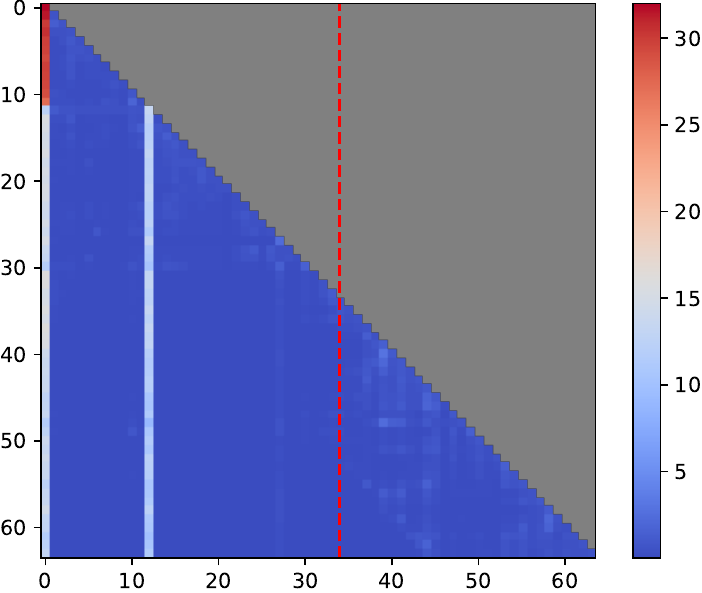}
    }
\subfloat[(g) Attention map of \\Vicuna-v1.5-7B Layer 16]{
        \includegraphics[width=0.24\linewidth]{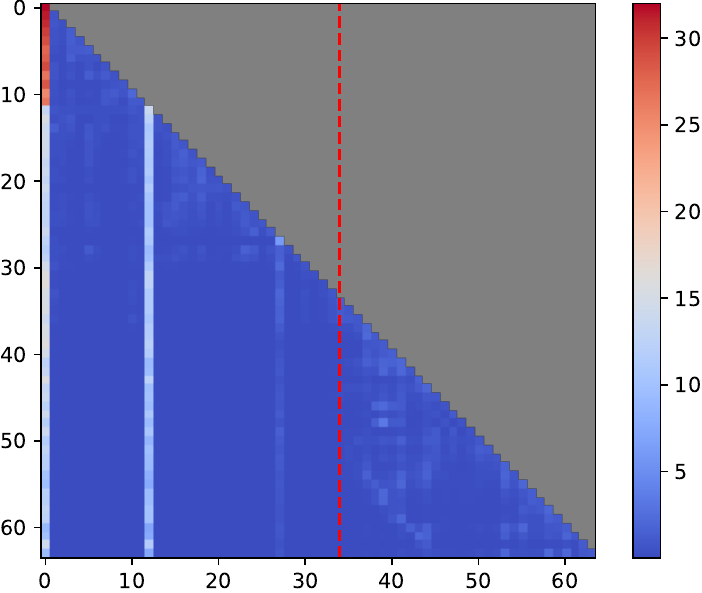}
    }
\subfloat[(h) Attention map of \\Vicuna-v1.5-7B Layer 24]{
        \includegraphics[width=0.24\linewidth]{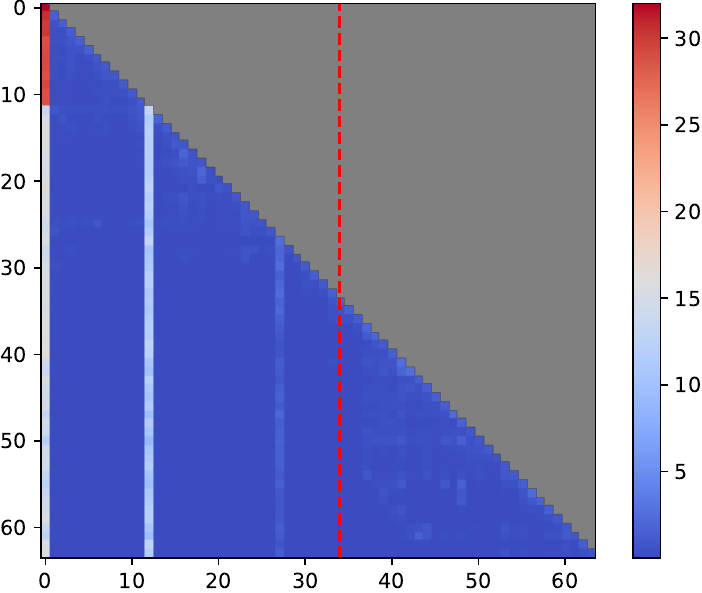}
    }
	\caption{Magnitude of the output activations and attention map in Vicuna-v1.5-7B. The tokens before the red dashed line correspond to the Vicuna system prompt.
 }
\end{figure*}
\captionsetup[subfloat]{labelsep=none,format=plain,labelformat=empty,justification=centering}
\captionsetup[subfloat]{labelsep=none,format=plain,labelformat=empty,justification=centering}
\captionsetup[subfloat]{labelsep=none,format=plain,labelformat=empty,justification=centering}

\end{document}